\documentclass[10pt,journal,compsoc]{IEEEtran}
\usepackage{times}
\usepackage{epsfig}
\usepackage{graphicx}
\usepackage{subfigure}
\usepackage{amsmath}
\usepackage{amssymb}
\usepackage{booktabs}
\usepackage{cite}
\usepackage{multirow}
\usepackage{CJKutf8}
\usepackage{booktabs}
\usepackage{dsfont}
\usepackage{algorithmic}
\usepackage{setspace}
\usepackage{algorithm}
\usepackage{xcolor}
\usepackage{url}
\usepackage[hidelinks]{hyperref}
\allowdisplaybreaks

\usepackage{eqparbox}

\newcommand{\hc}[1]{{\color{black} #1}}
\newcommand{\rhc}[1]{{\color{black} #1}}
\begin{document}
%
\title{A Self-Supervised Gait Encoding Approach with Locality-Awareness for 3D Skeleton Based Person Re-Identification}

%
%
%
%

\author{Haocong~Rao,
        Siqi~Wang,
        Xiping~Hu,
        Mingkui~Tan,
        Yi~Guo,
        Jun~Cheng,
        Xinwang~Liu,
        and~Bin~Hu

\IEEEcompsocitemizethanks{
\IEEEcompsocthanksitem Haocong Rao and Siqi Wang contributed equally to this work (Corresponding authors: Xiping Hu; Bin Hu).    
\IEEEcompsocthanksitem H. Rao and J. Cheng are with Shenzhen Institute of Advanced Technology, Chinese Academy of Sciences, Shenzhen 518055, China. E-mail: \{hc.rao, jun.cheng\}@siat.ac.cn, haocongrao@gmail.com.
\IEEEcompsocthanksitem S. Wang and X. Liu are with the National University of Defense Technology, Changsha 410073, China. Email: \{wangsiqi10c, xinwangliu\}@nudt.edu.cn.
\IEEEcompsocthanksitem X. Hu is with Shenzhen Institute of Advanced Technology, Chinese Academy of Sciences, Shenzhen 518055, China, and also with Lanzhou University, Gansu 730000, China. E-mail:  xp.hu@siat.ac.cn, huxp@lzu.edu.cn.
\IEEEcompsocthanksitem M. Tan is with South China University of Technology, Guangzhou 510006, China. Email: mingkuitan@scut.edu.cn.
\IEEEcompsocthanksitem Y. Guo is with the Second Clinical Medical College, Jinan University, Shenzhen 518055, China. E-mail: xuanyi\_guo@163.com.
\IEEEcompsocthanksitem B. Hu is with Beijing Institute of Technology, Beijing 100081, China, and also with Lanzhou University, Gansu 730000, China. Email: bh@bit.edu.cn, bh@lzu.edu.cn.

}
} 

%
%

\markboth{IEEE TRANSACTIONS ON PATTERN ANALYSIS AND MACHINE INTELLIGENCE, ~Vol.~X, No.~X, July~2021}%
{Shell \MakeLowercase{\textit{et al.}}: Bare Demo of IEEEtran.cls for Computer Society Journals}
%



\IEEEtitleabstractindextext{%
\begin{abstract}

Person re-identification (Re-ID) via gait features within 3D skeleton sequences is a newly-emerging topic with several advantages. Existing solutions either rely on hand-crafted descriptors or supervised gait representation learning. This paper proposes a \textit{self-supervised} gait encoding approach that can leverage \textit{unlabeled} skeleton data to learn gait representations for person Re-ID. Specifically, we first create self-supervision by learning to reconstruct unlabeled skeleton sequences reversely, which involves richer high-level semantics to obtain better gait representations. Other pretext tasks are also explored to further improve self-supervised learning. Second, inspired by the fact that motion's continuity endows adjacent skeletons in one skeleton sequence and temporally consecutive skeleton sequences with higher correlations (referred as \textit{locality} in 3D skeleton data), we propose a locality-aware attention mechanism and a locality-aware contrastive learning scheme, which aim to preserve locality-awareness on intra-sequence level and inter-sequence level respectively during self-supervised learning. Last, with context vectors learned by our locality-aware attention mechanism and contrastive learning scheme, a novel feature named Constrastive Attention-based Gait Encodings (CAGEs) is designed to represent gait effectively. Empirical evaluations show that our approach significantly outperforms skeleton-based counterparts by 15-40\% \textit{Rank-1} accuracy, and it even achieves superior performance to numerous multi-modal methods with extra RGB or depth information. 
Our codes are available at \href{https://github.com/Kali-Hac/Locality-Awareness-SGE}{https://github.com/Kali-Hac/Locality-Awareness-SGE.}

\end{abstract}

}

\maketitle

\IEEEdisplaynontitleabstractindextext

%
\IEEEpeerreviewmaketitle

\IEEEraisesectionheading{\section{Introduction}\label{sec:introduction}}

%
%
%
%
\IEEEPARstart{T}{he} goal of person re-identification (Re-ID) is to re-identify the same person in a different scene or view. It plays a pivotal role in various applications like security authentication, human tracking, and role-based activity understanding \cite{loy2010time,baltieri2011sarc3d,vezzani2013people,wang2016person,zhao2017person,su2018multi,chen2018person,li2019unsupervised,qian2019leader,yu2020unsupervised}. To perform person Re-ID effectively, \textit{gait} is one of the most useful human body clues, and it has aroused a growing interest in the research community since gait can be collected by unobtrusive methods without cooperative subjects \cite{connor2018biometric}. Physiological and psychological studies \cite{murray1964walking,cutting1977recognizing} reveal that human individuals behave differently when walking, and they are endowed with some relatively stable gait patterns ($e.g.$, stride length, angles of body joints). Such unique gait patterns are usually highly valuable for high-level tasks like gait recognition \cite{liao2020model} and person Re-ID \cite{andersson2015person}.

\begin{figure}[t]
\begin{center}
\scalebox{0.38}{
  \includegraphics[]{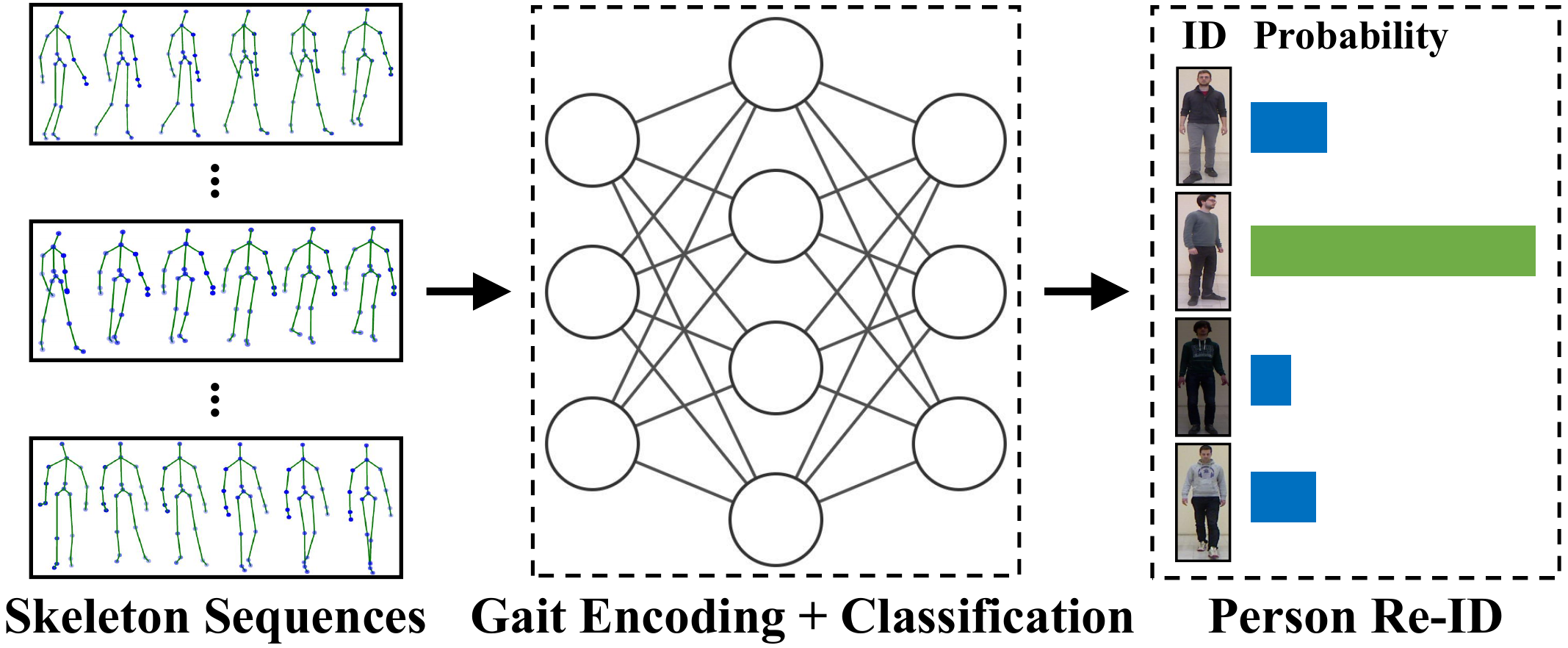}
  }
  \caption{Gait-based person Re-ID using 3D skeleton data.}
\label{figure1}
\end{center}
\end{figure}

To perform gait analysis, gait is typically described by two types of methods: \textbf{(1)} \textit{Appearance}-based methods \cite{wang2003silhouette,veeraraghavan2005matching,liu2006improved,guan2014reducing,chunli2010behavior,sivapalan2011gait,wang2011human,zhang2019comprehensive}, which leverage human silhouettes from aligned image sequences to depict gait. However, an important flaw of this type of methods is its vulnerability to body shape changes and appearance changes. \textbf{(2)} \textit{Model}-based methods \cite{991036,Wang2004149,6117582,andersson2015person,liao2020model}, which model gait by human body structure and motion of body joints. Unlike appearance-based methods, model-based methods are invariant to factors like scale and view \cite{nambiar2019gait}. Therefore, model-based methods possess better robustness in practice. Among various models, \textit{3D skeleton} model describes humans by the 3D coordinates of numerous key body joints, and it can often be used as a highly efficient representation of human body structure and motion \cite{han2017space}. 3D skeleton data are easily accessible with popular devices like Kinect \cite{shotton2011real-time}, and they enjoy several prominent advantages when compared with frequently-seen RGB or depth data. For example, 3D skeleton data are much less likely to be interfered by illumination changes than RGB data, and they enjoy much smaller data size and less information redundancy than depth data \cite{han2017space}. Therefore, exploiting 3D skeleton data to perform gait analysis for downstream tasks like person Re-ID (illustrated in Fig. \ref{figure1}) is an attractive and promising solution with increasing popularity \cite{liao2020model}. Nevertheless, the way to extract or learn discriminative gait features from 3D skeleton data remains to be an open problem.

For this purpose, most existing works like \cite{barbosa2012re,andersson2015person} resort to hand-crafted skeleton feature descriptors. They typically focus on describing human bodies in terms of geometric, morphological, and anthropometric attributes, and then extracting corresponding features from 3D skeleton data. However, hand-crafted feature engineering is usually complicated and tedious. For instance, the method in \cite{andersson2015person} requires defining 80 skeleton descriptors from the views of anthropometric and gait attributes for person Re-ID. Besides, such methods also heavily rely on domain knowledge like human anatomy \cite{yoo2002extracting}, thus lacking the ability to mine useful latent gait features beyond human comprehension. Motivated by the limitations of hand-crafted skeleton descriptors and the remarkable success achieved by recent deep neural networks (DNNs), few recent works like \cite{haque2016recurrent} resort to DNNs to learn gait representations automatically. However, the gait encoding process of such methods unexceptionally follows the classic \textit{supervised} learning paradigm, which requires the discriminative information from labeled 3D skeleton data. As a consequence, they cannot utilize unlabeled skeleton data directly for automatic gait encoding. 

\rhc{This paper for the first time proposes a 3D skeleton based person Re-ID approach guided by \textbf{\textit{self-supervision}} and \textbf{\textit{locality-awareness}}, and it realizes highly effective gait encoding with unlabeled 3D skeleton sequence data. By first creating self-supervision signals for gait encoding, our approach not only makes it possible to learn gait representations from unlabeled skeleton data, but also prompts learning richer high-level semantics ($e.g.$, sequence order, body part motion) and more discriminative gait features. To be more specific, we propose to leverage the reverse reconstruction of skeleton sequences as a primary self-supervised learning objective. Meanwhile, we also explore other pretext tasks and utilize them to further enhance self-supervised learning. Second, we notice that 3D skeleton sequences are endowed with a property named \textit{locality}: The continuity of human motion usually induces very small pose/skeleton changes in a local temporal interval \cite{aggarwal1999human}. As a result, for each skeleton in a skeleton sequence, its adjacent skeletons have higher correlations to itself, which is referred as \textit{intra-sequence} locality. Similarly, we also define \textit{inter-sequence} locality, which suggests that two temporally consecutive 3D skeleton sequences also enjoy higher relevance. To this end, we propose to incorporate \textit{locality-awareness} to enable better 3D skeleton reconstruction and gait encoding during self-supervised learning. Accordingly, during the gait encoding process, we propose a novel locality-aware attention mechanism and locality-aware contrasitive learning scheme to preserve locality on the intra-sequence and inter-sequence level respectively. Last, based on the proposed locality-aware attention mechanism and locality-aware contrastive learning, we devise a novel method to construct our gait representations, which 
are named Contrastive Attention-based Gait Encodings (CAGEs), from the learned model. Our empirical evaluations demonstrate that CAGEs, which can be learned without any skeleton label, can be directly applied to person Re-ID and achieve highly competitive performance.}

A preliminary version of this work was reported in \cite{DBLP:conf/ijcai/RaoW0TD0020}. Compared with \cite{DBLP:conf/ijcai/RaoW0TD0020}, this work not only systematically explores the design of self-supervised learning objective for 3D skeleton sequences with more pretext tasks, but also extends the conception of locality from the intra-sequence level to the inter-sequence level by devising the locality-aware contrastive learning. To our knowledge, this is also the first attempt that explores the contrastive learning technique for learning discriminative gait features. \rhc{In particular, we need to point out that ``self-supervised learning'' in this paper still refers to learning with our designed pretext tasks (\textit{e.g.,} reverse reconstruction or prediction of the 3D skeleton sequences). Although contrastive learning is a popular technique to realize self-supervised learning in the literature, it is specifically designed to encourage sequence-level locality here and should be distinguished from the previous ``self-supervised learning'' term.}  On the foundation of those improvements, this work also improves earlier gait features proposed in \cite{DBLP:conf/ijcai/RaoW0TD0020} into the more effective gait features CAGEs. To validate those improvements, this work carries out more extensive experiments and detailed discussions on three public Re-ID datasets and a new multi-view Re-ID dataset \cite{nambiar2017context}. \hc{Besides, we demonstrate that our approach is also effective with the skeleton data estimated from RGB videos \cite{yu2006framework}.
}
To sum up, our contributions can be summarized as follows:
\begin{itemize}
\item We propose a new self-supervised learning paradigm for the gait encoding of 3D skeleton based person Re-ID. The proposed paradigm enables us to yield more effective gait representations from unlabeled 3D skeleton sequences by learning a reverse sequential skeleton reconstruction. 

\item We explore other possible forms of pretext tasks for the proposed self-supervised learning paradigm, and showcase their effectiveness in further strengthening gait encoding.

\item We devise a locality-aware attention mechanism to exploit the intra-sequence locality within skeletons of one skeleton sequence, so as to facilitate better skeleton reconstruction and gait encoding during the self-supervised learning.

\item We propose a locality-aware contrastive learning scheme to preserve the inter-sequence locality among temporally adjacent 3D skeleton sequences, which is able to encourage better gait encoding on the sequence level. 

\item We propose a new method to construct our gait representations (CAGEs) from the learned model. CAGEs are shown to be highly effective for person Re-ID.

\end{itemize}

The rest of paper is organized as follows: Sec. \ref{related} introduces relevant works in the literature. Sec. \ref{proposed} elucidates each module of the proposed approach. Sec. \ref{experiments} presents the details of experiments, and extensively compares our approach with existing solutions. Sec. \ref{discuss} provides ablation studies and comprehensive discussions on the proposed approach. Sec. \ref{conclusion} concludes this paper.

\begin{figure*}[t]
    \centering
    \scalebox{0.70}{
    \includegraphics{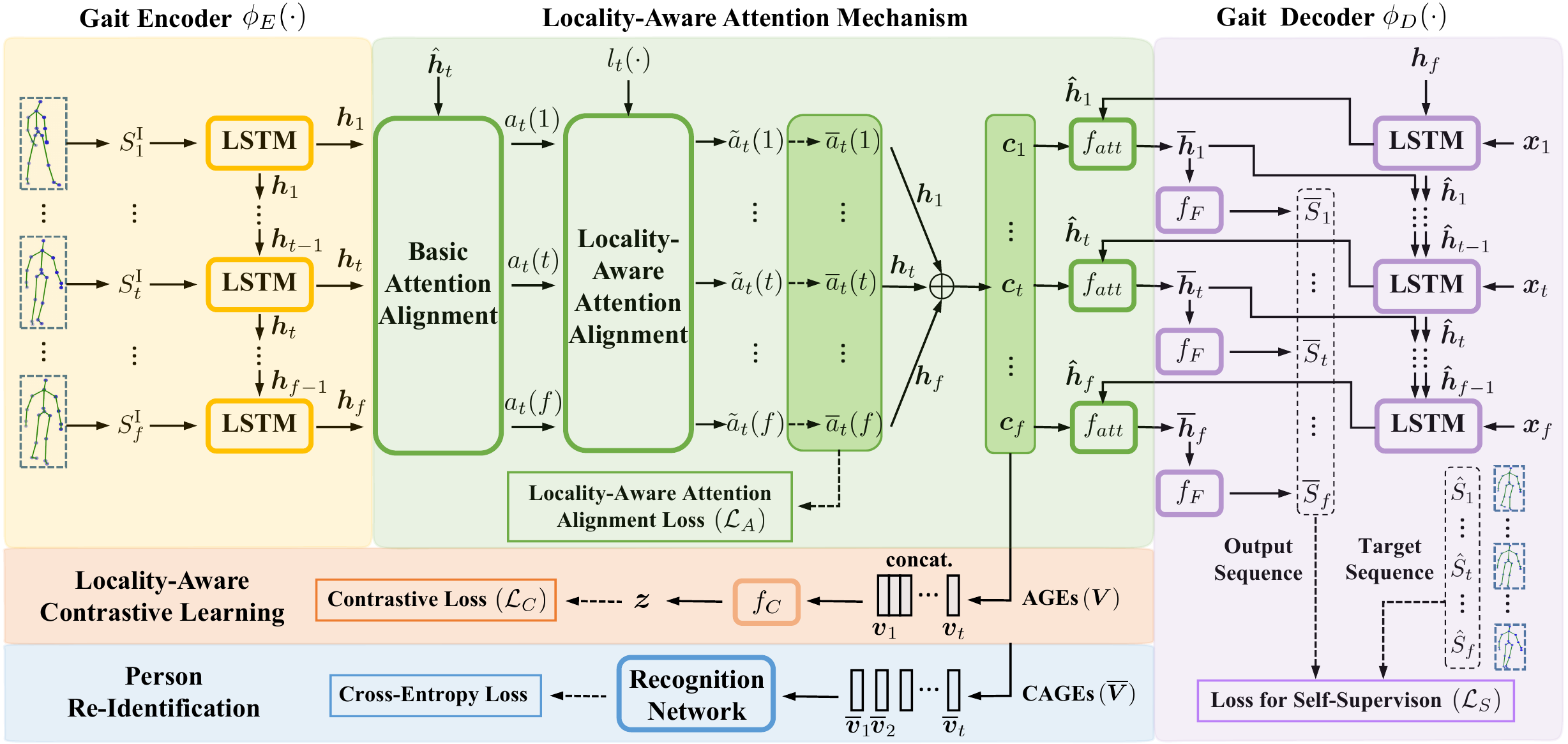}
    }
    \caption{Flow diagram of our model: (1) Gait Encoder (yellow) encodes each skeleton frame $S^{\text{I}}_t$ into an encoded gait state $\boldsymbol{h}_t$. (2) Locality-aware attention mechanism (green) first computes the basic attention alignment score ${a}_{t}(\cdot)$, so as to measure the content-based correlation between each encoded gait state and the decoded gait state $\boldsymbol{{\hat h}}_{t}$ from Gait Decoder (purple). Then, the locality mask $l_{t}(\cdot)$ provides an objective $\tilde{a}_{t}(\cdot)\!=\!{a}_{t}(\cdot)\ l_{t}(\cdot)$, which guides our model to learn locality-aware alignment scores $\overline{a}_{t}(\cdot)$ by the locality-aware attention alignment loss ${\mathcal L}_{A}$. 
    Next, $\boldsymbol{h}_1\cdots \boldsymbol{h}_f$ are weighted by $\overline{a}_{t}(\cdot)$ to compute the context vector $\boldsymbol{c}_{t}$. $\boldsymbol{c}_{t}$ and $\boldsymbol{{\hat h}}_{t}$ are fed into the concatenation layer $f_{att}(\cdot)$ to produce an attentional state vector $\boldsymbol{\overline{h}}_{t}$.
    Finally, $\boldsymbol{\overline{h}}_{t}$ is fed into the full connected layer $f_{F}(\cdot)$ to output $t^{th}$ skeleton $\boldsymbol{\overline{S}}_{t}$ and Gait Decoder for later decoding. (3) $\boldsymbol{c}_{t}$ is used to build Attention-based Gait Encodings (AGEs) $\boldsymbol{v}_{t}$, which are concatenated and fed into $f_{C}(\cdot)$ to perform locality-aware contrastive learning (red). (4) The Contrastive Attention-based Gait Encodings (CAGEs) $\boldsymbol{\overline{v}}_{t}$ learned by
   contrastive learning are fed into a recognition network for person Re-ID (blue).}
\label{model}
\end{figure*}

\section{Related Works}
\label{related}
\subsection{Person Re-Identification}
\quad \ \textit{Skeleton-based person Re-ID.} As an emerging topic, most existing works extract hand-crafted features to depict certain geometric, morphological or anthropometric attributes of 3D skeleton data.
Barbosa \textit{et al.} \cite{barbosa2012re} compute 7 Euclidean distances between the floor plane and joint or the joint pair to construct a distance matrix, which is learned by a quasi-exhaustive strategy to perform person Re-ID. 
Munaro \textit{et al.} \cite{munaro2014one} further extend them to 13 skeleton descriptors ($D^{13}$) and use support vector machine (SVM) and $k$-nearest neighbor (KNN) for classification.
In \cite{pala2019enhanced}, 16 Euclidean distances between body joints ($D^{16}$) are fed to an Adaboost classifier for Re-ID.
Since existing solutions that use features from 3D skeletons alone usually perform unsatisfactorily, features from other modalities ($e.g.$, 3D point clouds \cite{munaro20143d}, 3D face descriptor \cite{pala2019enhanced}) are often used to enhance the performance. Meanwhile, few recent works exploit supervised deep learning models to learn gait representations from skeleton data: In \cite{battistone2018tglstm}, a Time based Graph (TG) LSTM model is proposed for human recognition based on skeletal graphs that are transformed from binary images. \cite{haque2016recurrent} utilizes long short-term memory (LSTM) \cite{hochreiter1997long} to model temporal dynamics of body joints to perform person Re-ID; The latest work from Liao \textit{et al.} \cite{liao2020model} propose PoseGait, which feeds 81 hand-crafted pose features of 3D skeleton data into convolutional neural networks (CNN) for human recognition.

Our work differs from previous skeleton-based works in following aspects: (1) We propose a novel self-supervised approach to encode discriminative gait features from unlabeled 3D skeleton data. We do NOT need to extract hand-crafted features like \cite{barbosa2012re,munaro2014one,pala2019enhanced} or use identity labels to supervise gait representation learning \cite{battistone2018tglstm,haque2016recurrent,liao2020model}. In this work, the reverse skeleton reconstruction is proposed as the major pretext task to capture high-level semantics like skeleton motion patterns in unlabeled skeleton data, which facilitates us to yield more effective gait representations.
 (2) The property of locality induced by motion's continuity is exploited for better gait encoding: We propose the locality-aware attention mechanism and locality-aware contrastive learning scheme to preserve intra-sequence and inter-sequence locality embedded in 3D skeleton sequences respectively. To our best knowledge, this is also the first attempt to leverage attention mechanism and contrastive learning to realize gait encoding.

\textit{Depth-based and multi-modal person Re-ID.} Depth-based methods typically exploit human shapes or silhouettes from depth images to extract gait features for person Re-ID. For example, Sivapalan \textit{et al.} \cite{sivapalan2011gait} extend the Gait Energy Image (GEI) \cite{chunli2010behavior} to 3D domain and propose Gait Energy Volume (GEV) algorithm based on depth images to perform gait-based human recognition. 3D point clouds from depth data are also pervasively used to estimate body shape and motion trajectories. Munaro \textit{et al.} \cite{munaro2014one} propose point cloud matching (PCM) to compute the distances of multi-view point cloud sets, so as to discriminate different persons. Haque \textit{et al.} \cite{haque2016recurrent} adopt 3D LSTM to model motion dynamics of 3D point clouds for person Re-ID. As to multi-modal methods, they usually combine skeleton-based features with extra RGB or depth information ($e.g.,$ depth shape features based on point clouds \cite{munaro20143d,hasan2016long,wu2017robust}) to boost Re-ID performance. In \cite{karianakis2018reinforced}, CNN-LSTM with reinforced temporal attention (RTA) is proposed for person Re-ID based on a split-rate RGB-depth transfer approach.

\subsection{Contrastive Learning}
Recent years witness a surging popularity of contrastive learning in the unsupervised learning field \cite{hadsell2006dimensionality,wu2018unsupervised,oord2018representation,zhuang2019local}. It aims to learn effective data representations by separating positive pairs from negative pairs with contrastive losses, which measure the similarity of sample pairs in a latent representation space, and they are often combined with various pretext tasks to enhance unsupervised learning. To name a few, Wu \textit{et al.} \cite{wu2018unsupervised} propose an instance-level discrimination method based on exemplar related task \cite{dosovitskiy2014discriminative} and noise-contrastive estimation (NCE) \cite{gutmann2010noise}. Contrastive predictive coding (CPC) \cite{oord2018representation} adopts the context auto-encoding task with a probabilistic contrastive loss (InfoNCE) to learn representations from different modalities. Many previous works \cite{zhuang2019local,tian2019contrastive,misra2020self} adopt the memory bank \cite{wu2018unsupervised} to store the representation vectors of samples in the dataset, while some recent advances \cite{ye2019unsupervised,ji2019invariant,chen2020a} explore the use of in-batch samples for negative sampling instead of a memory bank. The latest contrastive learning framework is SimCLR \cite{chen2020a,NEURIPS2020_fcbc95cc}, which is highly efficient for unsupervised visual representation learning and inspires our work for skeletons. 

Our work differs from previous studies in the following aspects: (1) The proposed locality-aware contrastive learning scheme is proposed to incorporate the inter-sequence locality into the gait encoding process, during which the sequence-level representation of 3D skeleton sequence is viewed as an instance in contrastive learning. (2) The goal of locality-aware contrastive learning scheme is to maximize the agreement between adjacent sequences that enjoy higher correlations. Different from \cite{chen2020a,he2020momentum} that use augmented samples of images as contrastive instances, we exploit consecutive and non-consecutive 3D skeleton sequences as positive and negative pairs respectively for contrastive learning.

\section{The Proposed Approach}
\label{proposed}
Suppose that a skeleton sequence $\boldsymbol{S}=(S_{1},\cdots,S_{f})$ contains $f$ consecutive skeleton frames, where  $S_{i} \in \mathbb{R}^{J \times 3}$ contains 3D coordinates of $J$ body joints. The training set $\Phi=\{\boldsymbol{S}^{(i)}\}_{i=1}^{N}$ contains $N$ skeleton sequences collected from different persons. Each skeleton sequence $\boldsymbol{S}^{(i)}$ corresponds to a label $y_i$, where $ \ y_i \!\in\! \left \{ 1,\cdots, C \right \}$ and $C$ is the number of persons. Our goal is to learn discriminative gait features $\boldsymbol{v}$ from $\boldsymbol{S}$ without using any label. Then, the effectiveness of learned features $\boldsymbol{v}$ can be validated by using them to perform person Re-ID: Learned features and labels are used to train a simple recognition network (note that learned features $\boldsymbol{v}$ are frozen and NOT tuned by training at this stage). The overview of the proposed approach is given in Fig. \ref{model}, and we present details of each technical component below. 

\subsection{Self-Supervised Learning with 3D Skeletons}
\label{reconstruction_mechanism}

\subsubsection{Reverse Reconstruction as Self-Supervision}
\label{sec:rev_rec}
To learn gait representations without labeled 3D skeleton sequences, we propose to introduce self-supervision by learning to reconstruct input 3D skeleton sequences in a \textit{reverse} order, $i.e.$, by taking the input skeleton sequence $\boldsymbol{S}^{\text{I}}\!=\!({S}^{\text{I}}_{1},\cdots,{S}^{\text{I}}_{f})\!=\!({S}_{1},\cdots,{S}_{f})\!=\!\boldsymbol{S}$, we expect our model to output the sequence $\boldsymbol{{\hat S}}\!=\!({\hat S}_{1},\cdots,{\hat S}_{f})\!=\!(S_{f},\cdots,S_{1})$, which gives ${\hat S}_{t}=S_{f-t+1}$. Compared with the na\"ive reconstruction that learns to reconstruct exact inputs ($\boldsymbol{S}^{\text{I}}\!\longmapsto\! \boldsymbol{S}^{\text{I}}$), the proposed learning objective ($\boldsymbol{S}^{\text{I}}\!\longmapsto\! \boldsymbol{{\hat S}}$) is combined with more high-level information (\textit{e.g.}, skeleton order in the sequence) that are meaningful to human perception, which requires the model to capture richer high-level semantics to achieve this learning objective. In this way, our model is expected to learn more meaningful gait representations than frequently-used plain reconstruction. Formally, given an input 3D skeleton sequence, we use the encoder to encode each skeleton frame $S^{\text{I}}_{t}$ ($t\in\{1,\cdots f\}$) and the previous step's latent state $\boldsymbol{h}_{t-1}$ (if existed), which provides the temporal context information for the gait encoding process, into the current latent state $\boldsymbol{h}_{t}$:
\begin{align*}
\boldsymbol{h}_{t}=\left\{
\begin{aligned}
\ & \phi_{E}\left(S^{\text{I}}_{1} \right) \quad & \text{if} \quad t = 1 \\
\ & \phi_{E}\left(\boldsymbol{h}_{t-1}, S^{\text{I}}_{t} \right) \quad & \text{if} \quad t > 1
\end{aligned}
\right. \tag{$1$}\label{encode}
\end{align*}
where $\boldsymbol{h}_{t} \in \mathbb{R}^{K}$, $\phi_{E}(\cdot)$ denotes our Gait Encoder (GE). GE is built with an LSTM, which aims to capture long-term temporal dynamics of skeleton sequences. $\boldsymbol{h}_{1},\cdots,\boldsymbol{h}_{f}$ denote the \textit{encoded gait states} that contain preliminary gait encoding information. In the \textit{training} phase, encoded gait states are decoded by a Gait Decoder (GD) that aims to output the target sequence $\boldsymbol{{\hat S}}$, and the decoding process is performed below (see  Fig. \ref{model}):
\begin{align*}
(\boldsymbol{{\hat h}}_{t}, \overline{S}_{t})=\left\{
\begin{aligned}
\ & \phi_{D}\left(\boldsymbol{h}_{f}, \boldsymbol{x}_{1}\right) \quad & \text{if} \quad t = 1 \\
\ & \phi_{D}\left(\boldsymbol{{\hat h}}_{t-1}, \boldsymbol{x}_{t-1}, \boldsymbol{\overline{h}}_{t-1}\right) \quad & \text{if} \quad t > 1
\end{aligned}
\right.\tag{$2$}\label{decode_train}
\end{align*}
where $\phi_{D}(\cdot)$ denotes the GD. GD consists of an LSTM and a fully connected (FC) layer that outputs those joint coordinates of a 3D skeleton. $ \boldsymbol{{\hat h}}_{t}\!\in \mathbb{R}^{K}$ refers to the $t^{th}$ \textit{decoded gait state}, $i.e.$, the latent state output by GD's LSTM to generate the $t^{th}$ skeleton $\overline{S}_t$. $\boldsymbol{x}_{t}$ is the $t^{th}$ auxiliary input for \textit{training}. When the decoding is initialized ($t=1$), we feed $\boldsymbol{x}_{1}\!=\!\boldsymbol{Z}\in \mathbb{R}^{J}$, which is an all-0 skeleton placeholder, and the final encoded gait state $\boldsymbol{h}_f$ into GD to decode the first skeleton. Afterwards, to generate $t^{th}$ skeleton $\overline{S}_t$, $\phi_{D}(\cdot)$ takes three inputs from the ${(t-1)}^{th}$ decoding step: decoded gait state $\boldsymbol{{\hat h}}_{t-1}$, the auxiliary skeleton input $\boldsymbol{x}_{t-1}\!=\!\hat{S}_{t-1}$ (the ground-truth skeleton of the previous time step) that enables better convergence, and the \textit{attentional state vector} $\boldsymbol{\overline{h}}_{t-1}$ that fuses encoding and decoding information based on the proposed attention mechanism, which will be elaborated in Sec. \ref{sec:attention}. In this way, we define the objective function $\mathcal{L}_{S}$ for self-supervision, which minimizes the mean square errors (MSE) between a target sequence $\boldsymbol{{\hat S}}$ and an output sequence $\boldsymbol{\overline{S}}$: 
\begin{align*}
\mathcal{L}_{S}=\sum_{i=1}^{f}\sum_{j=1}^{J}(\boldsymbol{\overline{S}}_{ij}-\boldsymbol{{\hat S}}_{ij})^{2} \tag{$3$}\label{reconstruct}
\end{align*}
where $\boldsymbol{\overline{S}}_{ij}$, $\boldsymbol{{\hat S}}_{ij}$ represent the $j^{th}$ joint position of the $i^{th}$ output or target skeleton. In the \textit{testing} phase, to test the reconstruction ability of our model, it should be noted that we use the output skeleton $\overline{S}_{t-1}$ rather than the target skeleton $\hat S_{t-1}$ as the auxiliary input to $\phi_{D}$ in the $t>1$ case, $i.e.$,  $\boldsymbol{x}_{t-1}\!=\!\overline{S}_{t-1}$. To facilitate training, our implementation actually optimizes Eq. \ref{reconstruct} on each individual dimension of the skeleton's 3D coordinates: $\boldsymbol{S}^{[d]}{\longmapsto}\boldsymbol{{\hat S}}^{[d]}$, where $d\in\{X, Y, Z\}$ corresponds to a certain dimension of the 3D data space, and $\boldsymbol{S}^{[d]},\boldsymbol{{\hat S}}^{[d]}\in \mathbb{R}^{f \times J}$.

\begin{figure}
    \centering\
    \scalebox{0.44}{
    \includegraphics{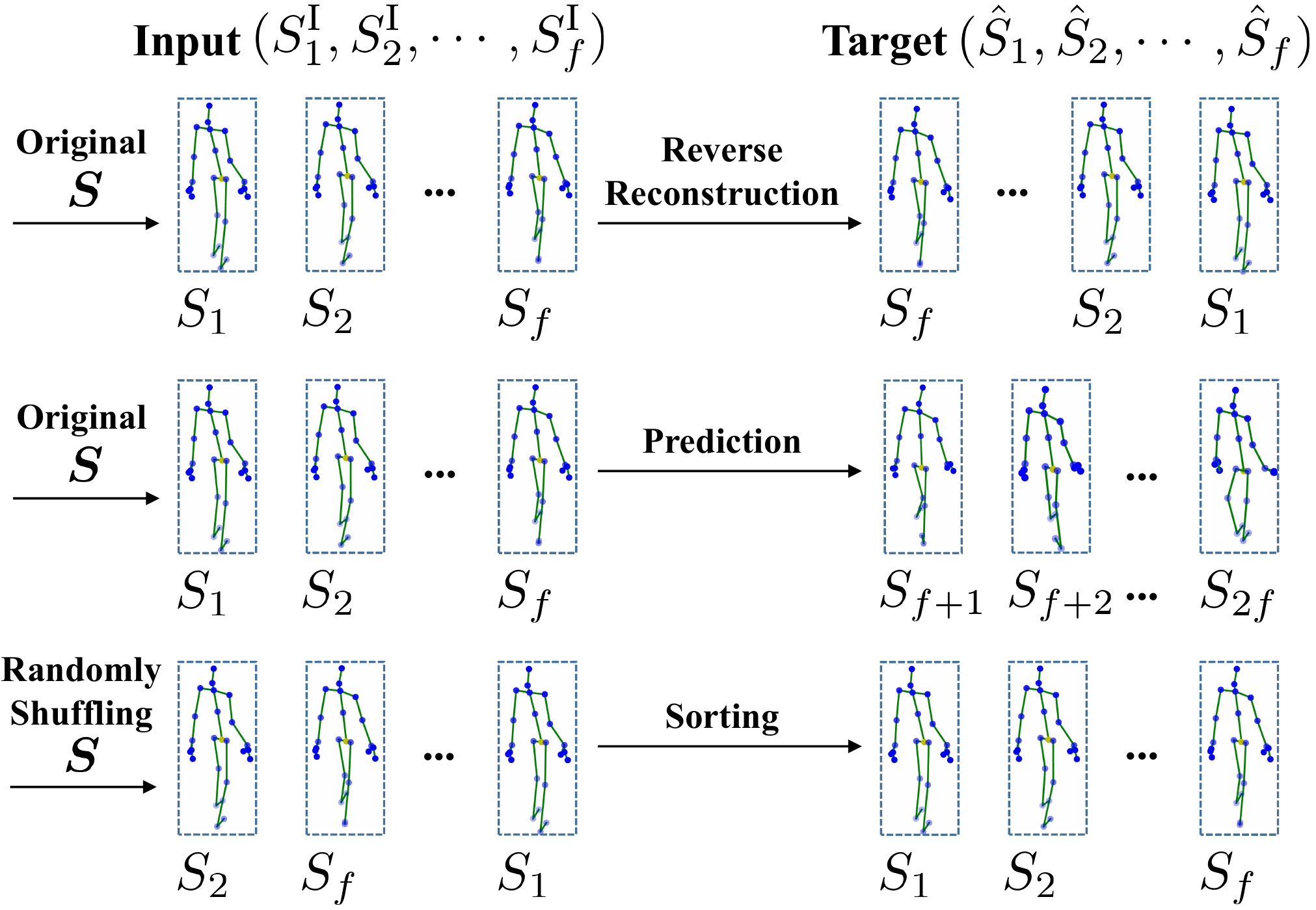}
    }
    \caption{Schematic diagrams of three pretext tasks: Reverse reconstruction (top), prediction (middle), sorting (bottom). The original sequence $\boldsymbol{S}=(S_{1},\cdots,S_{f})$ is the input $(S^{\text{I}}_{1},\cdots,S^{\text{I}}_{f})$ for reverse reconstruction and prediction, and a random shuffle of $\boldsymbol{S}$ is the input for sorting.}
    \label{pretext_diagrams}
\end{figure}
\subsubsection{Other Pretext Tasks for Self-Supervision}
\label{other_pretext_tasks}
Our self-supervised gait encoding approach can also be equipped with other pretext tasks, which exploit different inputs $\boldsymbol{S}^{\text{I}}$ and learning targets $\boldsymbol{\hat S}$ to provide self-supervision for gait encoding of 3D skeleton sequences. 
To this end, we design two additional pretext tasks in this work: (1) Future skeleton frame prediction (``Prediction''). As shown in Fig. \ref{pretext_diagrams}, the prediction task takes the original skeleton sequence as the input, namely $(S^{\text{I}}_{1},\cdots,S^{\text{I}}_{f})=(S_{1},\cdots,S_{f})$, and the learning goal is to predict the next $f$ skeleton frames: $(\hat{S}_{1},\cdots,\hat{S}_{f})=(S_{f+1},\cdots,S_{2f})$. The motivation of this task is that the model must capture key motion patterns in a skeleton sequence to predict unseen future skeletons, and learning to acquire such inference ability enables the model to mine more latent gait features. (2) 
Skeleton sequence sorting (``Sorting'').  The sorting task attempts to sort a randomly shuffled 3D skeleton sequence $\boldsymbol{S}$ 
back to the original sequence. Specifically, 
the input is $(S^{\text{I}}_{1},\cdots,S^{\text{I}}_{f})=(S_{r_1},\cdots,S_{r_f})$ ($r_1, \cdots, r_f \!\in\!\{1, \cdots, f\}$ are shuffled indexes), and the target sequence is $(\hat{S}_{1},\cdots,\hat{S}_{f})=(S_{1},\cdots,S_{f})$. In this way, it enables the model to learn the inherent temporal coherence embedded in skeleton sequences during gait encoding. Besides, it is easy to know that reverse reconstruction is a special case of sorting. As a comparison, sorting is usually more difficult for the model, since a random shuffle often removes the sequence order information completely. By contrast, reverse reconstruction still retains the sequence order information at inputs, which makes it possible to utilize the locality property discussed in Sec. \ref{sec:attention}.  

For those new pretext tasks of self-supervised learning, we can still leverage the same model structure while changing the inputs $\boldsymbol{S}^{\text{I}}$, targets $\boldsymbol{\hat S}$, and auxiliary input $\boldsymbol{x}$ accordingly during \textit{training}: For prediction and sorting, we alternate the auxiliary input from $\boldsymbol{x}_{t}=\hat{S}_{t}$ (the ground-truth skeleton) to $\boldsymbol{x}_{t}=\overline{S}_{t}$ (the predicted skeleton) in the $t>1$ case, and in the testing phase we keep the auxiliary inputs unchanged. For prediction, we also test the half-prediction case ($i.e.$, target sequence is $\boldsymbol{\hat{S}}=(S_{\frac{f}{2}+1},\cdots,S_{\frac{3f}{2}})$), which is shown to yield better gait representations for person Re-ID than the full-prediction (see supplementary material).
Our later experiments compare the gait features learned by different pretext tasks for person Re-ID, and the results demonstrate that the proposed reverse reconstruction achieves the best performance (see Sec. \ref{performance_comp}), which explains the center role of reverse reconstruction in the proposed self-supervised gait encoding approach. However, our experiments also show that gait features learned by other pretext tasks can be readily combined with features learned by reverse reconstruction (referred as the ``Rev. Rec. '' configuration in later experiments) to further improve Re-ID performance. 
\hc{First, we extensively evaluate different combinations of two pretext tasks:  ``Rev. Rec. + Pred. '' (reverse reconstruction+prediction),  ``Rev. Rec. + Sort. '' (reverse reconstruction+sorting),``Pred. + Sort. '' (prediction+sorting) in Sec. \ref{results_of_pretexts}.} Then, we propose the ``Rev. Rec. Plus'' configuration that synthesizes gait features learned from all three pretext tasks for person Re-ID. Consequently, the exploration of more specific pretext tasks will be beneficial to our self-supervised skeleton sequence learning paradigm.


\subsection{Locality-Aware Attention Mechanism}
\label{sec:attention}

As learning gait features essentially requires capturing motion patterns from 3D skeleton sequences, it is instinctive to consider a natural property of motion--continuity. The continuity of motion ensures that those skeletons in a small temporal interval will NOT undergo drastic changes, thus resulting in higher correlations among adjacent skeletons in a local context of the skeleton sequence. This property is referred as \textit{\textbf{intra-sequence locality}} here. Due to such intra-sequence locality, when reconstructing a certain skeleton in a sequence, we expect our model to pay more attention to its neighboring skeletons that are located in the same local temporal context. 
To this end, we propose a locality-aware attention mechanism, the details of which are presented below.

\subsubsection{Basic Attention Alignment} We first introduce Basic Attention (BA) alignment \cite{luong-etal-2015-effective} to measure the content ($i.e.$, latent state) based correlations between the input sequence and the output sequence. As shown in Fig. \ref{model}, at the $t^{th}$ decoding step, we compute the \textit{BA Alignment Scores (BAS)} between $\boldsymbol{{\hat h}}_{t}$ and the encoded gait state $\boldsymbol{h}_{j}$ ($j\in\{1,\dots,f\}$):
\begin{align*}
     {a}_{t}(j)&=\operatorname{align}\left(\boldsymbol{{\hat h}}_{t},\boldsymbol{h}_{j}\right)=\frac{\exp \left({\boldsymbol{{\hat h}}_{t}}^{\top}\boldsymbol{h}_{j}\right)}{\sum^{f}_{i=1} \exp \left({\boldsymbol{{\hat h}}_{t}}^{\top}\boldsymbol{h}_{i}\right)}
    \tag{$4$}\label{align}
 \end{align*}
 BAS aims to focus on those more correlative skeletons in the encoding stage, and provides preliminary attention weights for skeleton decoding. However, BA alignment only considers the content based correlations and does not explicitly take the intra-sequence locality into consideration, which motivates us to design the locality mask and locality-aware attention alignment below.  

 \begin{figure}[t]
    \centering
    \subfigure[BAS Attention Matrix]{\scalebox{0.34}{\includegraphics[]{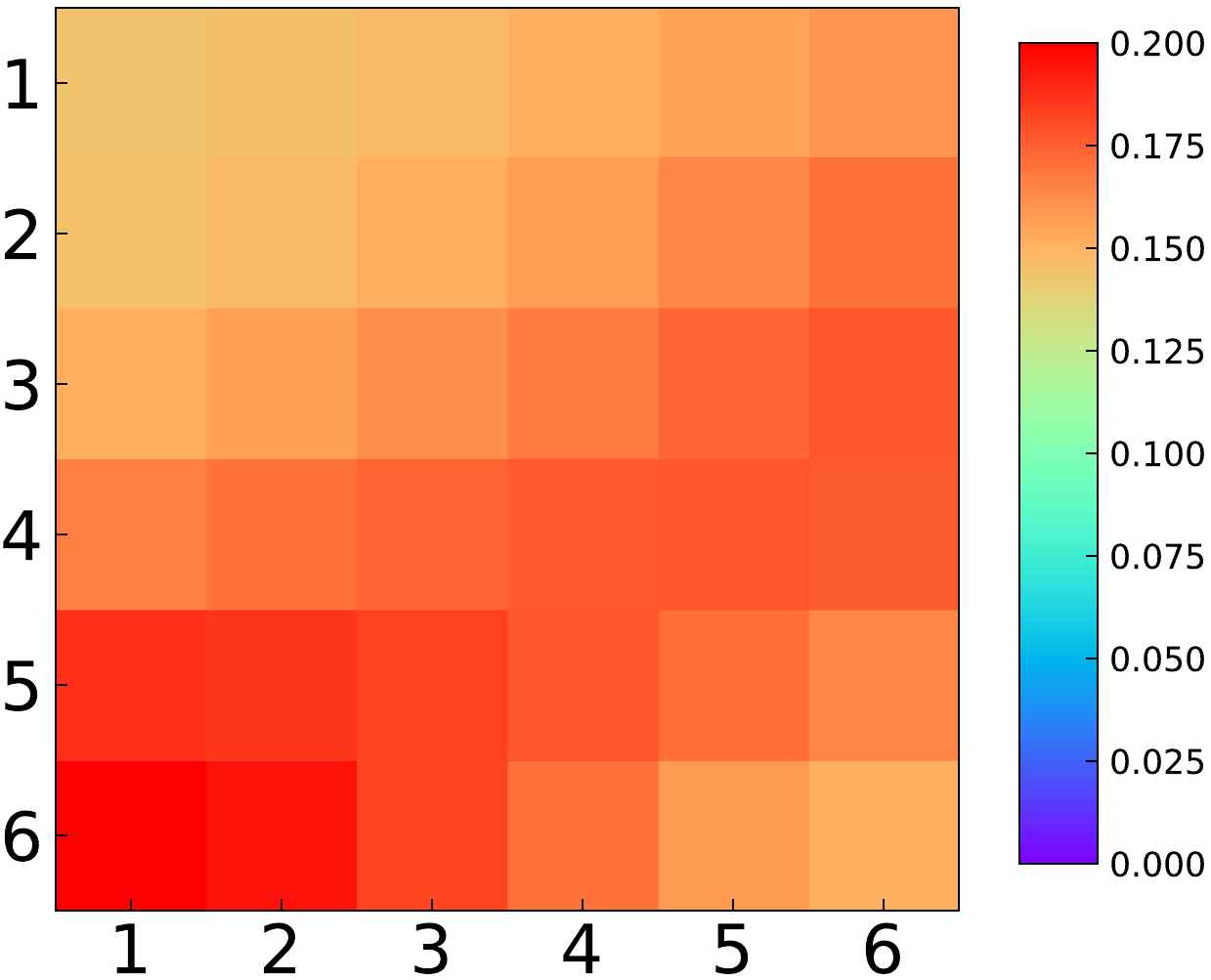}}
     }\
     \subfigure[LAS Attention Matrix]{\scalebox{0.34}{\includegraphics[]{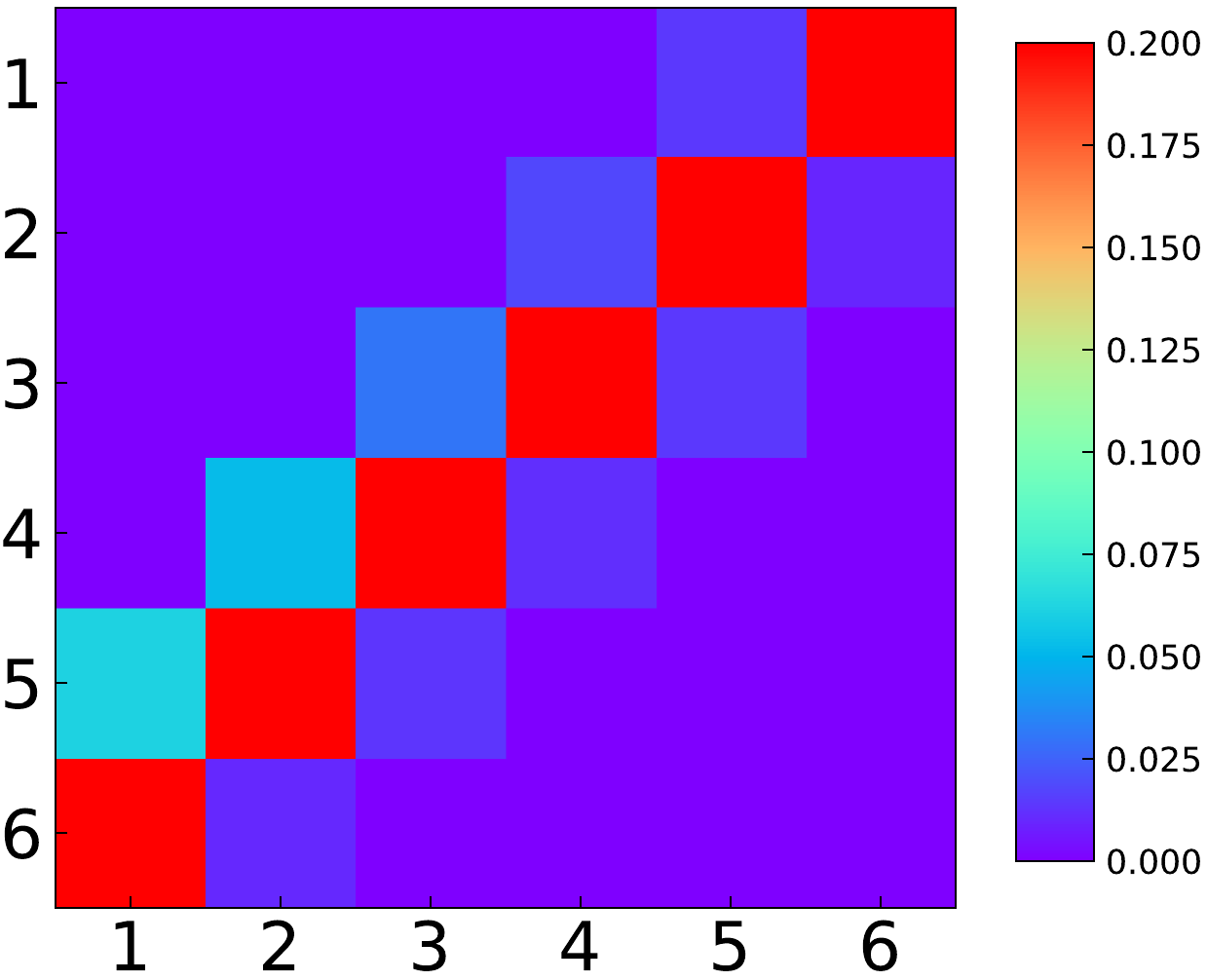}}
     }
    \caption{Visualization of the BAS (left) and LAS (right) attention matrices that represent average attention alignment scores. Note that the abscissa and ordinate denote indices of input skeletons and output skeletons respectively. The LA alignment improves the learning of locality by assigning larger alignment scores near the clinodiagonal line.} 
    \label{attention_comparison}
\end{figure}

\subsubsection{Locality Mask} The motivation to design locality mask is to incorporate intra-sequence locality directly into the gait encoding process for better skeleton reconstruction. As the goal is to decode the $t^{th}$ skeleton $\overline{S}_t$ as $\hat S_t$, we consider those skeletons in the local temporal context of $S_{p_{t}}$ to be highly correlated to $\overline{S}_t$, where $p_{t}=f-t+1$ (note that we use the reverse reconstruction). To describe the local context centered at $S_{p_{t}}$, we define an attentional window $[p_{t}-D, p_{t}+D]$, where $D$ is a selected integer to control the attentional range. Since the locality will favor temporal positions near $p_{t}$ (\textit{i.e.}, closer positions are more correlative), a direct solution is to place a \textit{Gaussian} distribution centered around $p_{t}$ as the locality mask:
\begin{align*}
{l}_{t}(j)&=\mathrm{exp}\left ( -\frac{(j-p_{t})^2}{2\sigma^{2}} \right )
\tag{$5$}\label{local_align}
\end{align*}
where we empirically set $\sigma \!= \!\frac{D} {2}$, $j$ is a position within the window centered at $p_{t}$. We can weight BAS by this locality mask to compute \textit{Masked BA Alignment Scores (MBAS)} below, which directly forces alignment scores to obtain locality:
\begin{align*}
\tilde{a}_{t}(j)={l}_{t}(j)\cdot{a}_{t}(j)
\tag{$6$}\label{local_align_scores}
\end{align*}
Besides, the locality mask is only valid for sequential reconstruction, so it cannot be directly combined with sorting or prediction. This is exactly an advantage of the reverse reconstruction task. 

\subsubsection{Locality-Aware Attention Alignment} Despite that the locality mask is straightforward to yield the intra-sequence locality, it is a very coarse solution that brutally constrains the alignment scores. Therefore, instead of using MBAS (${\tilde{a}}_{t}(j)$) directly, we propose the \textit{Locality-aware Attention (LA) alignment}. Specifically, an LA alignment loss term $\mathcal{L}_A$ is used to encourage LA alignment to learn similar locality like ${\tilde{a}}_{t}(j)$:
\begin{align*}
\mathcal{L}_{A}=\sum_{t=1}^{f}\sum_{j=1}^{f}({a}_{t}(j)-\tilde{a}_{t}(j))^{2} \tag{$7$}\label{LA_loss}
\end{align*}
 By adding the loss term $\mathcal{L}_{A}$, we can obtain \textit{LA Alignment Scores (LAS)}. Note that in Eq. \ref{align}, the final learned ${a}_{t}(j)$ is BAS. For clarity, we use $\overline{a}_{t}(j)$ to represent LAS learned by Eq. (\ref{LA_loss}). With the guidance of $\mathcal{L}_{A}$, our model learns to allocate more attention to the local temporal context by itself rather than using a hard locality mask. To utilize alignment scores to yield an attention-weighted encoded gait state at the $t^{th}$ step, we can calculate the \textit{context vector} $\boldsymbol{c}_t$ by a sum of weighted encoded gait states:
\begin{align*}
\boldsymbol{c}_{t}=\sum^{f}_{j=1} \overline{a}_{t}(j)\boldsymbol{h}_{j} \tag{$8$}\label{context}
\end{align*}
Note that the context vector $\boldsymbol{c}_{t}$ can also be computed with BAS or MBAS. $\boldsymbol{c}_{t}$ provides a synthesized gait encoding that is more relevant to $\boldsymbol{\hat{h}}_{t}$, which facilitates the reconstruction of $t^{th}$ skeleton.  To combine both encoding and decoding information for reverse reconstruction, we use a concatenation layer $f_{att}(\cdot)$ that combines $\boldsymbol{c}_{t}$ and $\boldsymbol{{\hat h}}_{t}$ into an attentional state vector $\boldsymbol{\overline{h}}_{t}$: 
\begin{align*}
     \boldsymbol{\overline{h}}_{t}&=f_{att}(\boldsymbol{c}_{t}, \boldsymbol{{\hat h}}_{t})=\tanh (\boldsymbol{W}_{att}[\boldsymbol{c}_{t};\boldsymbol{{\hat h}}_{t}]) \tag{$9$}\label{attention_state}
\end{align*}
where $\boldsymbol{W}_{att}$ represents the learnable weight matrix in the layer. 
Finally, we generate the joint coordinates of $t^{th}$ output skeleton by the FC layer $f_F(\cdot)$ of the GD:
\begin{align*}
         \overline{S}_{t}=f_{F}\left(\boldsymbol{\overline{h}}_{t}\right)=\boldsymbol{W}_{F}\boldsymbol{\overline{h}}_{t} \tag{$10$}\label{prediction}
\end{align*}
where $\boldsymbol{W}_{F}$ is the weights to be learned in this FC layer.

\begin{figure}
    \centering
    \scalebox{0.39}{\includegraphics[]{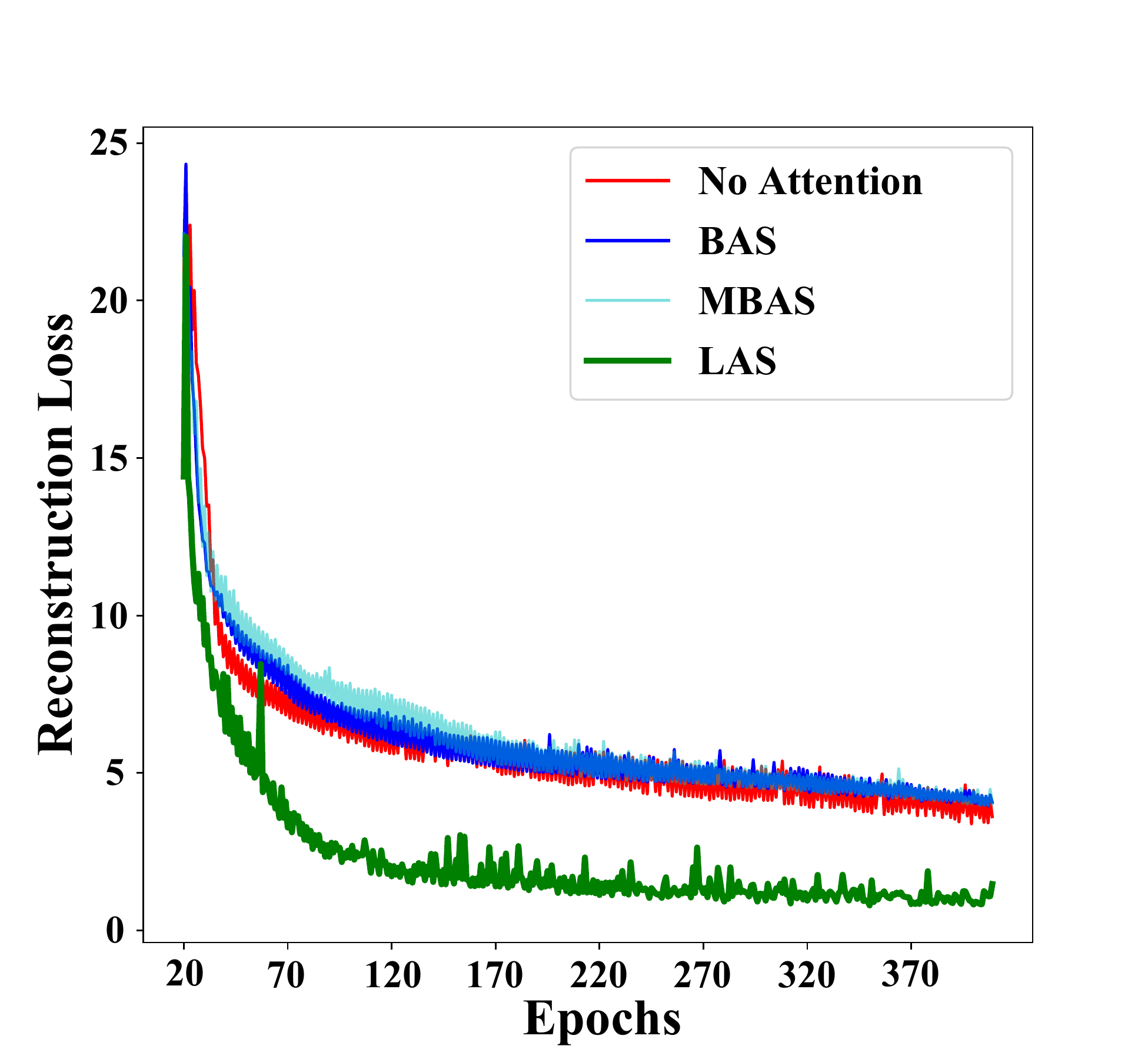}
    }
    \caption{Reconstruction loss curves when using no attention, BAS, MBAS or LAS for skeleton reconstruction (note that here we compare different attention mechanisms without using contrastive learning). Using LAS achieves better reverse reconstruction with smaller reconstruction loss.}
    \label{attention_recon_loss}
\end{figure}

\subsubsection{Analysis on Different Attention Mechanisms}
First, to provide a more intuitive impression of the proposed locality-aware attention mechanism, we visualize the BAS and LAS attention matrices, which are formed by the average alignment scores computed with the $f=6$ skeleton sequence on dimension $X$ (explained in Sec. \ref{sec:rev_rec}) as an example. As shown by Fig. \ref{attention_comparison}, LA alignment significantly improves the locality of learned alignment scores: It can be observed that relatively large alignment scores are densely distributed near the clinodiagonal line of the attention matrix (note that when reverse reconstruction is performed, the clinodiagonal line reflects each skeletons' correlations to themselves), which means temporally adjacent skeletons are assigned with larger attention than those comparatively remote skeletons. By contrast, despite that BA alignment also learns locality to a certain extent, BA's alignment weights exhibit a much more uniform distribution and many non-adjacent skeletons are also given large alignment scores. Similar trends are also observed when learning on dimension $Y$ and $Z$. Such observations justify LAS's effectiveness to encourage intra-sequence locality. 

Second, to illustrate how different attention mechanisms contribute to the self-supervised learning goal, \textit{i.e.} the reverse reconstruction of 3D skeleton sequences, we visualize the corresponding reconstruction loss during training in four cases: No attention mechanism, BAS, MBAS and LAS. As shown by Fig. \ref{attention_recon_loss}, it can be observed that training with LAS converges at a faster speed with a smaller reconstruction loss, which justifies our intuition that exploiting locality will facilitate the reverse reconstruction. Interestingly, we observe that using the locality mask directly in fact does not benefit the reduction of reconstruction loss, which also indicates that learning is a better way to accomplish intra-sequence locality than imposing a hard locality mask.

\subsubsection{Attention-based Gait Encodings}
\label{GAEV_construction}
Instead of simply fulfilling the pretext tasks, the ultimate goal here is to learn good gait features to conduct effective person Re-ID. Thus, we need to extract certain 3D skeleton sequence embeddings from the internal layers of neural networks to construct gait representations. Unlike traditional LSTM based methods that basically rely on the last hidden state to compress the temporal dynamics of a sequence \cite{weston2014memory}, we recall that the dynamic context vector $\mathbf{c}_t$ 
learned from the attention mechanism integrates the key encoded gait states of input skeletons and retains crucial spatio-temporal information to recover target skeleton sequences. Hence, we utilize them instead of the last hidden state to build the preliminary gait representations--Attention-based Gait Encodings (AGEs). Specifically, skeleton-level AGE ($\boldsymbol{v}_t$) is defined as follows:
\begin{align*}
\boldsymbol{v}_{t}=[\boldsymbol{c}^{X}_{t};\boldsymbol{c}^{Y}_{t};\boldsymbol{c}^{Z}_{t}]
\tag{$11$}\label{eq9}
\end{align*}
where $\boldsymbol{c}^{d}_{t}$ denotes the context vector computed on dimension $d\in\{X, Y, Z\}$ at the $t^{th}$ step of decoding. As reported in our earlier work \cite{DBLP:conf/ijcai/RaoW0TD0020}, AGEs can be directly utilized to perform person Re-ID. However, AGEs only incorporate the locality on the intra-sequence level, \textit{i.e.}, among different skeletons in one 3D skeleton sequence. They do not consider the relationship between different 3D skeleton sequences on the inter-sequence level, which can also be involved to improve the gait representation learning. This motivates us to propose the Locality-aware Contrastive Learning (LCL) scheme below, so as to further encourage the inter-sequence locality among different skeleton sequences. Besides, since LCL is still performed on each individual dimension and AGEs will be further tuned by LCL, we use a slightly abused notation by defining $\boldsymbol{v}_{t}=\boldsymbol{c}^{d}_{t}$ in the next section, where $d\in\{X, Y, Z\}$.

\begin{figure}[tb]
    \centering
     \scalebox{0.375}{\includegraphics[]{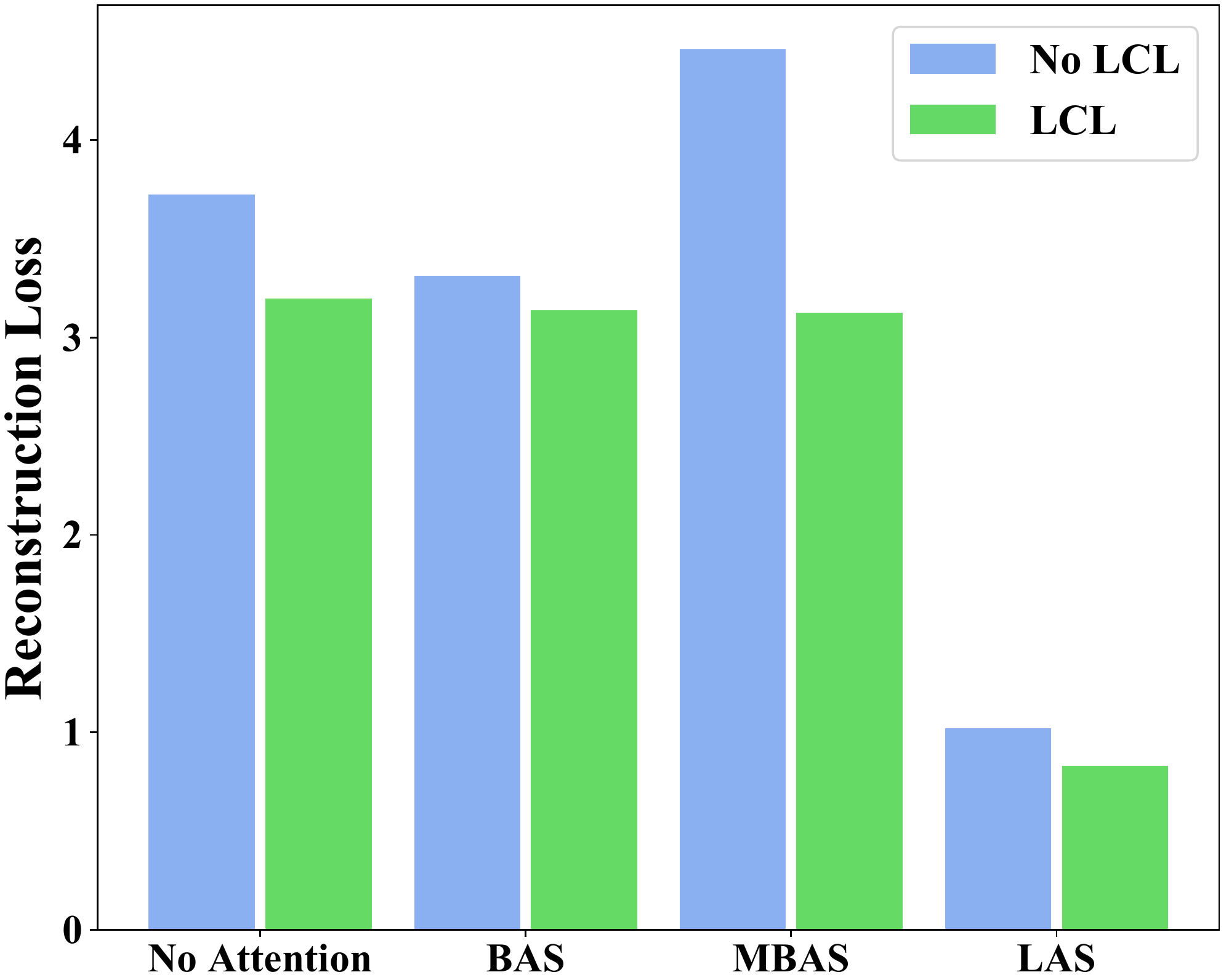}}
    \caption{Skeleton reconstruction loss when using no attention, BAS, MBAS or LAS for skeleton reconstruction. The comparison between applying LCL (``LCL'') and not applying LCL (``No LCL'') is reported.}
    \label{LCL_loss}
\end{figure}

\subsection{Locality-Aware Contrastive Learning Scheme}
\label{LCL_mechanism}
Similar to intra-sequence locality among skeletons in a sequence, we can also assume that consecutive skeleton sequences in a local temporal context are more likely to share similar gait representations than those non-consecutive ones. Such relationship among different 3D skeleton sequences (referred as \textbf{\textit{inter-sequence locality}}) can be exploited to enhance self-supervised gait encoding.
 To this end, we propose a Locality-aware Contrastive Learning (LCL) scheme to impose inter-sequence locality on skeleton sequences.

\subsubsection{Skeleton Sequence Contrastive Learning}
To compute the correlations between skeleton sequences and learn the inter-sequence locality, we first construct sequence-level gait representations by concatenating skeleton-level AGEs as follows:
\begin{align*}
\boldsymbol{V}^{(i)}=[\boldsymbol{v}_{1};\boldsymbol{v}_{2};\dots;\boldsymbol{v}_{t}]
\tag{$12$}\label{AGE_seq}
\end{align*}
where $\boldsymbol{V}^{(i)}$ denotes attention-based gait encodings of the $i^{th}$ skeleton sequence $\boldsymbol{S}^{(i)}$ in the training set $\Phi$. Here we adopt the same setting in \cite{chen2020a} to improve  representation learning: We first use a multi-layer perceptron (MLP) with one hidden layer to map $\boldsymbol{V}^{(i)}$ to the contrasting space where the contrastive loss is applied:
\begin{align*}
\boldsymbol{z}_{i}=f_{C}\left(\boldsymbol{V}^{(i)}\right)=\boldsymbol{W}^{2} \sigma\left(\boldsymbol{W}^{1} \boldsymbol{V}^{(i)}\right)
\tag{$13$}\label{MLP_map}
\end{align*}
where $\boldsymbol{z}_{i}\in\mathbb{R}^{K}$ is the representation of $i^{th}$ skeleton sequence in the contrasting space. $f_{C}(\cdot)$ is the function that denotes the MLP layer for contrastive learning. $\sigma$ is the non-linear activation function like ReLU. $\boldsymbol{W}^{1}$ and $\boldsymbol{W}^{2}$ are weights to be learned in the MLP layer. The LCL scheme contrasts the similarity between representations ($\boldsymbol{z}_{i}$) of different skeleton sequences. During the training stage of LCL scheme, each batch of skeleton sequences $\{\boldsymbol{S}^{{(k)}}\}^{n}_{k=1}$ is drawn without random shuffle from the training subset $\Phi_{i} \ (i \in \{1,\cdots,C\})$ that corresponds to the $i^{th}$ person, and $\Phi=\{\Phi_{1},\cdots\Phi_{C}\}$. Next, we define two consecutive skeleton sequences as a positive pair, while two non-consecutive skeleton sequences in the batch are defined as a negative pair.
Given a positive skeleton sequence pair $\boldsymbol{S}^{(i)}$ and $\boldsymbol{S}^{(j)}$ from $\{\boldsymbol{S}^{{(k)}}\}^{n}_{k=1}$, the LCL scheme aims to maximize the agreement between representations of $\boldsymbol{S}^{(i)}$ and $\boldsymbol{S}^{(j)}$. We define the loss function below for a positive sequence pair and summarize the entire LCL scheme in Algorithm \ref{LCL_algorithm}:
\begin{align*}
\ell(i, j)=-\log \frac{\exp \left(\operatorname{sim}\left(\boldsymbol{z}_{i}, \boldsymbol{z}_{j}\right) / \tau\right)}{\sum_{k=1}^{2 n-2} \mathds{1}_{[k \neq i]} \exp \left(\operatorname{sim}\left(\boldsymbol{z}_{i}, \boldsymbol{z}_{k}\right) / \tau\right)}
\tag{$14$}\label{c_loss}
\end{align*}
where $\operatorname{sim}(\boldsymbol{z}_{i}, \boldsymbol{z}_{j})=\boldsymbol{z}_{i}^{\top} \boldsymbol{z}_{j} /\|\boldsymbol{z}_{i}\|\|\boldsymbol{z}_{j}\|$ denotes the cosine similarity between two representation vectors $\boldsymbol{z}_{i}$ and $\boldsymbol{z}_{j}$, $\tau$ denotes the temperature parameter, and $\mathds{1}_{[k \neq i]}\in \{0,1\}$ is an indicator function: $\mathds{1}_{[k \neq i]}=1$ iff $k \neq i$. $2n-2$ is the number of samples to contrast in a training batch.
As presented in Algorithm \ref{LCL_algorithm}, we contrast every two sequences in a batch of size $n$ (note that the number of positive pairs is $n-1$), and the final contrastive loss $\mathcal{L}_{C}$ is computed among all positive pairs.

\begin{algorithm}[t]
    \setstretch{1.35}
    \caption{Main algorithm of LCL scheme}
    \label{LCL_algorithm}
    \begin{algorithmic}
    \REQUIRE Batch size $n$, temperature $\tau$, gait encoding model $\phi$, MLP function $f_{C}$ for contrastive learning.
    \FOR{a batch of consecutive sequences $\{\boldsymbol{S}^{{(k)}}\}^{n}_{k=1}$}
    \FORALL{$k\in\{1, \dots, n-1\}$}
    \STATE$\boldsymbol{V}^{(k)}=\phi(\boldsymbol{S}^{(k)})$ \COMMENT{gait representation} 
    \STATE$\boldsymbol{z}_{k}=f_{C}(\boldsymbol{V}^{(k)})$ 
    \COMMENT{map to contrasting space} 
     \STATE$\boldsymbol{V}^{(k+1)}=\phi(\boldsymbol{S}^{(k+1)})$
     \COMMENT{$\boldsymbol{S}^{(k+1)}$ and $\boldsymbol{S}^{(k)}$ are adjacent} 
    \STATE $\boldsymbol{z}_{k+n-1}=f_{C}(\boldsymbol{V}^{(k+1)})$ 
    \ENDFOR 
    \FORALL{$i\in\{1,\dots, 2n-2\}$ and $j\in\{1,\dots, 2n-2\}$} 
    \STATE $\alpha_{i, j}=\frac{\boldsymbol{z}_{i}^{\top} \boldsymbol{z}_{j}}{\tau\left\|\boldsymbol{z}_{i}\right\|\left\|\boldsymbol{z}_{j}\right\|}$ \COMMENT{similarity between sequences}
    \ENDFOR \\
    \textbf{define} $\ell(i, j)=-\log \frac{\exp \left(\alpha_{i, j}\right)}{\sum_{k=1}^{2n-2}\mathds{1}_{[k \neq i]} \exp \left(\alpha_{i, k}\right)}$ \\
    $\mathcal{L}_{C}=\frac{1}{2 n - 2} \sum_{k=1}^{n-1}[\ell(k,\ k\!+\!n\!-\!1)+\ell(k\!+\!n\!-\!1,\ k)]$ \\
    update $\phi$ and $f_{C}$ to minimize $\mathcal{L}_{C}$
    \ENDFOR
    \RETURN{gait encoding model $\phi$ and $f_{C}$}
    \end{algorithmic}
\end{algorithm}
\subsubsection{Analysis on Locality-Aware Contrastive Learning}
The proposed LCL aims to improve the learned gait representations AGEs. It is noted that AGEs are constructed with dynamic context vectors of the model, and such context vectors actually play a important role in carrying out the reverse reconstruction of skeleton sequences. Hence, we can also visualize the final reconstruction loss before and after the LCL scheme is applied to improve the learned context vectors, so as to check whether it can improve context vectors and enable better reverse reconstruction. We compare two cases where LCL is applied and not applied (``No LCL''). As shown in Fig. \ref{LCL_loss}, it is found that the LCL constantly enables the model to achieve lower reconstruction loss, regardless of the used attention mechanism (no attention mechanism, BAS, MBAS, LAS). These results indicate that the gait encoding model with the LCL scheme can achieve better skeleton reconstruction, and we will empirically demonstrate that gait features learned by LCL can boost the person Re-ID performance as well in Sec. \ref{ablation_sec}. 


\subsubsection{Contrastive Attention-based Gait Encodings}
By applying the LCL scheme to sequence-level AGEs (Eq. \ref{AGE_seq}), we can incorporate the inter-sequence locality and tune AGEs into the final gait representations named Contrastive Attention-based Gait Encodings (CAGEs), which preserve locality by both locality-aware attention mechanism and LCL scheme:
\begin{align*}
\boldsymbol{S}^{(i)}\stackrel{\text{\uppercase\expandafter{\romannumeral1}}}{\longmapsto}\text{AGEs}(\boldsymbol{V}^{(i)})\stackrel{\text{\uppercase\expandafter{\romannumeral2}}}{\longmapsto}\text{CAGEs}(\boldsymbol{\overline{V}}^{(i)}) 
\tag{$15$}\label{CAGEs_v}
\end{align*}
where $\boldsymbol{\overline{V}}^{(i)}$ denotes CAGEs of the $i^{th}$ skeleton sequence $\boldsymbol{S}^{(i)}$, \uppercase\expandafter{\romannumeral1} and \uppercase\expandafter{\romannumeral2} are the learning process of locality-aware attention mechanism and LCL scheme respectively. In this work, our model performs \uppercase\expandafter{\romannumeral1} and \uppercase\expandafter{\romannumeral2} simultaneously at each training step, while we conduct only \uppercase\expandafter{\romannumeral1} in \cite{DBLP:conf/ijcai/RaoW0TD0020}.
Note that CAGEs are also built by dynamic context vectors like AGEs in Eq. \ref{eq9}. For simplicity, we use $\boldsymbol{\overline{V}}$ to represent sequence-level CAGEs, and use $\boldsymbol{\overline{v}}_{t}$ to represent the skeleton-level CAGE at the $t^{th}$ step of decoding here.

To perform person Re-ID, we use CAGEs to train a simple recognition network $f_{RN}(\cdot)$ that consists of a hidden layer and a softmax layer. In particular, we explore two specific Re-ID strategies: \textbf{(1)} \textbf{S}equence-level \textbf{C}oncatenation (SC): It directly uses sequence-level CAGEs $\boldsymbol{\overline{V}}$, which is the concatenation of skeleton-level CAGEs ($[\boldsymbol{\overline{v}}_{1};\boldsymbol{\overline{v}}_{2};\dots;\boldsymbol{\overline{v}}_{f}])$, to train the recognition network and predict the sequence label $f_{RN}(\boldsymbol{\overline{V}};\boldsymbol{\theta_{r}})$, where $\boldsymbol{\theta_{r}}$ refers to parameters of $f_{RN}(\cdot)$. \textbf{(2)} \textbf{A}verage \textbf{P}rediction (AP): It exploits skeleton-level CAGEs to train the recognition network, and averages the prediction of each skeleton-level CAGE $f_{RN}(\boldsymbol{\overline{v}}_{t};\boldsymbol{\theta_{r}})$ ($t\in\{1,\cdots,f\}$) in a skeleton sequence to be the final sequence-level prediction for person Re-ID. We compare AP and SC under different pretext tasks and demonstrate that AP constantly achieves better Re-ID performance than SC (see Sec. \ref{results_of_pretexts}). Note that during training, each skeleton in one sequence shares the same skeleton sequence label $y_i$. Besides, skeleton labels are only used to train the recognition network, $i.e.$, CAGEs are frozen during training. Our later evaluations show that CAGEs, which are learned with unlabeled 3D skeleton data only, are surprisingly discriminative and produce remarkable person Re-ID performance.  


\begin{table}[t]
\label{dataset_detail}
\caption{Statistics of different datasets. Note: Here we present the number of skeletons estimated from the original CASIA B dataset.}
\begin{center}
\scalebox{0.82}{
\setlength{\tabcolsep}{1.5mm}{
\begin{tabular}{lrrrrr}
\specialrule{0.1em}{0.45pt}{0.45pt}
\textbf{}                     & \multicolumn{1}{c}{\textbf{KGBD}} & \multicolumn{1}{c}{\textbf{BIWI}} & \multicolumn{1}{c}{\textbf{KS20}} & \multicolumn{1}{c}{\textbf{IAS-Lab}} & \multicolumn{1}{c}{\textbf{CASIA B}} \\ \specialrule{0.1em}{0.45pt}{0.45pt}
\textbf{\# Unique Subjects}   & \textbf{164}                      & 50                                & 20                                & 11                                   & 124                                  \\
\textbf{\# Original Sequences}   & 822                               & 50                                & 300                               & 11                                   & \textbf{13639}                       \\
\textbf{\# Sequences/Classes} & 5                                 & 1                                 & 15                                & 1                                    & \textbf{110}                         \\
\textbf{\# Skeletons/Classes} & 2898                              & 369                               & 537                               & 690                                  & \textbf{8872}                        \\ \specialrule{0.1em}{0.2pt}{0.2pt}
\end{tabular}
}
}
\end{center}
\end{table}

\subsection{The Entire Approach}
\label{optimization}
As a summary, the computation flow of the entire approach during self-supervised learning is $\boldsymbol{h}\rightarrow\boldsymbol{\hat {h}}\rightarrow \overline{a}\rightarrow \boldsymbol{c}\rightarrow \boldsymbol{\overline{h}}\rightarrow \boldsymbol{\overline{S}}$. To guide model training in the gait encoding process, we combine the loss for self-supervision $\mathcal{L}_{S}$ (Eq. \ref{reconstruct}), LA alignment loss $\mathcal{L}_{A}$  (Eq. \ref{LA_loss}), and contrastive loss $\mathcal{L}_{C}$ (Algorithm \ref{LCL_algorithm}) as follows:
\begin{align*}
&\mathcal{L}=\lambda_{S} \mathcal{L}_{S}+\lambda_{A}\mathcal{L}_{A}+\lambda_{C}\mathcal{L}_{C}+\beta \left\|\boldsymbol{\Theta}\right\|^{2}_{2} \tag{$16$}\label{eq11}
\end{align*}
where $\boldsymbol{\Theta}$ denotes the parameters of the model,  $\lambda_{S}$, $\lambda_{A}$, $\lambda_{C}$ are weight coefficients to trade off the importance of the loss for self-supervision, LA alignment loss and contrastive loss. $\left\|\boldsymbol{\Theta}\right\|^{2}_{2}$ is $L_{2}$ regularization.
For the person Re-ID task, we employ standard cross-entropy loss to train the recognition network with CAGEs. 

\begin{figure}[t]
    \centering
      \scalebox{0.075}{\includegraphics{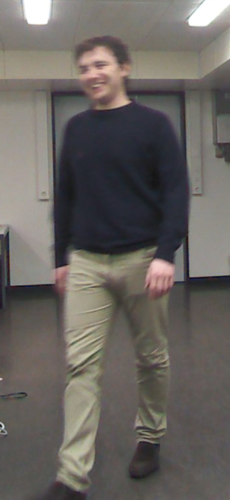}}
       \scalebox{0.075}{\includegraphics{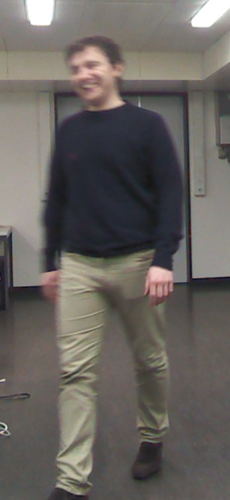}}
        \scalebox{0.075}{\includegraphics{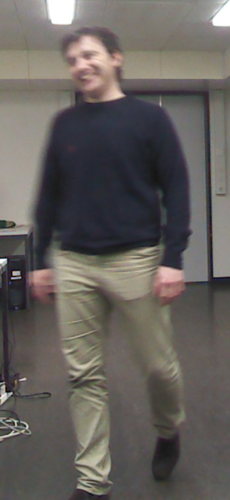}}
         \scalebox{0.075}{\includegraphics{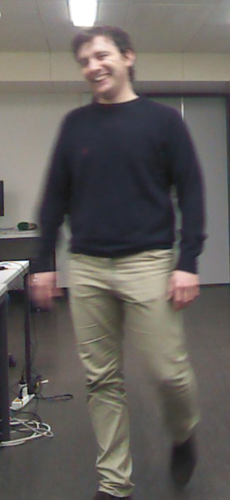}}
          \scalebox{0.075}{\includegraphics{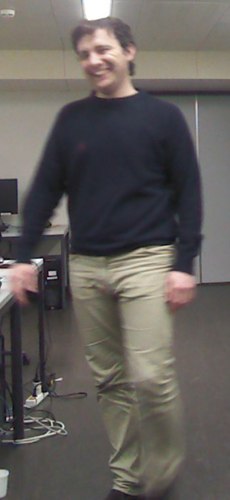}}
           \scalebox{0.075}{\includegraphics{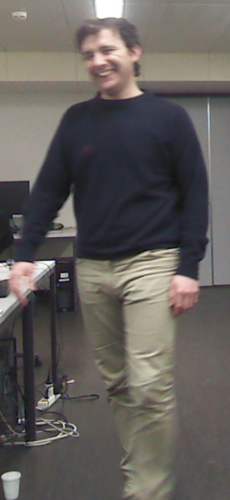}}      \quad \quad\quad
            \scalebox{0.115}{\includegraphics{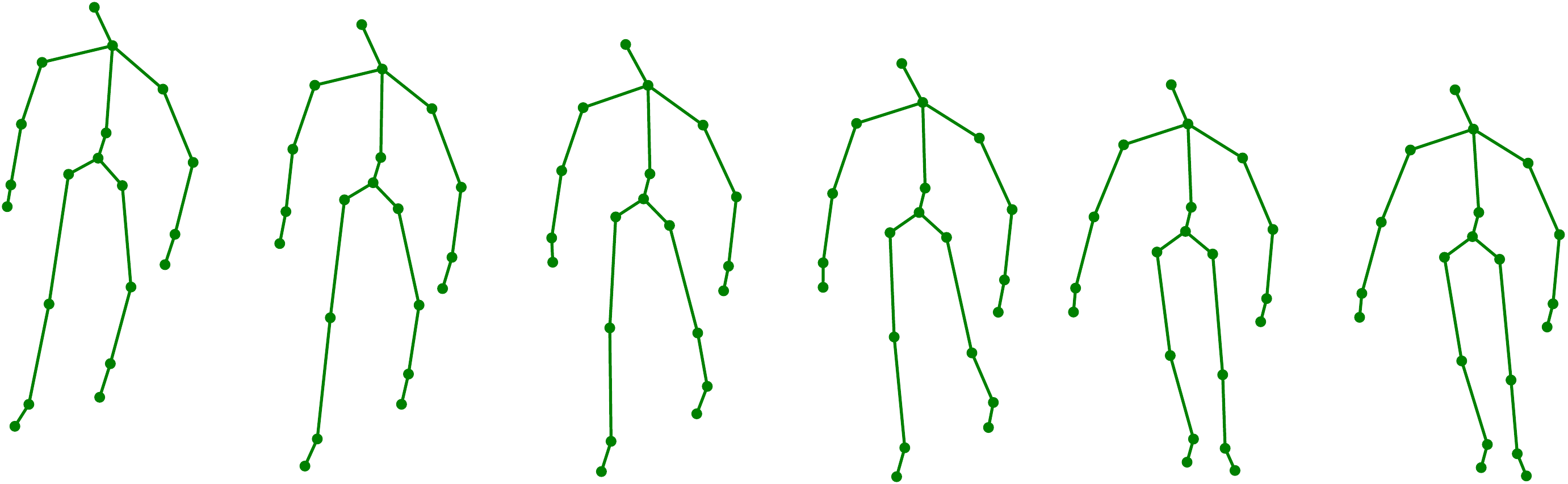}}
     \quad \quad \quad  
     \scalebox{0.10}{\includegraphics{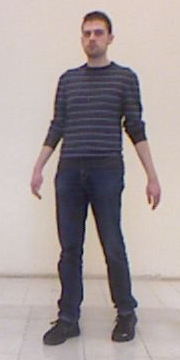}}
       \scalebox{0.10}{\includegraphics{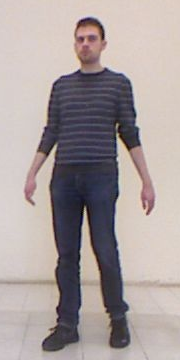}}
        \scalebox{0.10}{\includegraphics{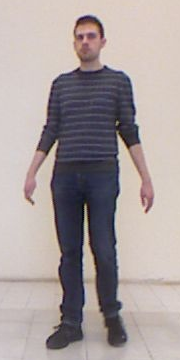}}
         \scalebox{0.10}{\includegraphics{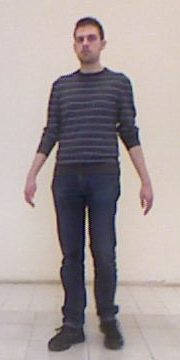}}
          \scalebox{0.10}{\includegraphics{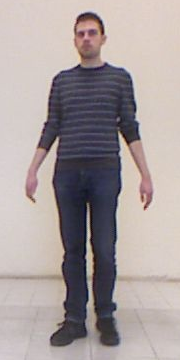}}
           \scalebox{0.10}{\includegraphics{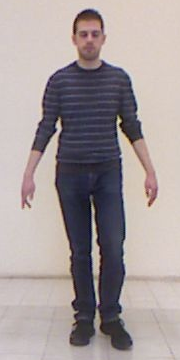}}    \quad \quad  \ \ \scalebox{0.165}{\includegraphics{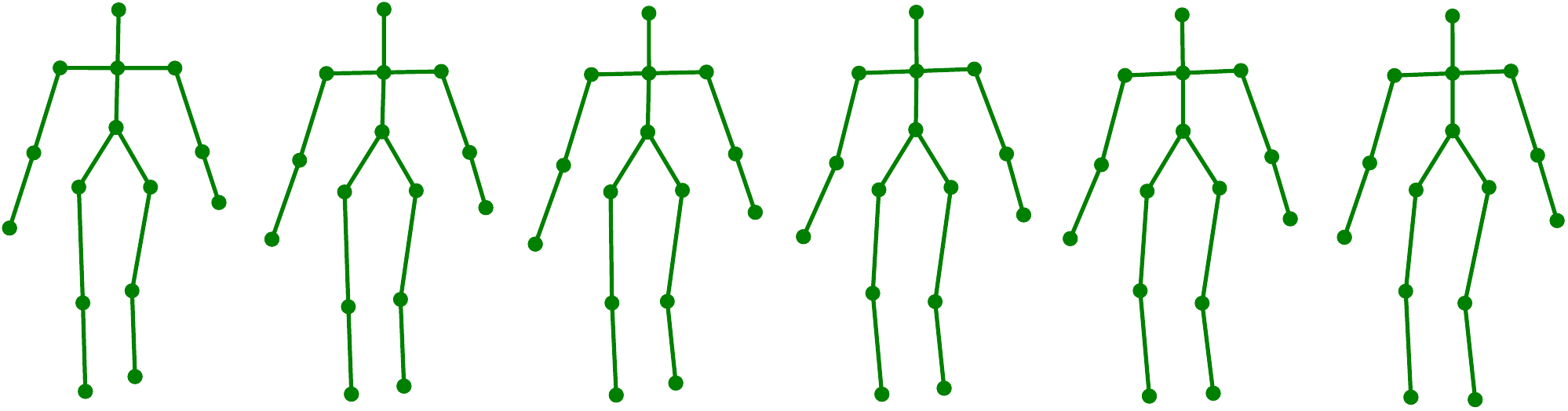}}       \quad  \quad \quad \quad 
 \scalebox{0.06}{\includegraphics{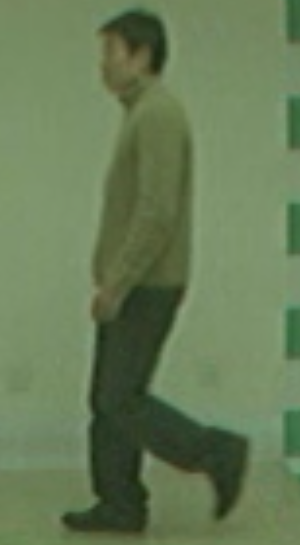}}
       \scalebox{0.06}{\includegraphics{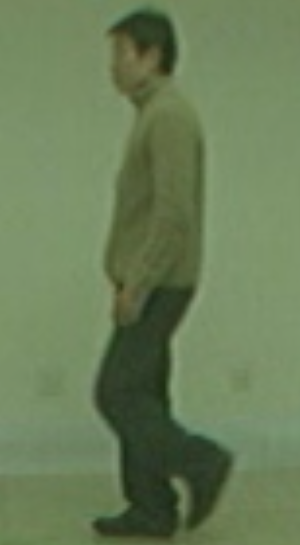}}
        \scalebox{0.06}{\includegraphics{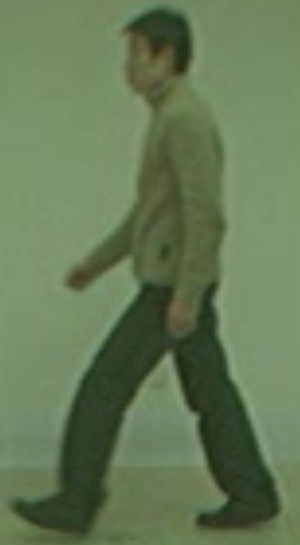}}
         \scalebox{0.06}{\includegraphics{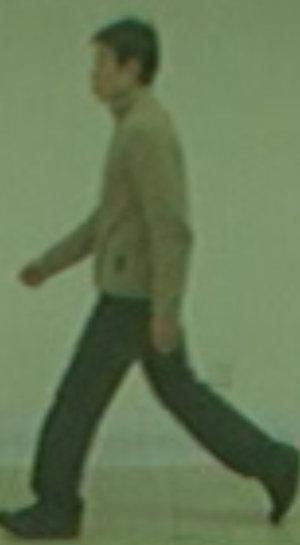}}
          \scalebox{0.06}{\includegraphics{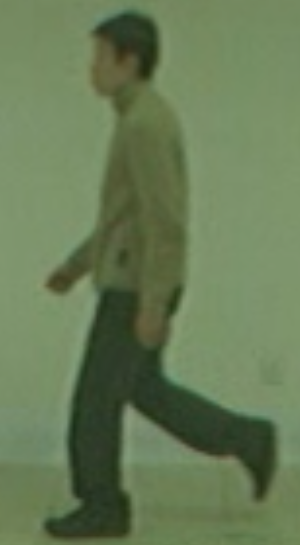}}
           \scalebox{0.06}{\includegraphics{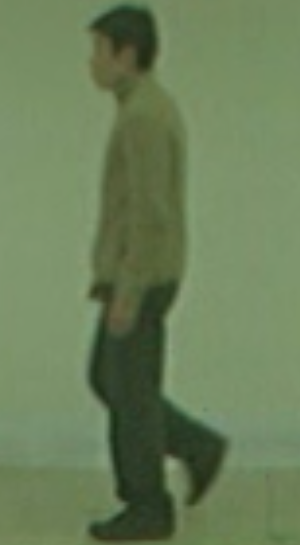}} \quad \quad   \scalebox{0.09}{\includegraphics{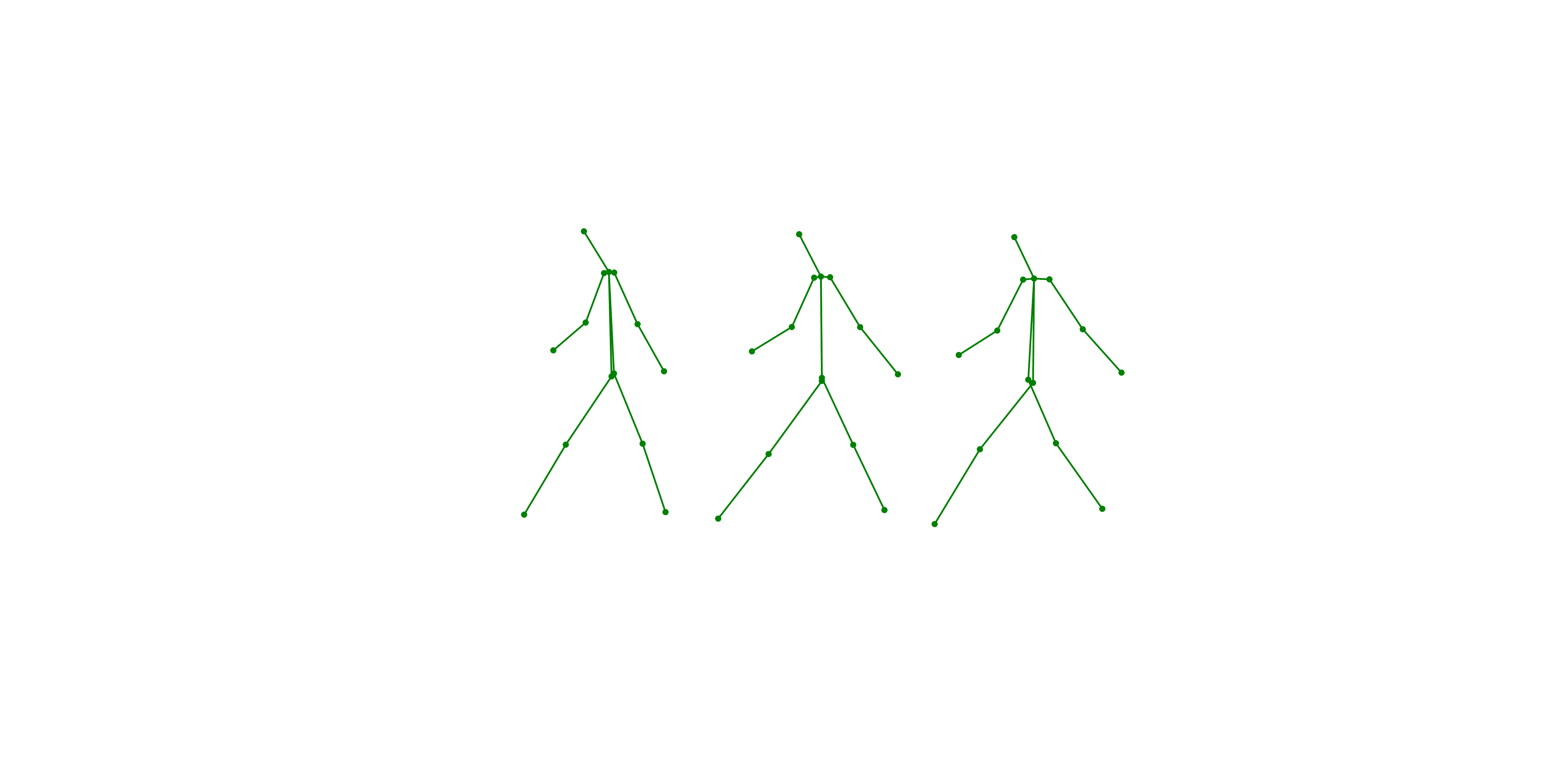}} \  \scalebox{0.09}{\includegraphics{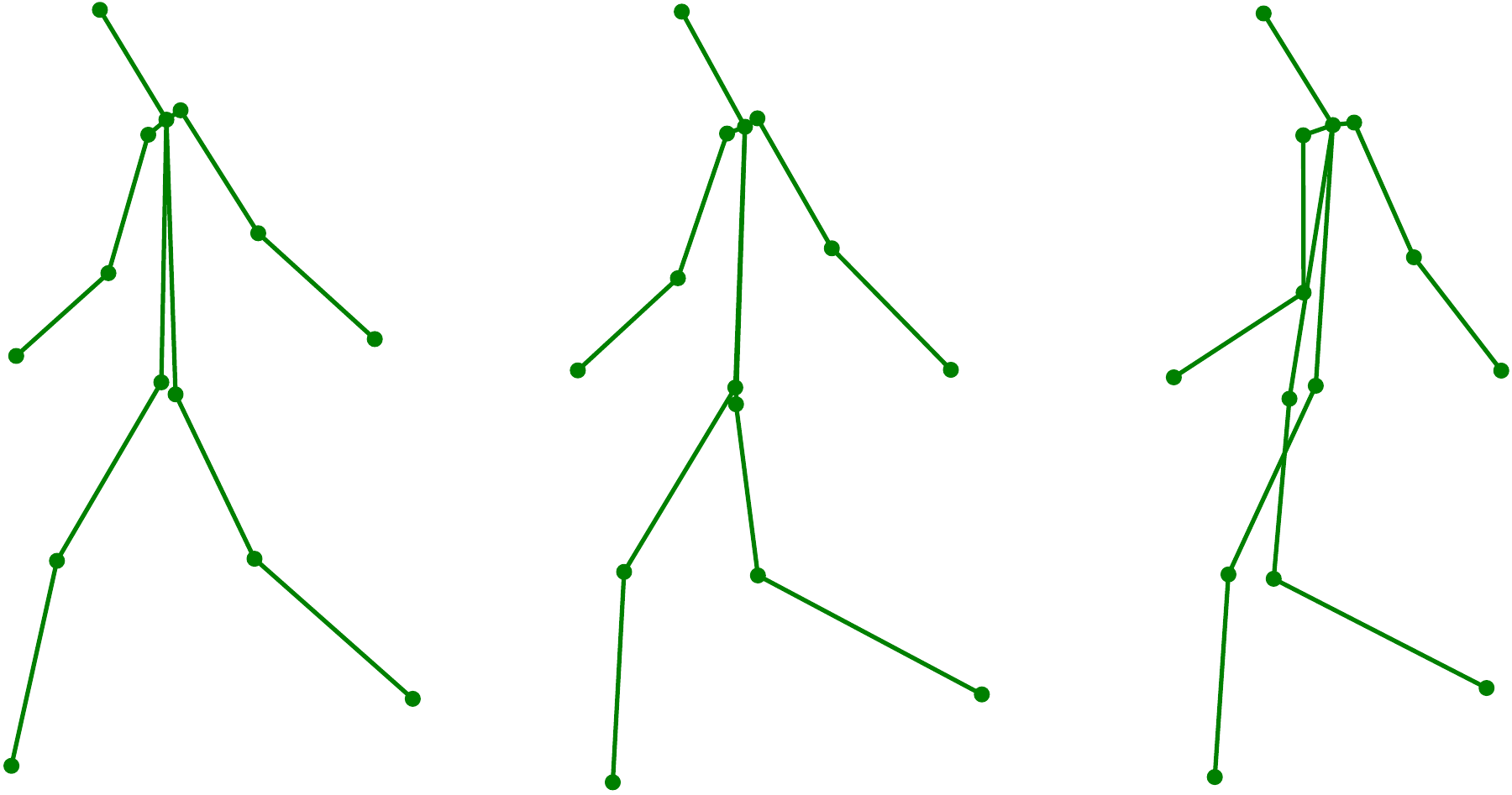}}   
    \caption{\hc{Examples of RGB images and 3D skeletons in BIWI (first row), IAS-Lab (second row) and CASIA B (third row). Note that the last skeleton sequence is estimated from RGB images of CASIA B. }}
    \label{sample_skeleton}
\end{figure}

\section{Experiments}
\label{experiments}
\subsection{Experimental Setup}
 We evaluate our method on four person Re-ID datasets that provide 3D skeleton data: \textit{BIWI} \cite{munaro2014one}, \textit{IAS-Lab} \cite{munaro2014feature}, \textit{KS20 VisLab Multi-View Kinect Skeleton Dataset} \cite{nambiar2017context}, \textit{Kinect Gait Biometry Dataset (KGBD)} \cite{andersson2015person}, \hc{and a large RGB video based multi-view gait dataset \textit{CASIA B} \cite{yu2006framework}.
 They collect skeleton data from 50, 11, 20, 164 and 124 different individuals respectively} (detailed in Table \ref{dataset_detail}). \hc{As to the former four skeleton-based Re-ID datasets,} we follow the evaluation setup in \cite{haque2016recurrent}, which is frequently used in the literature: For BIWI, we use the training set and \textit{walking} testing set, which contain dynamic skeleton data; For IAS-Lab, we use the full training set and two test splits, IAS-A and IAS-B; For KGBD, since no training and testing splits are given, we randomly leave one skeleton video of each person for testing and use the rest of videos for training. For KS20, we design different split setup to evaluate the multi-view Re-ID performance of our approach: (1) Random Splits (RS): For each viewpoint, we randomly select two skeleton videos for training and use the rest of videos for testing. (2) Cross-View Splits (CVS): We test each viewpoint in KS20 (including left lateral at $0^{\circ}$, left diagonal at $30^{\circ}$, frontal at $90^{\circ}$, right diagonal at $130^{\circ}$, and right lateral at $180^{\circ}$) and use the remaining four viewpoints for training. For each original skeleton sequence that corresponds to an individual person in the dataset, we discard the first and last 10 skeleton frames to avoid ineffective skeleton recording. Then, we spilt the given original skeleton sequences in the dataset into multiple shorter skeleton sequences (\textit{i.e.}, $\boldsymbol{S}^{(i)}$) with length $f$ by a step of $\frac{f}{2}$, which aims to obtain as many 3D skeleton sequences as possible to train our model. Unless explicitly specified, the skeleton sequence $\boldsymbol{S}^{(i)}$ in this paper refers to those split sequences used in learning, rather than those original skeleton sequences provided by datasets.

\begin{table*}[t]
\caption{Comparison with existing skeleton-based methods (11-16). Depth-based methods (1-4) and multi-modal methods (5-10) are also included as a reference. Bold numbers refer to the best performers among skeleton-based methods. ``—'' indicates no published result. ``Rev. Rec. '' (17) denotes using CAGES learned by the proposed reverse reconstruction, and ``Rev. Rec. Plus'' (18) represents the proposed enhanced model that concatenates CAGEs learned from three pretext tasks for person Re-ID. Best results using average prediction (AP) are reported in 17-18.}
\label{skeleton_results}
\scalebox{0.94}{
\setlength{\tabcolsep}{1.2mm}{
\begin{tabular}{ccl|rrrrr|rrrrr}
\specialrule{0.1em}{0.45pt}{0.45pt}
\multicolumn{1}{l}{}                                                                          & \multicolumn{1}{l}{}            &                              & \multicolumn{5}{c|}{\textit{\textbf{Rank-1 (\%)}}}                                                                                                                                   & \multicolumn{5}{c}{\textit{\textbf{nAUC}}}                                                                                                                                          \\ \specialrule{0.1em}{0.45pt}{0.45pt}
\multicolumn{1}{l}{}                                                                          & \multicolumn{1}{l}{\textbf{Id}} & \textbf{Methods}             & \multicolumn{1}{l}{\textbf{BIWI}} & \multicolumn{1}{l}{\textbf{IAS-A}} & \multicolumn{1}{l}{\textbf{IAS-B}} & \multicolumn{1}{l}{\textbf{KS20}} & \multicolumn{1}{l|}{\textbf{KGBD}} & \multicolumn{1}{l}{\textbf{BIWI}} & \multicolumn{1}{l}{\textbf{IAS-A}} & \multicolumn{1}{l}{\textbf{IAS-B}} & \multicolumn{1}{l}{\textbf{KS20}} & \multicolumn{1}{l}{\textbf{KGBD}} \\ \specialrule{0.1em}{0.45pt}{0.45pt}
\multirow{4}{*}{\textbf{\begin{tabular}[c]{@{}c@{}}Depth-Based\\ \\ Methods\end{tabular}}}    & 1                               & Gait Energy Image \cite{chunli2010behavior}            & 21.4                              & 25.6                               & 15.9                               & —                            & —                             & 73.2                              & 72.1                               & 66.0                               & —                            & —                            \\
                                                                                              & 2                               & Gait Energy Volume \cite{sivapalan2011gait}           & 25.7                              & 20.4                               & 13.7                               & —                            & —                             & 83.2                              & 66.2                               & 64.8                               & —                            & —                            \\
                                                                                              & 3                               & 3D LSTM \cite{haque2016recurrent}                      & 27.0                              & 31.0                               & 33.8                               & —                            & —                             & 83.3                              & 77.6                               & 78.0                               & —                            & —                            \\
                                                                                              & 4                               & 3D CNN + Average Pooling \cite{boureau2010theoretical}     & 27.8                              & 33.4                               & 39.1                               & —                            & —                             & 84.0                              & 81.4                               & 82.8                               & —                            & —                            \\ \specialrule{0.1em}{0.45pt}{0.45pt}
\multirow{6}{*}{\textbf{\begin{tabular}[c]{@{}c@{}}Multi-Modal\\ \\ Methods\end{tabular}}}    & 5                               & PCM + Skeleton \cite{munaro20143d}               & 42.9                              & 27.3                               & 81.8                               & —                            & —                             & —                            & —                             & —                             & —                            & —                            \\
                                                                                              & 6                               & Size-Shape decriptors + SVM \cite{hasan2016long}  & 20.5                              & —                             & —                             & —                            & —                             & —                            & —                             & —                             & —                            & —                            \\
                                                                                              & 7                               & Size-Shape decriptors + LDA \cite{hasan2016long}  & 22.1                              & —                             & —                             & —                            & —                             & —                            & —                             & —                             & —                            & —                            \\
                                                                                              & 8                               & DVCov + SKL \cite{wu2017robust}                 & 21.4                              & 46.6                               & 45.9                               & —                            & —                             & —                            & —                             & —                             & —                            & —                            \\
                                                                                              & 9                               & ED + SKL \cite{wu2017robust}                    & 30.0                              & 52.3                               & 63.3                               & —                            & —                             & —                            & —                             & —                             & —                            & —                            \\
                                                                                              & 10                              & CNN-LSTM with RTA \cite{karianakis2018reinforced}           & 50.0                              & —                             & —                             & —                            & —                             & —                            & —                             & —                             & —                            & —                            \\ \specialrule{0.1em}{0.45pt}{0.45pt}
\multirow{8}{*}{\textbf{\begin{tabular}[c]{@{}c@{}}Skeleton-Based\\ \\ Methods\end{tabular}}} & 11                              & $D^{13}$ descriptors + SVM \cite{munaro2014one}        & 17.9                              & —                             & —                             & —                            & —                             & —                            & —                             & —                             & —                            & —                            \\
                                                                                              & 12                              & $D^{13}$ descriptors + KNN \cite{munaro2014one}       & 39.3                              & 33.8                               & 40.5                               & 58.3                              & 46.9                               & 64.3                              & 63.6                               & 71.1                               & 78.0                              & 90.0                              \\
                                                                                              & 13                              & $D^{16}$ descriptors + Adaboost \cite{pala2019enhanced}  & 41.8                              & 27.4                               & 39.2                               & 59.8                              & 69.9                               & 74.1                              & 65.5                               & 78.2                               & 78.8                              & 90.6                              \\
                                                                                              & 14                              & Single-layer LSTM \cite{haque2016recurrent}           & 15.8                              & 20.0                               & 19.1                               & 80.9                              & 39.8                               & 65.8                              & 65.9                               & 68.4                               & 92.3                              & 87.2                              \\
                                                                                              & 15                              & Multi-layer LSTM \cite{zheng2019relational}            & 36.1                              & 34.4                               & 30.9                               & 81.6                              & 46.2                               & 75.6                              & 72.1                               & 71.9                               & 94.2                              & 89.8                              \\
                                                                                              & 16                              & PoseGait \cite{liao2020model}                    & 33.3                              & 41.4                               & 37.1                               & 70.5                              & 90.6                               & 81.8                              & 79.9                               & 74.8                               & 94.0                              & 97.8                              \\
                                                                                              & 17                              & Ours (Rev. Rec. )   & 62.9                              & \textbf{60.1}                      & \textbf{62.5}                      & 86.9                              & 86.9                               & 86.8                              & \textbf{82.9}                      & \textbf{86.9}                      & \textbf{94.9}                     & 97.1                              \\
                                                                                              & 18                              & Ours (Rev. Rec. Plus) & \textbf{63.3}                     & 59.1                               & 62.2                               & \textbf{92.0}                     & \textbf{90.6}                      & \textbf{88.3}                     & 81.5                               & 86.2                               & \textbf{94.9}                     & \textbf{98.1}                     \\ \specialrule{0.1em}{0.2pt}{0.2pt}
\end{tabular}
}
}
\end{table*}

\begin{table}[t]
\centering
\caption{Re-ID performance comparison on cross-view splits (CVS) of KS20 dataset. $0^{\circ}, 30^{\circ}, 90^{\circ}, 130^{\circ},$ and $180^{\circ}$ represent different viewpoints.}
\label{CVS_RS_results}
\setlength{\tabcolsep}{1.1mm}{
\begin{tabular}{c|l|ccccc}
\specialrule{0.1em}{0.45pt}{0.45pt}
                                          & \textbf{Methods}      & $\boldsymbol{0^{\circ}}$ & $\boldsymbol{30^{\circ}}$ & $\boldsymbol{90^{\circ}}$ & $\boldsymbol{130^{\circ}}$ & $\boldsymbol{180^{\circ}}$ \\ \specialrule{0.1em}{0.45pt}{0.45pt}
\multirow{3}{*}{\textit{\textbf{Rank-1}}} & PoseGait              & 24.6       & 19.1        & 29.7        & 27.3         & 25.0         \\
                                          & Ours (Rev. Rec.)      & 44.4       & \textbf{54.9}        & \textbf{55.0}        & 41.9         & 53.4         \\
                                          & Ours (Rev. Rec. Plus) & \textbf{48.8}       & 53.6        & 54.9        & \textbf{44.5}         & \textbf{57.5}         \\ \hline
\multirow{3}{*}{\textit{\textbf{nAUC}}}   & PoseGait              & 81.2       & 75.4        & 81.0        & 79.6         & 85.1         \\
                                          & Ours (Rev. Rec.)      & 84.8       & \textbf{89.1}        & \textbf{87.8}        & 83.3         & 89.3         \\
                                          & Ours (Rev. Rec. Plus) & \textbf{86.5}       & 87.1        & 84.8        & \textbf{83.8}         & \textbf{91.8}         \\ \specialrule{0.1em}{0.2pt}{0.2pt}
\end{tabular}
}
\end{table}

\hc{Different from the aforementioned skeleton-based datasets, CASIA B dataset is a large-scale RGB video based dataset without providing original skeleton data. To apply our method to RGB-based datasets for person Re-ID tasks, we exploit pre-trained pose estimation model \cite{cao2019openpose, chen20173d} to extract 3D skeletons (see Fig. \ref{sample_skeleton}) from RGB videos in CASIA B (detailed in supplementary material). CASIA B contains 124 individuals with 11 views---$0^{\circ}$, $18^{\circ}$, $36^{\circ}$, $54^{\circ}$, $72^{\circ}$, $90^{\circ}$, $108^{\circ}$, $126^{\circ}$, $144^{\circ}$, $162^{\circ}$, $180^{\circ}$ and three conditions---pedestrians wearing a bag (``bag''), wearing a coat (``clothes''), and without any coat or bag (``normal''). We adopt two setups for performance evaluation: (1) Cross-View Evaluation (CVE): We evaluate each view in CASIA B while adjacent views are used for training. (2) Condition-based Matching Evaluation (CME) \cite{liu2015enhancing}: We randomly and equally divide 124 people IDs to training set and testing set, and divide the testing set by three original conditions (``normal'', ``bag'', ``clothes'') to be gallery or probe sets. In particular, we evaluate our approach on single-condition ($i.e.,$ gallery set and probe set keep the same condition without appearance changes) and on cross-condition settings ($i.e.,$ probe set is under normal condition (``Nm'') while gallery sets are under bag (``Bg'') or clothes condition (``Cl'')). Hence, the CME setup contains five combinations, $i.e.,$ ``Nm-Nm'', ``Bg-Bg'', ``Cl-Cl'', ``Bg-Nm'', and ``Cl-Nm'', where the first condition is used for probe set and the second one is for gallery. The details for CME setup can be found in \cite{liu2015enhancing}. Experiments under each evaluation setup are repeated for multiple times and the average performance is reported in Sec. \ref{CASIA_B_evaluation}.} \rhc{The implementation details are provided in the supplementary material, and our codes are open at \href{https://github.com/Kali-Hac/Locality-Awareness-SGE}{https://github.com/Kali-Hac/Locality-Awareness-SGE. }}

\begin{table*}[ht]
    \centering
    \caption{Ablation study of our model. ``\checkmark'' \ indicates that the corresponding model component is used: GE, GD, reverse skeleton reconstruction (Rev. Rec.), locality-aware attention alignment scores (LAS).
    ``AGEs'' indicates using AGEs ($\boldsymbol{v}_{t}$) rather than encoded gait states of GE's LSTM $\boldsymbol{h}_{t}$ to perform person Re-ID task \hc{(note that here $\boldsymbol{h}_{t}$ and AGEs  ($\boldsymbol{v}_{t}$) are learned without the locality-aware contrastive learning (LCL) scheme)}, and ``LCL+CAGEs'' ($\boldsymbol{\overline{v}}_{t}$) represents exploiting CAGEs learned by the LCL scheme for person Re-ID. }
\label{ablation_t}
\scalebox{0.95}{
\setlength{\tabcolsep}{1.75mm}{
\begin{tabular}{cccclc|cc|cc|cc|cc}
\specialrule{0.1em}{0.45pt}{0.45pt}
\multicolumn{6}{c|}{\textbf{Model Configuration}}                                                                                                                                                            & \multicolumn{2}{c|}{\textbf{BIWI}}                                                         & \multicolumn{2}{c|}{\textbf{IAS-A}}                                                        & \multicolumn{2}{c|}{\textbf{IAS-B}}                                                        & \multicolumn{2}{c}{\textbf{KS20}}                                                         \\ \specialrule{0.1em}{0.45pt}{0.45pt}
\multicolumn{1}{l}{\textbf{GE}} & \multicolumn{1}{l}{\textbf{GD}} & \multicolumn{1}{l}{\textbf{Rev. Rec.}} & \multicolumn{1}{l}{\textbf{LAS}} & \textbf{AGEs}         & \multicolumn{1}{l|}{\textbf{LCL+CAGEs}} & \multicolumn{1}{l}{\textit{\textbf{Rank-1}}} & \multicolumn{1}{l|}{\textit{\textbf{nAUC}}} & \multicolumn{1}{l}{\textit{\textbf{Rank-1}}} & \multicolumn{1}{l|}{\textit{\textbf{nAUC}}} & \multicolumn{1}{l}{\textit{\textbf{Rank-1}}} & \multicolumn{1}{l|}{\textit{\textbf{nAUC}}} & \multicolumn{1}{l}{\textit{\textbf{Rank-1}}} & \multicolumn{1}{l}{\textit{\textbf{nAUC}}} \\ \specialrule{0.1em}{0.45pt}{0.45pt}
\checkmark                               &                                 &                                   &                                  &                       &                                           & 36.1                                         & 75.6                                        & 34.4                                         & 72.1                                        & 30.9                                         & 71.9                                        & 80.9                                         & 92.3                                       \\
\checkmark                              & \checkmark                               &                                   &                                  &                       &                                           & 41.5                                         & 80.1                                        & 48.1                                         & 77.5                                        & 48.4                                         & 76.2                                        & 83.3                                         & 92.2                                       \\
\checkmark                               & \checkmark                               & \checkmark                                 &                                  &                       &                                           & 46.7                                         & 81.5                                        & 50.9                                         & 78.3                                        & 52.9                                         & 80.3                                        & 84.5                                         & 94.2                                       \\
\checkmark                               & \checkmark                               & \checkmark                                 & \checkmark                                &                       &                                           & 57.7                                         & 85.8                                        & 55.4                                         & 81.6                                        & 57.4                                         & 83.6                                        & 85.7                                         & 94.1                                       \\
\checkmark                               & \checkmark                               &                                   & \checkmark                                & \multicolumn{1}{c}{\checkmark} &                                           & 57.2                                         & 85.7                                        & 55.6                                         & 80.7                                        & 57.0                                         & 84.8                                        & 86.5                                         & 94.7                                       \\
\checkmark                               & \checkmark                               & \checkmark                                 & \checkmark                                & \multicolumn{1}{c}{\checkmark} &                                           & 59.1                                         & 86.5                                        & 56.1                                         & 80.7                                        & 58.2                                         & 85.3                                        & 86.7                                         & 93.2                                       \\
\checkmark                               & \checkmark                               &                                   & \checkmark                                &                       & \checkmark                                         & 59.7                                         & 86.6                                        & 57.9                                         & 82.7                                        & 60.9                                         & 84.3                                        & 85.9                                         & 94.7                                       \\
\checkmark                              & \checkmark                               & \checkmark                                 & \checkmark                                &                       & \checkmark                                         & 62.9                                         & 86.8                                        & 60.1                                         & 82.9                                        & 62.5                                         & 86.2                                        & 86.9                                         & 95.7                                       \\ \specialrule{0.1em}{0.2pt}{0.2pt}
\end{tabular}
}
}
\end{table*}

\begin{table}[t]
\centering
\caption{Performance comparison of different pretext tasks and the proposed enhanced approach under two Re-ID manners (``AP'': Average prediction. ``SC'': Sequence-level concatenation). Bold numbers refer to the best \textit{Rank-1} accuracy and \textit{nAUC} among different configurations. }
\label{pretext_results_comp}
\scalebox{0.93}{
\setlength{\tabcolsep}{2.6mm}{
\begin{tabular}{cl|cccc}
\specialrule{0.1em}{0.45pt}{0.45pt}
\textbf{}                       & \multicolumn{1}{c|}{\textbf{}} & \multicolumn{2}{c}{\textit{\textbf{Rank-1}}} & \multicolumn{2}{c}{\textit{\textbf{nAUC}}} \\ \specialrule{0.1em}{0.45pt}{0.45pt}
\textbf{Dataset}                & \textbf{\hc{Pretext Task(s)}}          & \textbf{AP}            & \textbf{SC}         & \textbf{AP}           & \textbf{SC}        \\ \hline
\multirow{7}{*}{\textbf{BIWI}}  & Prediction                     & 40.4                   & 32.0                & 82.0                  & 79.7               \\
                                & Sorting                        & 55.7                   & 43.4                & 71.4                  & 85.5               \\
                                & Rev. Rec.                      & 62.9                   & 51.3                & 86.8                  & 84.1               \\
                                 & \hc{Pred. + Sort.}                  & 62.1                   & 51.5                & 86.7                  & 86.9               \\
                                & \hc{Rev. Rec. + Pred.}                 & 62.3                   & 52.5                & 87.5                  & 86.6             \\
                                & \hc{Rev. Rec. + Sort.}                 & 63.1                   & 51.3                & 88.0                  & 86.0               \\
                                & Rev. Rec. Plus                 & \textbf{63.3}          & 53.7                & \textbf{88.3}         & 87.1               \\ \specialrule{0.1em}{0.45pt}{0.45pt}
\multirow{7}{*}{\textbf{IAS-A}} & Prediction                     & 56.7                   & 54.6                & 82.3                  & 72.9               \\
                                & Sorting                        & 56.8                   & 51.9                & \textbf{82.9}         & 78.9               \\
                                & Rev. Rec.                      & \textbf{60.1}          & 55.0                & \textbf{82.9}         & 81.1               \\
                                  & \hc{Pred. + Sort.}                  & 58.8                   & 54.0                & 80.6                  & 80.1               \\
                                & \hc{Rev. Rec. + Pred.}                 & 58.8                   & 52.6                & 81.1                  & 79.1               \\
                                & \hc{Rev. Rec. + Sort.}                 & 58.7                   & 54.1                & 80.9                  & 80.3               \\
                                & Rev. Rec. Plus                 & 59.1                   & 53.0                & 81.5                  & 79.3               \\ \specialrule{0.1em}{0.45pt}{0.45pt}
\multirow{7}{*}{\textbf{IAS-B}} & Prediction                     & 58.5                   & 55.8                & 84.6                  & 76.0               \\
                                & Sorting                        & 54.1                   & 49.5                & 83.6                  & 78.5               \\
                                & Rev. Rec.                      & \textbf{62.5}          & 53.5                & \textbf{86.9}         & 85.6               \\
                                & \hc{Pred. + Sort.}                  & 61.9                   & 54.7                & 86.2                  & 84.7               \\
                                & \hc{Rev. Rec. + Pred.}                 & 61.2                   & 56.7                & 86.7                  & 85.7               \\
                                & \hc{Rev. Rec. + Sort.}                 & 61.5                   & 55.0                & 86.0                  & 85.1               \\
                                & Rev. Rec. Plus                 & 62.2                   & 57.6                & 86.2                  & 85.2               \\ \specialrule{0.1em}{0.45pt}{0.45pt}
\multirow{7}{*}{\textbf{KS20}}  & Prediction                     & 81.6                   & 74.0                & 77.5                  & 89.9               \\
                                & Sorting                        & 83.5                   & 78.1                & 90.6                  & 91.7               \\
                                & Rev. Rec.                      & 86.9                   & 84.1                & \textbf{94.9}         & 85.7               \\
                               & \hc{Pred. + Sort.}                  & 89.5                   & 86.3                & 92.8                  & 88.4               \\
                                & \hc{Rev. Rec. + Pred.}                 & 91.9                   & 86.0                & 94.5                  & 90.1               \\
                                & \hc{Rev. Rec. + Sort.}                 & 90.9                   & 84.7                & 93.1                  & 87.4               \\
                                & Rev. Rec. Plus                 & \textbf{92.0}          & 89.4                & \textbf{94.9}         & 89.5               \\ \specialrule{0.1em}{0.45pt}{0.45pt}
\multirow{7}{*}{\textbf{KGBD}}  & Prediction                     & 85.5                   & 85.1                & 96.5                  & 97.9               \\
                                & Sorting                        & 85.4                   & 84.2                & 97.4                  & 98.0               \\
                                & Rev. Rec.                      & 86.9                   & 84.1                & 97.1                  & 97.4               \\
                                  & \hc{Pred. + Sort.}                  & 90.4                   & 87.6                & 97.9                  & 97.8               \\
                                & \hc{Rev. Rec. + Pred.}                 & \textbf{90.6}          & 88.5                & 97.1                  & 97.7               \\
                                & \hc{Rev. Rec. + Sort.}                 & 90.5                   & 87.5                & 97.4                  & 96.9               \\
                                & Rev. Rec. Plus                 & \textbf{90.6}          & 88.3                & \textbf{98.1}         & 98.0               \\ \specialrule{0.1em}{0.2pt}{0.2pt}
\end{tabular}
}
}
\end{table}

\subsection{Evaluation Metrics} 
Person Re-ID is typically evaluated in a multi-shot manner, and the sequence label can be produced by either predictions of multiple frames or a sequence-level representation. In this work, we report the performance of both strategies (see Table \ref{skeleton_results} and Table \ref{pretext_results_comp}): (1) (SC) Using sequence-level CAGEs for person Re-ID. (2) (AP) Averaging the prediction of each skeleton-level CAGE in a skeleton sequence to be the final sequence-level prediction for person Re-ID. We compute \textit{Rank-1} accuracy and \textit{nAUC} (area under the cumulative matching curve (CMC) normalized by the number of ranks \cite{gray2008viewpoint}) to quantify multi-shot person Re-ID performance. \hc{It should be noted that we adopt AP strategy for Cross-View Evaluation (CVE) on CASIA B. For Condition-based Matching Evaluation (CME), each sequence-level 
CAGEs in probe set is used to match the one of the same identity in gallery set using Euclidean distance, and the \textit{Rank-1} matching rate is computed. }

\subsection{Performance Comparison}
\label{performance_comp}
In this section, we conduct an comprehensive comparison with existing skeleton based person Re-ID methods (Id = 11-16) in the literature. In the meantime, we also include classic depth-based methods (Id = 1-4) and representative multi-modal methods (Id = 5-10) as a reference. The results are reported as follows.

\subsubsection{Comparison with Skeleton-based Methods} As shown by Table \ref{skeleton_results}, our approach enjoys obvious advantages over existing skeleton-based methods in terms of both Re-ID performance metrics: First, our approach evidently outperforms those methods that rely on manually-designed geometric or anthropometric skeleton descriptors (Id = 11-13). For example, $D^{13}$ (Id = 12) and recent $D^{16}$ (Id = 13) are two most representative hand-crafted feature based methods, and our model outperforms both of them by a large margin ($20.7\%$-$43.7\%$ \textit{Rank-1} accuracy and $7.5\%$-$24.0\%$ \textit{nAUC} on different datasets). Second, our approach is also superior to recent skeleton based methods that utilize deep neural networks (Id = 14-16) on all datasets by up to $50.8\%$ \textit{Rank-1} accuracy and $22.5\%$ \textit{nAUC} improvement. Although the latest PoseGait can achieve comparable performance to our approach on the KGBD dataset, it still requires extracting 81 hand-crafted features for CNN learning. Besides, labeled skeleton data are indispensable for the gait encoding stage of existing deep learning based methods, while our approach can learn better gait representations by simply exploiting unlabeled 3D skeleton data.

Besides, we also evaluate the cross-view Re-ID performance of our approach on the KS20 dataset using multi-view 3D skeleton data provided by this dataset. We compare its performance with the latest PoseGait \cite{liao2020model} approach, which specially considers the multi-view scenario and achieves the best overall performance among existing skeleton based methods. The results are displayed in Table \ref{CVS_RS_results} and highlight the following conclusions:
Our approach consistently outperforms PoseGait by a considerable margin (up to $35.8\%$ \textit{Rank-1} accuracy and $13.7\%$ \textit{nAUC}) under all viewpoints. In the meantime, it can stably achieve comparatively satisfactory performance under different viewpoints, which validates the robustness of our approach against viewpoint variations. 


\subsubsection{Comparison with Depth-based Methods and Multi-modal Methods} 
\label{multi_approaches} Despite that our approach only takes 3D skeleton data as inputs, our approach consistently outperforms baselines of classic depth-based methods (Id = 1-4) by at least $23.4\%$ \textit{Rank-1} and $1.5\%$ \textit{nAUC} gain. Considering the fact that 3D skeletons are of much smaller data size than depth image data, our approach is both effective and efficient. As to the comparison with recent methods that exploit multi-modal inputs (Id = 5-10), the performance of our approach is still highly competitive: Although in few cases multi-modal methods perform better on IAS-B, our skeleton based method achieves the best \textit{Rank-1} accuracy on BIWI and IAS-A. Interestingly, we note that the multi-modal approach that uses both point cloud matching (PCM) and skeletons yields the best accuray on IAS-B, but it performs markedly worse on datasets that undergo more frequent shape and appearance changes (IAS-A and BIWI). By contrast, our approach consistently achieves stable and satisfactory performance on each dataset. Thus, with 3D skeleton data as the sole input, our approach can be a promising solution to person Re-ID and other potential skeleton-related tasks.



\section{Further Analysis}
\label{discuss}

\begin{figure*}[t]
    \centering
    \scalebox{0.345}{\includegraphics{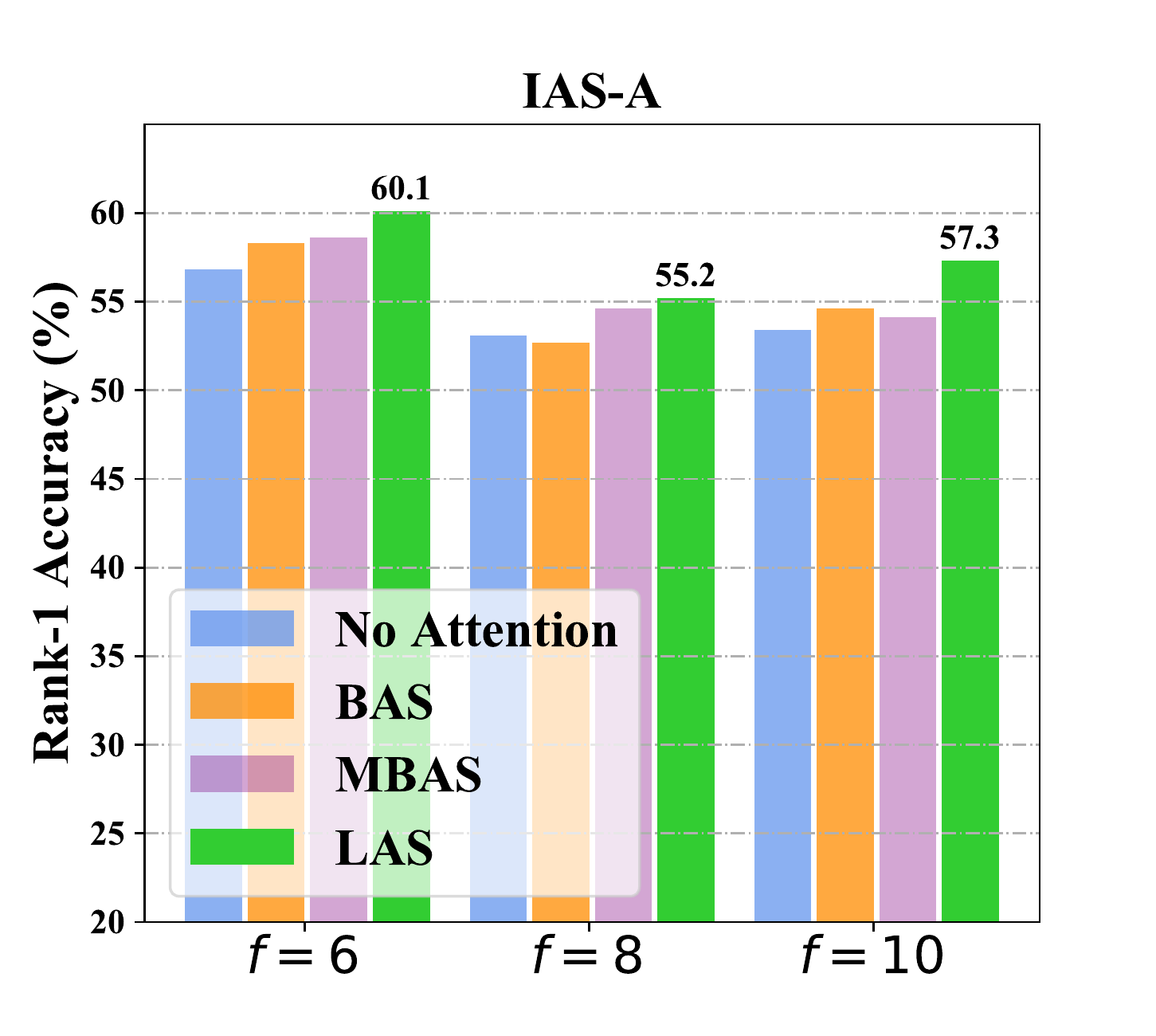}}
    \scalebox{0.345}{\includegraphics{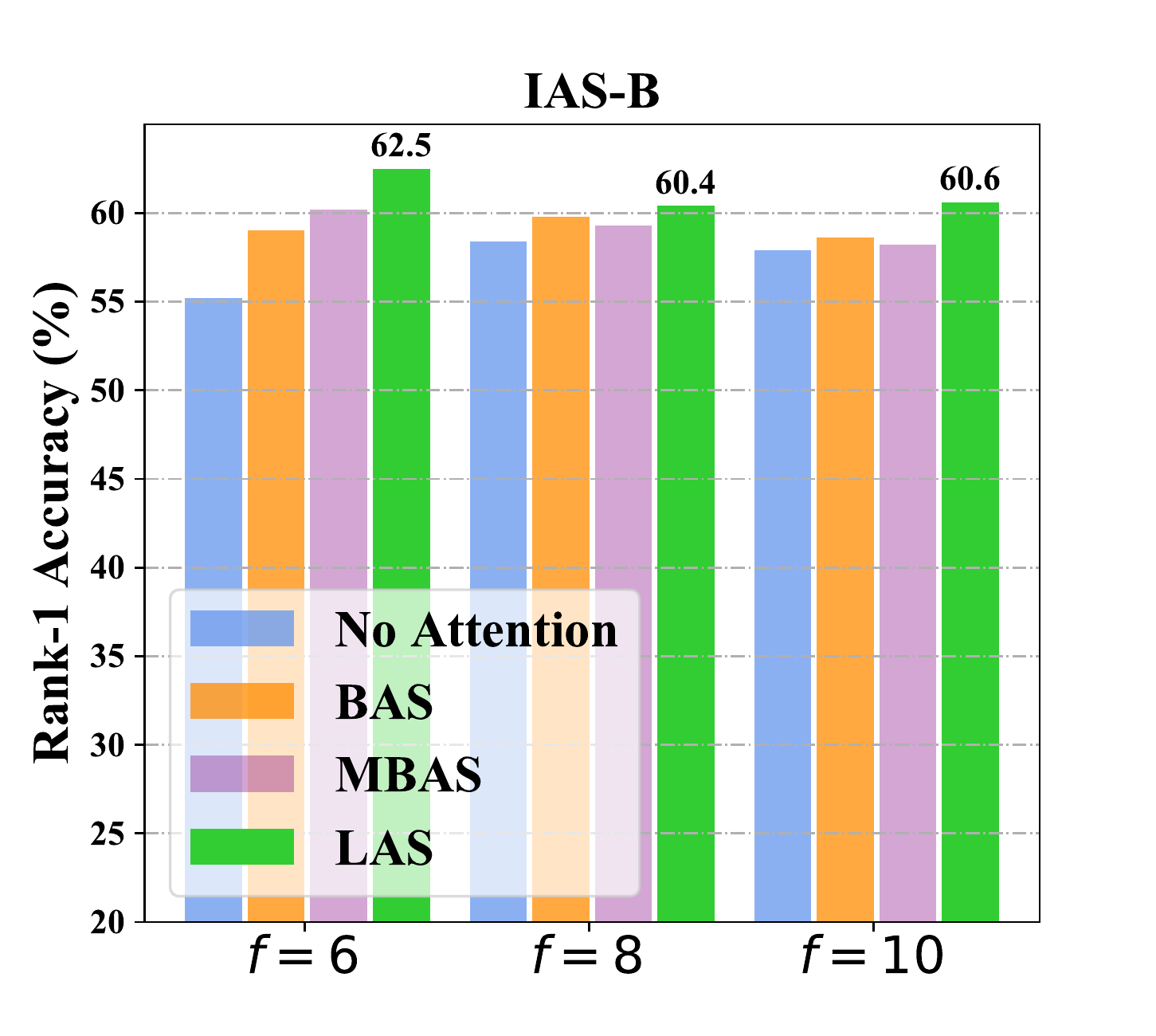}}
    \scalebox{0.345}{\includegraphics{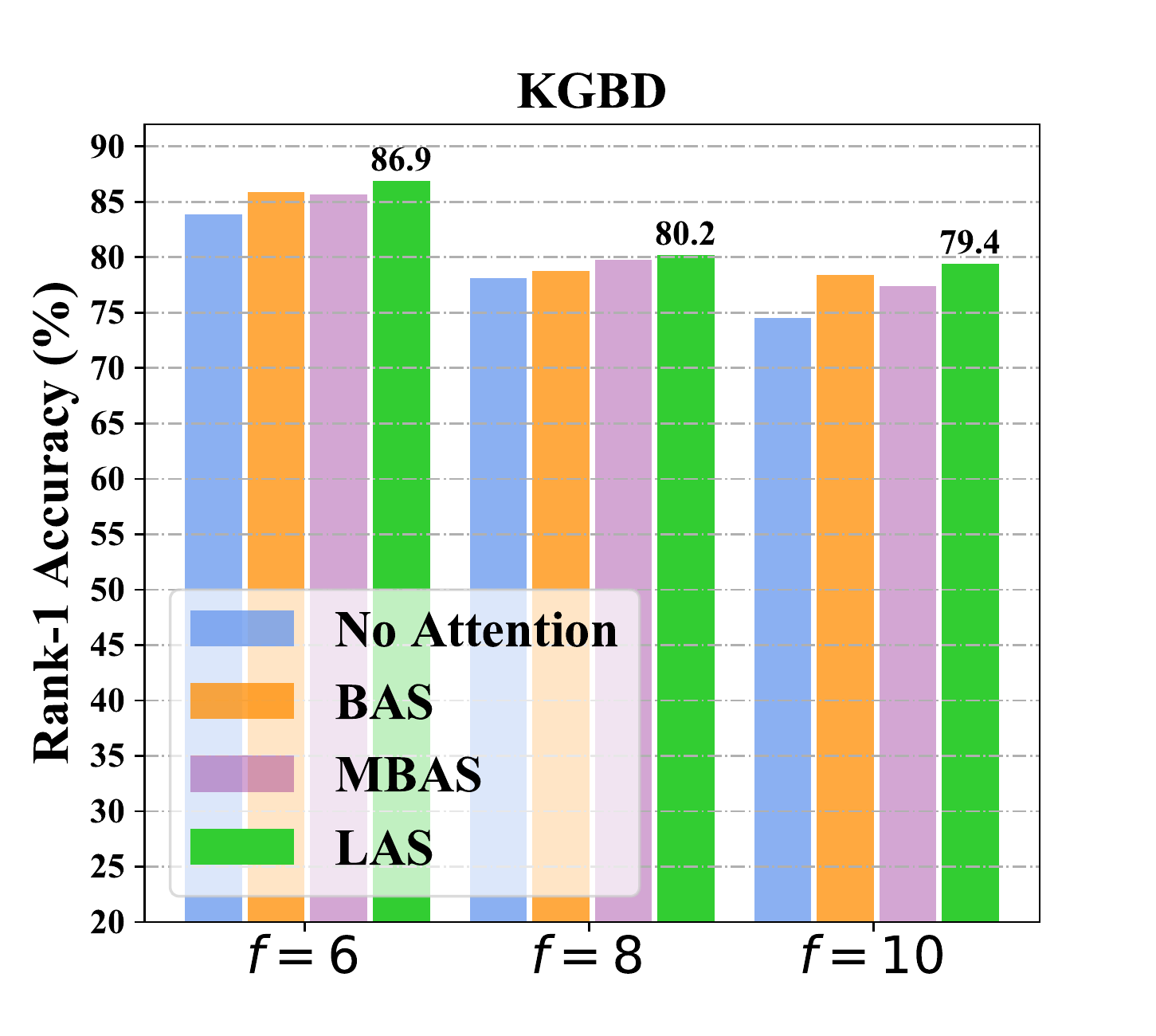}}
    \scalebox{0.345}{\includegraphics{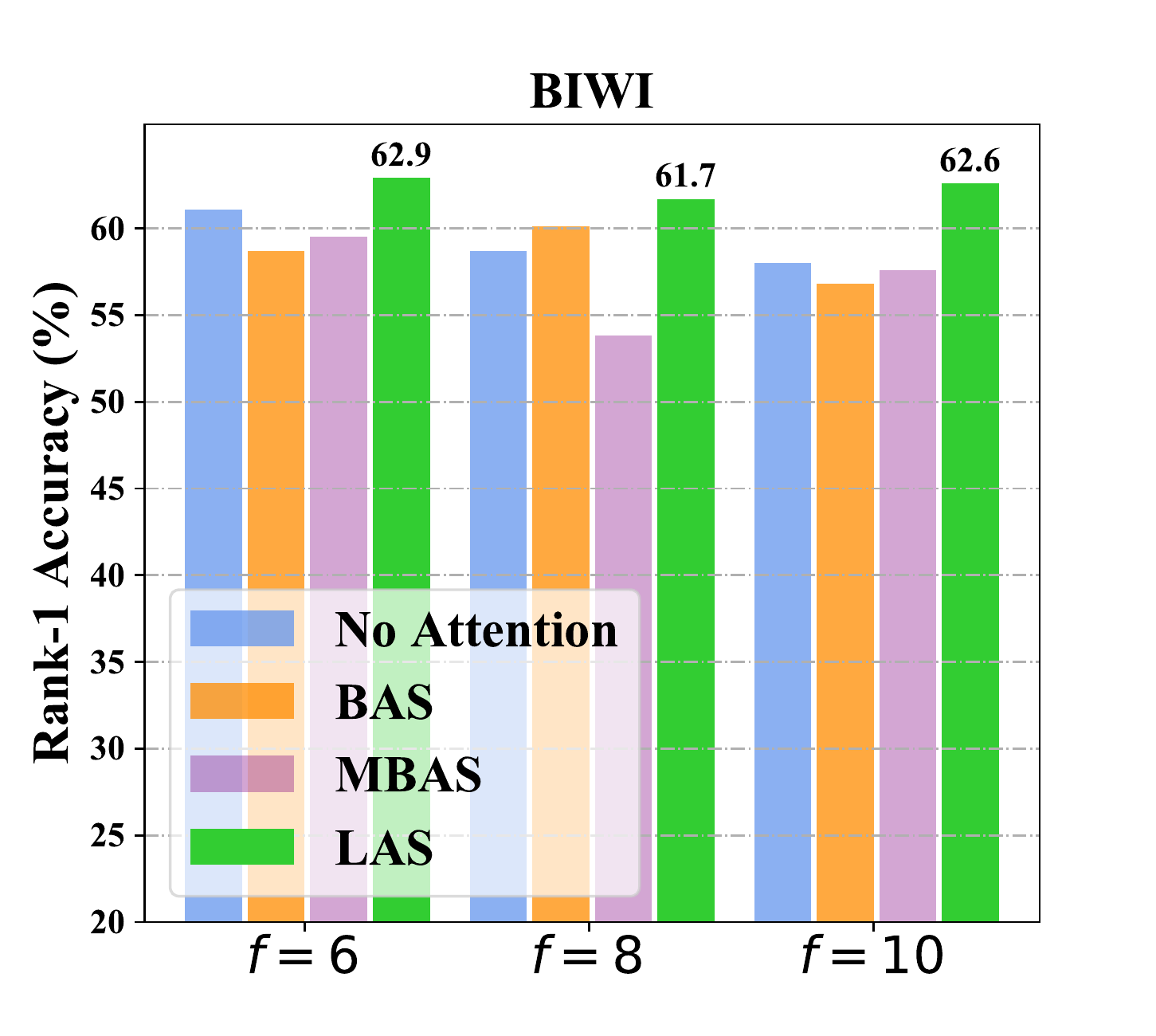}}
    \caption{\textit{Rank-1} accuracy on different datasets when using no attention, BAS, MBAS or LAS for model learning. $f$ denotes the sequence length.}
    \label{attention_comp}
\end{figure*}

\begin{table*}[h]
\centering
\caption{Performance of our approach when setting different contrasting intervals for learning (``Interval=1'' indicates contrasting adjacent sequences).}
\label{contrastive_interval_results}
\setlength{\tabcolsep}{3.5mm}{
\begin{tabular}{c|ccccc|ccccc}
\specialrule{0.1em}{0.45pt}{0.45pt}
\textbf{}         & \multicolumn{5}{c|}{\textbf{\textit{Rank-1}}}                                            & \multicolumn{5}{c}{\textbf{\textit{nAUC}}}                                               \\ \specialrule{0.1em}{0.45pt}{0.45pt}
\textbf{Interval} & \textbf{BIWI} & \textbf{IAS-A} & \textbf{IAS-B} & \textbf{KS20} & \textbf{KGBD} & \textbf{BIWI} & \textbf{IAS-A} & \textbf{IAS-B} & \textbf{KS20} & \textbf{KGBD} \\ \specialrule{0.1em}{0.45pt}{0.45pt}
\textbf{1}        & 62.9 & 60.1  & 62.5  & 86.9 & 86.9 & 86.8          & 82.9  & 86.9  & 94.9          & 97.1 \\
\textbf{2}        & 61.1          & 57.9           & 61.4           & 86.5          & 86.4          & 86.5          & 82.5           & 84.8           & 94.4 & 96.8          \\
\textbf{3}        & 61.9          & 59.4           & 58.1           & 86.1          & 86.7          & 86.0          & 82.3           & 84.8           & 94.1          & 96.8          \\
\textbf{4}        & 60.8          & 59.2           & 58.9           & 85.9          & 86.5          & 86.7 & 82.4           & 82.7           & 94.9          & 96.7          \\
\specialrule{0.1em}{0.2pt}{0.2pt}
\end{tabular}
}
\end{table*}
\begin{table*}[t]
\begin{center}
\caption{Performance of our approach when setting different temperatures ($\tau=0.05, 0.1, 0.5, 0.8, 1$) for the LCL scheme on different datasets.}
\label{CAGE_t}
\setlength{\tabcolsep}{3.8mm}{
\begin{tabular}{c|ccccc|ccccc}
\specialrule{0.1em}{0.45pt}{0.45pt}
\textbf{}     & \multicolumn{5}{c|}{\textbf{\textit{Rank-1}}}                                            & \multicolumn{5}{c}{\textbf{\textit{nAUC}}}                                               \\ \specialrule{0.1em}{0.45pt}{0.45pt}
$\boldsymbol{\tau}$    & \textbf{BIWI} & \textbf{IAS-A} & \textbf{IAS-B} & \textbf{KS20} & \textbf{KGBD} & \textbf{BIWI} & \textbf{IAS-A} & \textbf{IAS-B} & \textbf{KS20} & \textbf{KGBD} \\ \specialrule{0.1em}{0.45pt}{0.45pt}
\textbf{1}    & 60.6          & 58.6           & 59.7           & 85.9          & 86.8          & 85.3          & 81.3           & 84.5           & 94.5          & 97.1          \\
\textbf{0.8}  & 61.2          & 59.3           & 60.2           & 86.3          & 86.6          & 85.6          & 82.3           & 85.2           & 95.0          & 97.0          \\
\textbf{0.5}  & 61.1          & 58.5           & 61.5           & 86.3          & 86.9          & 86.8          & 82.0           & 85.8           & 95.0          & 97.1          \\
\textbf{0.1}  & 62.9          & 60.1           & 62.5           & 86.9          & 86.8          & 86.8          & 82.9           & 86.9           & 94.9          & 96.6          \\
\textbf{0.05} & 62.6          & 59.0           & 61.9           & 86.7          & 86.6          & 86.3          & 83.0           & 86.4           & 95.2          & 96.8          \\ \specialrule{0.1em}{0.2pt}{0.2pt}
\end{tabular}
}
\end{center}
\end{table*}

\subsection{Ablation Studies}
\label{ablation_sec}
In this section, we carry out ablation studies to verify the necessity of each model component in the proposed approach. As shown in Table \ref{ablation_t}, we can arrive at the following conclusions: 
\textbf{(1)} The proposed encoder-decoder architecture (GE-GD) performs remarkably better
than the supervised learning paradigm, which uses GE only to perform person Re-ID directly (up to $17.5\%$ \textit{Rank-1} accuracy and $5.4\%$ \textit{nAUC} gain). Such results demonstrate the effectiveness of our self-supervised gait encoding model, which leverages an encoder-decoder architecture and skeleton sequence reconstruction mechanism. 
\textbf{(2)} Using reverse reconstruction (``Rev. Rec.'' in Table \ref{ablation_t}) typically produces evident performance gain (up to $5.2\%$ \textit{Rank-1} accuracy and $4.1\%$ \textit{nAUC}) when compared with those configurations without reverse reconstruction. Such results justify reverse reconstruction as an effective pretext task for gait encoding, as it enables the model to learn more discriminative gait features for person Re-ID.
\textbf{(3)} Adding the proposed locality-aware attention mechanism (``LAS'') is able to improve the performance remarkably by up to $11.0\%$ \textit{Rank-1} accuracy and $4.3\%$ \textit{nAUC}. This observation is consistent with our previous analysis that LAS facilitates reverse reconstruction and contributes to person Re-ID performance. We will compare different attention mechanisms in the next section as well.
\textbf{(4)} The proposed context vector based gait representations, CAGEs and AGEs, are both able to achieve superior performance to frequently-used features ($\boldsymbol{h}_{t}$). \hc{When it comes to the comparison between AGEs (learned without contrastive learning) and CAGEs (learned with locality-aware contrastive learning), CAGEs consistently outperforms AGEs with a $0.2\%$-$4.3\%$ \textit{Rank-1} accuracy and $0.3\%$-$2.5\%$ \textit{nAUC} gain on different datasets (see row 6 and row 8 in Table \ref{ablation_t}). This demonstrates the advantage to incorporate inter-sequence locality by the proposed locality-aware contrastive learning scheme.}

\subsection{Discussions}
\subsubsection{Different Pretext Tasks}
\label{results_of_pretexts}
In this section, we systematically explore the influence of pretext tasks on self-supervised learning in Table \ref{pretext_results_comp}. First, we evaluate the Re-ID performance of different pretext tasks (Prediction, Sorting, and Rev. Rec.) under two Re-ID strategies (average prediction (AP) and sequence-level concatenation (SC)) on different datasets. Note that we apply the basic attention mechanism to prediction and sorting, because they cannot exploit intra-sequence locality and using locality-aware attention actually does not improve their performance. 
Then, we also compare their performance with \hc{all potential combinations of pretext tasks: (1) Pred. + Sort. , (2) Rev. Rec. + Pred. , (3) Rev. Rec. + Sort. and (4) the proposed enhanced configuration Rev. Rec. Plus ($i.e.,$ Rev. Rec. + Pred. + Sort. ) } (introduced in Sec. \ref{other_pretext_tasks}). The results are exhibited in Table \ref{pretext_results_comp}, and we draw the following conclusions:
\textbf{(1)} When a single pretext task is used for self-supervised learning, reverse reconstruction typically performs comparably or superior to other pretext tasks in terms of both AP and SC. This is due to the fact that reverse reconstruction utilizes the skeleton order information embedded in inputs, which enables the exploitation of the intra-sequence locality during training. Such results also justify the center role of reverse reconstruction in self-supervised learning.
\textbf{(2)} \hc{Combining gait features learned from different pretext tasks could achieve higher Re-ID performance than using Rev. Rec. alone in many cases. Notably, the enhanced configuration Rev. Rec. Plus, which combines CAGEs learned from all three pretext tasks, obtains the best overall performance in terms of \textit{Rank-1} and \textit{nAUC} on three out of four datasets (BIWI, KS20, KGBD).}
Such observations reveal the potential to extract richer gait features for performance improvement through introducing more pretext tasks into self-supervised learning. Nevertheless, Rev. Rec. Plus requires learning three different pretext tasks, and suffers from higher computational cost and feature dimension. By contrast, reverse reconstruction usually obtains similarly competitive performance with much less training cost and simpler gait representations. Hence, we recommend to use reverse reconstruction as the primary pretext task, and our later analysis is also performed based on this pretext task.
\textbf{(3)} As shown in Table \ref{pretext_results_comp}, using AP  almost constantly achieves better Re-ID performance. The reason is that AP is able to reduce the influence of noisy frames that give wrong predictions, so it can encourage better sequence-level predictions.

\subsubsection{Different Attention Mechanisms}
In order to show the effects of attention mechanisms, we evaluate the person Re-ID performance of our approach under four different cases (no attention, BAS, MBAS, or LAS). To provide a more comprehensive evaluation, we also evaluate attention mechanisms with different sequence lengths.
By results reported in Fig. \ref{attention_comp}, we can draw the following conclusions:
\textbf{(1)} The application of attention mechanisms (BAS, MBAS, LAS) can improve the model performance by $1.0\%$-$7.2\%$ \textit{Rank-1} accuracy when compared with the case without attention mechanism. This is because the attention mechanisms can help the model focus on more correlative skeletons, thus leading to better sequence reconstruction and more effective gait representations for person Re-ID. \textbf{(2)} Among different attention mechanisms, the proposed locality-aware attention mechanism (LAS) is the best performer, which surpasses BAS and MBAS by
up to $7.9\%$ \textit{Rank-1} accuracy on different datasets. These results justify our claim that intra-sequence locality enables better gait representation learning for person Re-ID.

\subsubsection{Different Contrasting Intervals}
In this section, we discuss the performance of our approach under different contrasting intervals. To be more specific, here we not only contrast adjacent skeleton sequences (Interval$=$1) by the proposed LCL scheme, but also exploit those non-adjacent skeleton sequences (Interval$>$1) for contrastive learning. As shown in Table \ref{contrastive_interval_results}, when compared with other contrasting interval settings, using adjacent sequences to perform LCL scheme constantly achieves the best \textit{Rank-1} accuracy as well as \textit{nAUC} on all datasets. The comparison validates that adjacent skeleton sequences in a local temporal context enjoy higher correlations. In other words, such results justify our motivation to integrate the inter-sequence locality, which can be effectively learned by our LCL scheme, for the enhancement of the gait encoding for person Re-ID.

\begin{figure*}[t]
    \centering
     \subfigure[\textit{Rank-1} Accuracy Comparison]{\scalebox{0.35}{\label{transfer_rank}\includegraphics[]{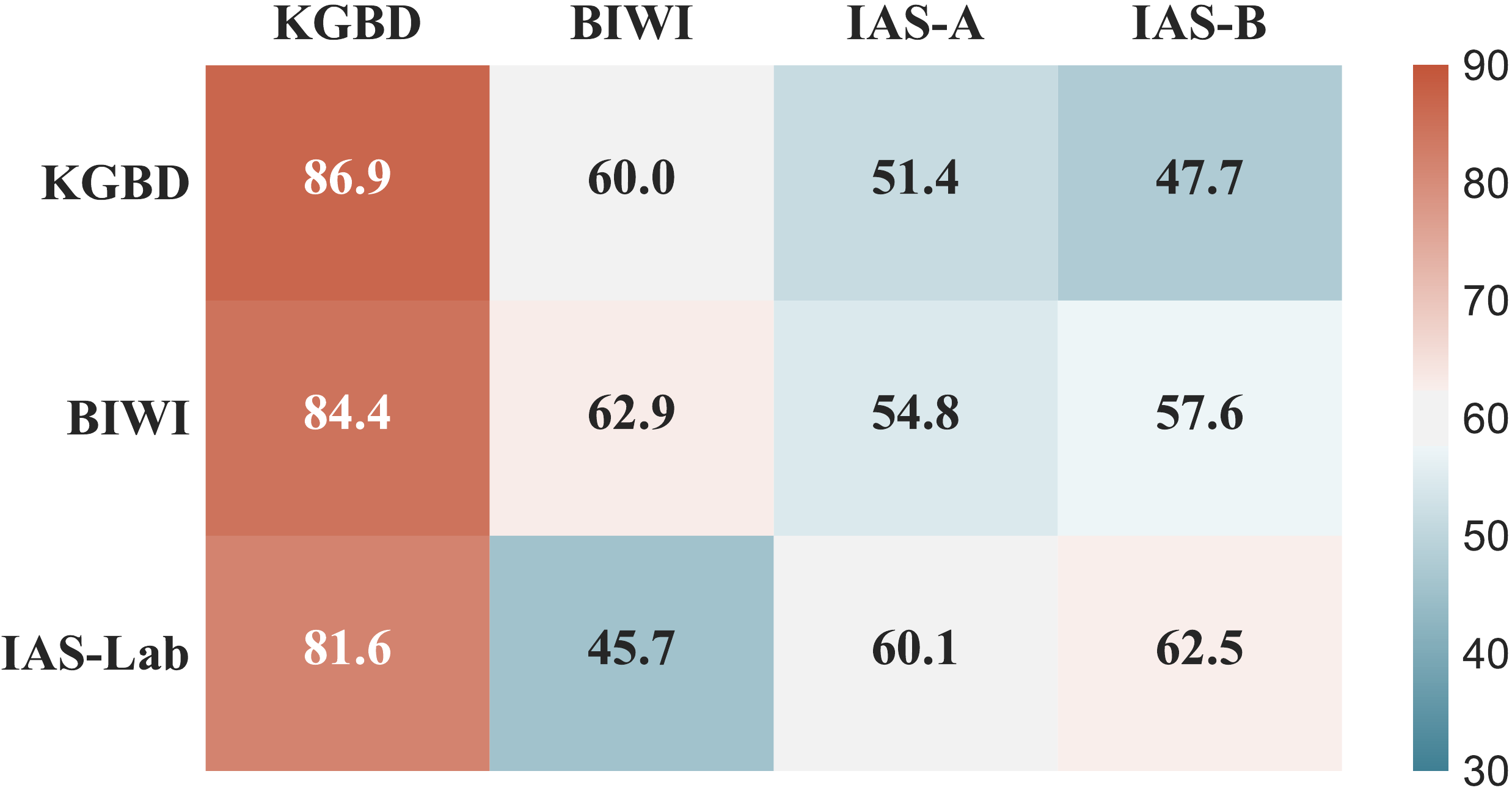}}
     }
     \ \ 
    \subfigure[\textit{nAUC} Comparison]{\scalebox{0.35}{\label{transfer_nauc}\includegraphics[]{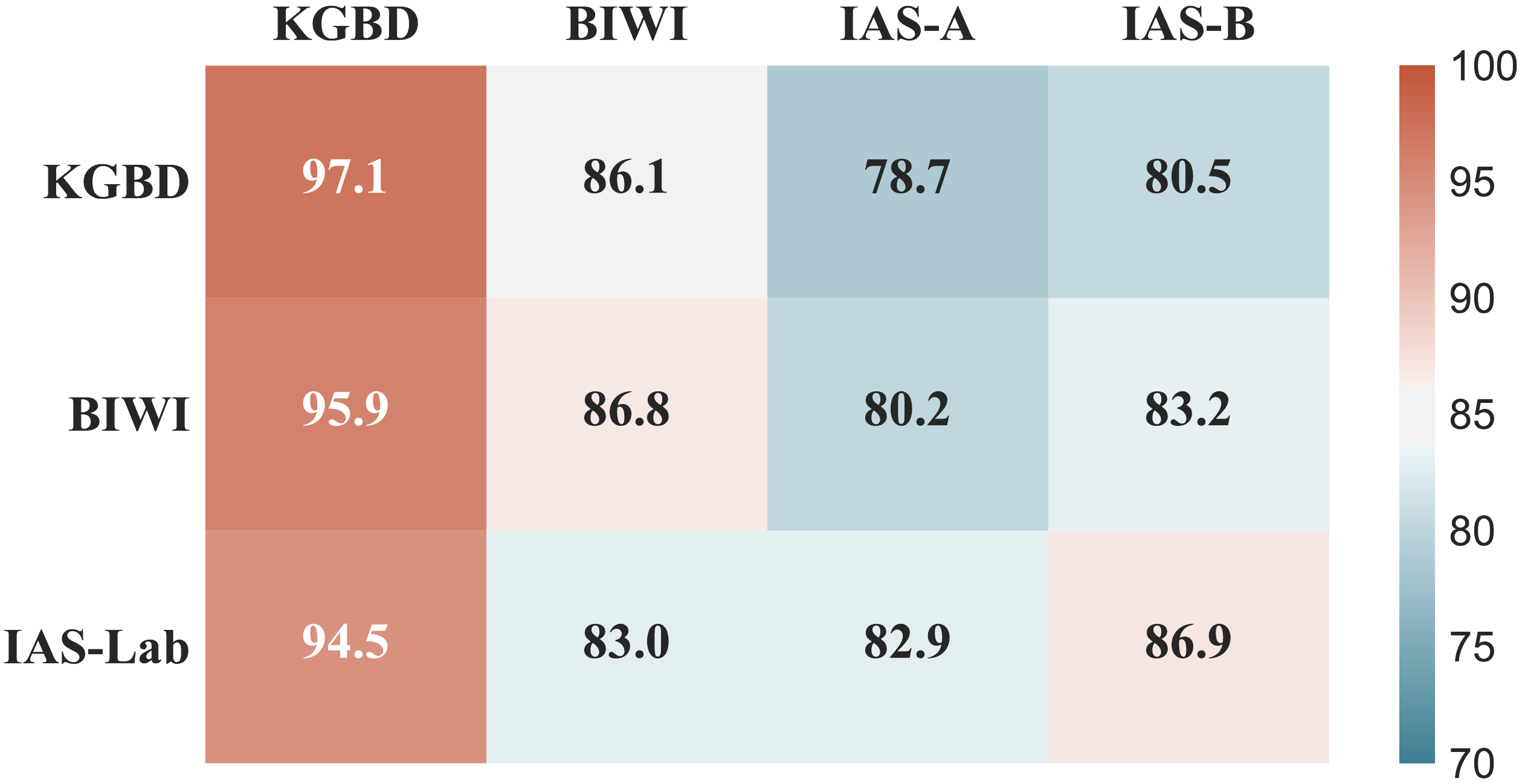}}
     }
    \caption{\textit{Rank-1} accuracy and \textit{nAUC} comparison between the original model and the transferred model on different datasets. Note that the abscissa and ordinate denote target datasets and source datasets (for training gait encoding models) respectively.}
    \label{transfer_comp}
\end{figure*}

\subsubsection{Temperature Setting for LCL Scheme}
\label{CAGEs_vs_AGEs_Sec}
When the proposed LCL scheme is applied, we need to determine the value of temperature $\tau$. In this section, we evaluate the performance of our approach when different values are set to the temperature $\tau$ of LCL scheme. As shown in Table \ref{CAGE_t}, we observe that the performance of our model typically enjoys a stable performance when the value of $\tau$ is varied, and no drastic change is observed. The results suggest that out LCL scheme is actually insensitive to $\tau$, so we simply set $\tau$ empirically in our experiments. 

\subsection{Transferability of Gait Encoding Model}
\label{transfer}
Interestingly, we discover that the gait encoding model learned on one dataset can be readily transferred to other datasets. Specifically, we use the gait encoding model pre-trained on training sets (``source datasets'') from KGBD, BIWI or IAS-Lab to directly encode 3D skeleton data from other datasets (``target datasets'') into CAGEs, which are then used for training the recognition network $f_{RN}$ to perform person Re-ID.  
We compare their Re-ID performance with the gait encoding model trained on their original training sets: As shown by Fig. \ref{transfer_comp}, the transferred gait encoding models ($i.e.,$ trained on a different source dataset) are also able to achieve highly competitive performance when compared with existing methods in literature, despite that our original model is still the best performer. Such transferability enables the pre-trained model to extract discriminative gait features from unseen skeleton data of a new dataset, which demonstrates that our approach indeed captures transferable high-level semantics of 3D skeleton data. 




\hc{\subsection{Evaluation on Model-Estimated Skeleton Data} 
\label{CASIA_B_evaluation}

\begin{table*}[t]
\centering
\hc{\caption{Rank-1 accuracy on different views of CASIA B under CVE setup.}
\label{CVE_comparison}}
\scalebox{0.95}{
\setlength{\tabcolsep}{4.0mm}{
\begin{tabular}{lccccccccccc}
\specialrule{0.1em}{0.45pt}{0.45pt}
\multicolumn{1}{l}{\textbf{Methods}} & $\boldsymbol{0^{\circ}}$    & $\boldsymbol{18^{\circ}}$   & $\boldsymbol{36^{\circ}}$   & $\boldsymbol{54^{\circ}}$   & $\boldsymbol{72^{\circ}}$   & $\boldsymbol{90^{\circ}}$   & $\boldsymbol{108^{\circ}}$  & $\boldsymbol{126^{\circ}}$  & $\boldsymbol{144^{\circ}}$  & $\boldsymbol{162^{\circ}}$  & $\boldsymbol{180^{\circ}}$  \\ \specialrule{0.1em}{0.45pt}{0.45pt}
PoseGait \cite{liao2020model}                             & 10.7          & 37.4          & 52.5          & 28.3          & 24.3          & 18.9          & 23.5          & 17.2          & 23.6          & 18.8          & 4.3           \\
\textbf{Ours}                        & \textbf{35.5} & \textbf{78.5} & \textbf{79.5} & \textbf{58.0} & \textbf{66.1} & \textbf{76.0} & \textbf{64.4} & \textbf{66.9} & \textbf{68.4} & \textbf{49.9} & \textbf{46.3} \\ \specialrule{0.1em}{0.2pt}{0.2pt}
\end{tabular}
}
}
\end{table*}

\begin{table}[t]
\centering
\hc{\caption{Rank-1 matching rate on CASIA B compared with appearance-based methods under CME setup. Note: ``Cl-Nm'' denotes the probe set (under ``Clothes'' condition) and gallery set (under ``Normal'' condition).}
\label{CME_comparison}}
\scalebox{0.90}{
\setlength{\tabcolsep}{1.0mm}{
\begin{tabular}{lccccc}
\specialrule{0.1em}{0.45pt}{0.45pt}
                  \textbf{Methods}  & \textbf{Nm-Nm} & \textbf{Bg-Bg} & \textbf{Cl-Cl} & \textbf{Cl-Nm} & \textbf{Bg-Nm} \\ \specialrule{0.1em}{0.45pt}{0.45pt}
LMNN \cite{weinberger2009distance}                & 3.9            & 18.3           & 17.4           & 11.6           & 23.1           \\
ITML \cite{davis2007information}                & 7.5            & 19.5           & 20.1           & 10.3           & 21.8           \\
EFL \cite{gray2008viewpoint}                 & 12.3           & 5.8            & 19.9           & 5.6            & 17.1           \\
SDALF \cite{farenzena2010person}               & 4.9            & 10.2           & 16.7           & 11.6           & 22.9           \\
Score-Level + MLR \cite{liu2015enhancing} & 13.6           & 13.6           & 13.5           & 9.7            & 14.7           \\
Feature-level + MLR \cite{liu2015enhancing} & 16.3           & 18.9           & 25.4           & 20.3           & 31.8           \\
\textbf{Ours}       & \textbf{54.3}  & \textbf{37.5}  & \textbf{31.9}  & \textbf{27.0}  & \textbf{36.3}  \\ \specialrule{0.1em}{0.2pt}{0.2pt}
\end{tabular}
}
}
\end{table}

To further evaluate our skeleton-based approach on the large-scale gait dataset CASIA B, we exploit pre-trained pose estimation models \cite{cao2019openpose,chen20173d} to extract skeleton data from RGB videos of CASIA B, and evaluate the performance of our approach with the estimated skeleton data. We compare our model with the latest skeleton-based method PoseGait \cite{liao2020model} and representative appearance-based methods \cite{weinberger2009distance,davis2007information,gray2008viewpoint,farenzena2010person,liu2015enhancing}, and we can draw the following observations and conclusions: 

\textbf{(1)} As shown in Table \ref{CVE_comparison}, our approach outperforms the latest skeleton-based model PoseGait by a significant margin ($24.8\%$-$57.1\%$ \textit{Rank-1} accuracy) on all views of CASIA B. Notably, both methods achieve their own best performance on the view $36^\circ$, while PoseGait is still inferior to our method by $27\%$ \textit{Rank-1} accuracy. On the two most challenging views ($0^\circ$ and $180^\circ$), our approach also achieves
better Re-ID performance than PoseGait by more than $24\%$ \textit{Rank-1} accuracy.
\textbf{(2)} In Table \ref{CME_comparison}, our skeleton-based approach could also outperform representative classic appearance-based methods that utilize visual features ($e.g.,$ RGB features, silhouettes). For example, our method outperforms LMNN \cite{weinberger2009distance} and ITML \cite{davis2007information}, which leverage visual features (RGB, HSV color and texture) and metric learning for recognition \cite{liu2015enhancing}, with an evident performance gain up to $50.4\%$ on different CME settings. Compared with the method that fuses RGB appearance features and GEI features (``Feature-level + MLR'' \cite{liu2015enhancing}), our model also yields evidently better \textit{Rank-1} matching rate by $4.5\%$-$38.0\%$ on all conditions of CASIA B. Therefore, despite that our approach is trained on the estimated skeleton data with noises, the learned gait representations still show higher discriminative power and achieve superior performance to those appearance-based methods on CASIA B. These results show the effectiveness of our approach on multi-view Re-ID tasks when using model-estimated skeleton data, and also demonstrate the great potential of our approach to be applied to large RGB-based datasets.

}

\hc{
\subsection{Visualization of Typical Samples}
\begin{figure*}[t]
    \centering
     \subfigure[Confusion Matrix (KS20)]{\scalebox{0.55}{\includegraphics[]{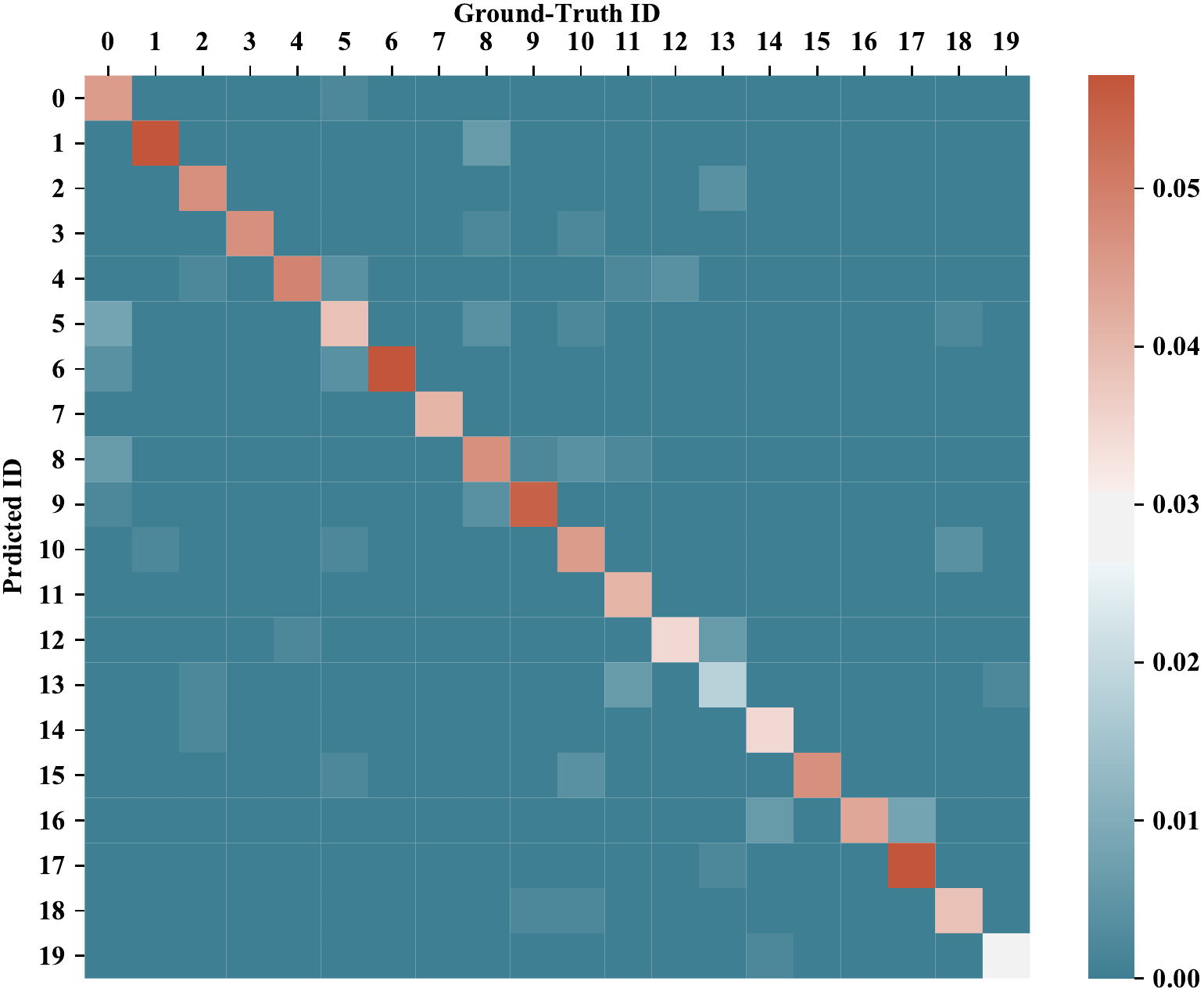}}
     }
     \ \ 
    \subfigure[Confusion Matrix (BIWI)]{\scalebox{0.55}{\includegraphics[]{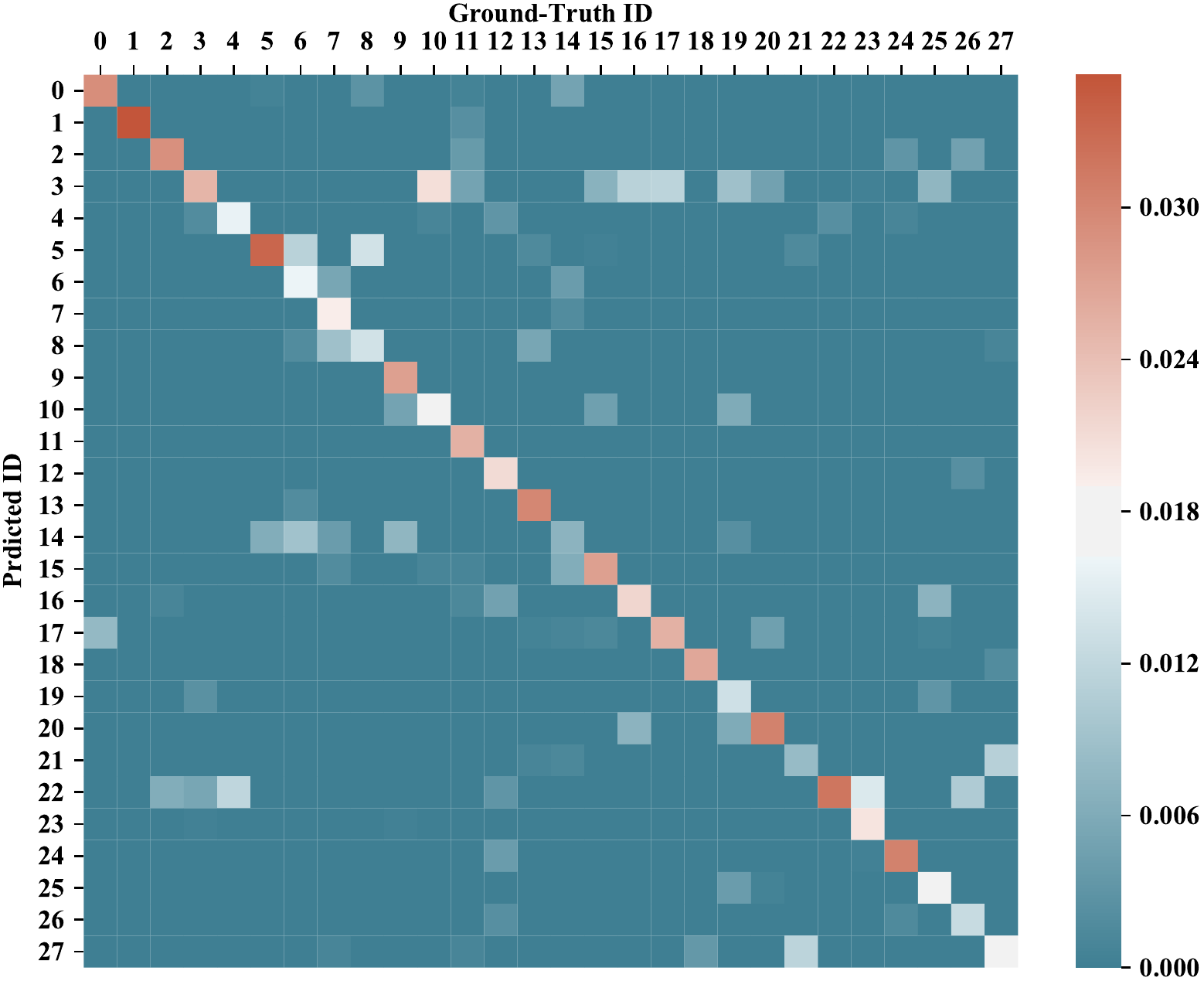}}
     }
    \caption{\hc{Confusion matrices of KS20 and BIWI. Note that abscissa  and ordinate denote the ground-truth and predicted IDs respectively. The position in the $a^{th}$ column and $b^{th}$ row indicates that the testing samples belonging to the $a^{th}$ ID is predicted as the $b^{th}$ ID, while the corresponding value is the proportion of such samples to samples in the whole testing set. 
   Full results are provided in the supplementary material.}}
    \label{confusion_matrices}
\end{figure*}

\begin{figure}[t]
    \centering
     \subfigure[Id. = 9, Pred. = 9 ]{\scalebox{0.155}{\includegraphics[]{main/IAS_right.pdf}}
     }
     \quad
     \subfigure[Id. = 9, Pred. = 8]{\scalebox{0.155}{\includegraphics[]{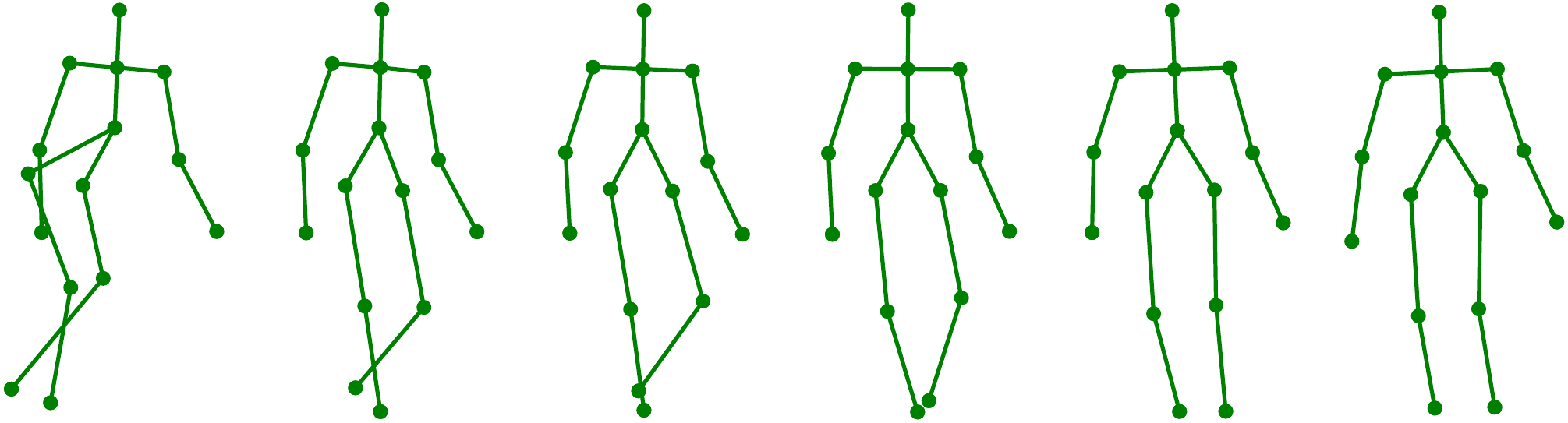}}
     }
     \quad 
        \subfigure[Id. = 8, Pred. = 8]{\scalebox{0.145}{\includegraphics[]{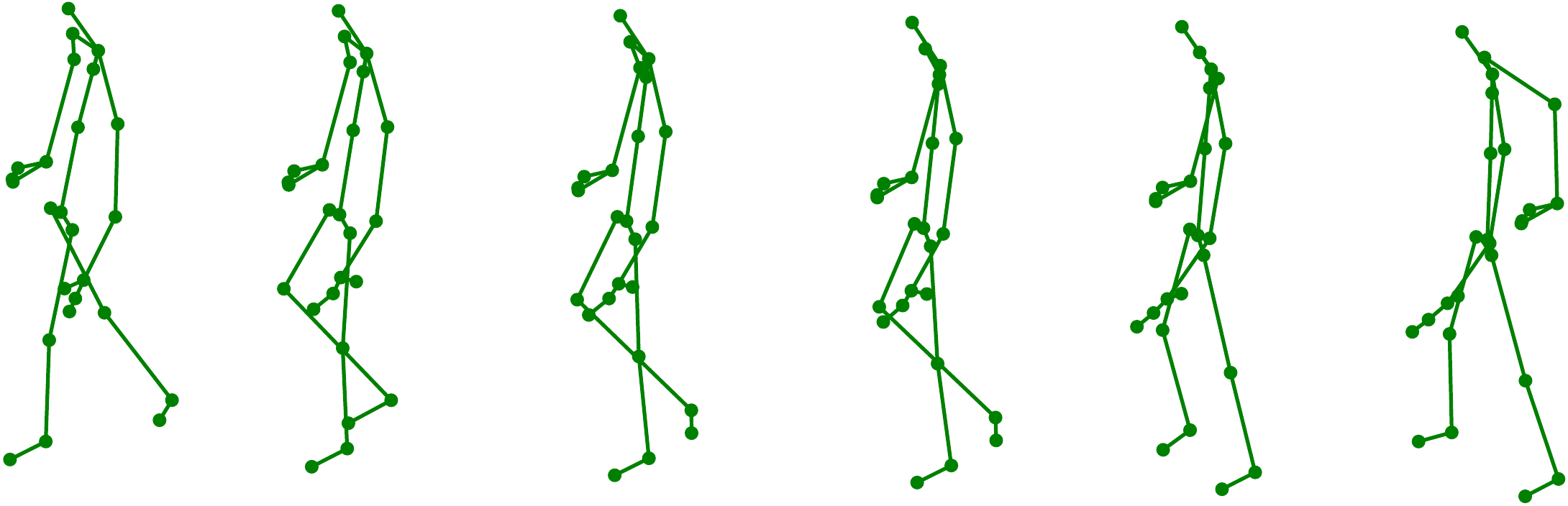}}
     }
     \quad
    \subfigure[Id. = 8, Pred. = 9]{\scalebox{0.138}{\includegraphics[]{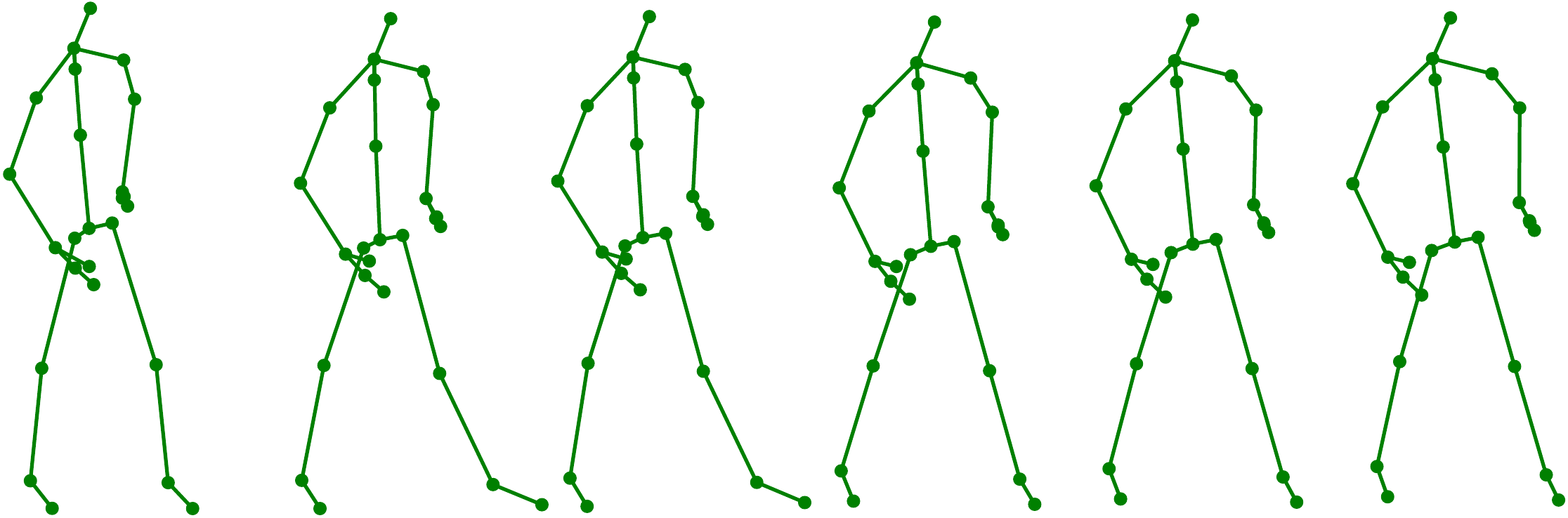}}
     }
          \quad 
            \subfigure[Id. = 7, Pred. = 7]{\scalebox{0.115}{\includegraphics[]{main/BIWI_right.pdf}}
     }
     \quad
    \subfigure[Id. = 7, Pred. = 6]{\scalebox{0.115}{\includegraphics[]{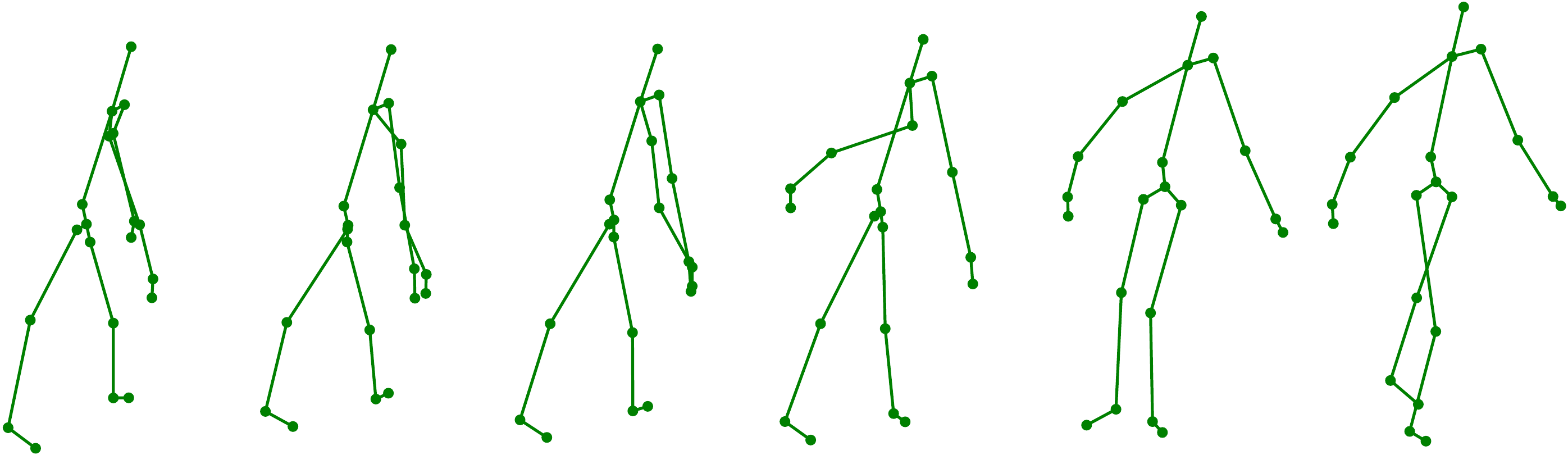}}
    }
      \quad  \
        \subfigure[Id. = 6, Pred. = 6]{\scalebox{0.085}{\includegraphics[]{main/CASIA_right_1.pdf} \quad \quad
         \includegraphics[]{main/CASIA_right_2.pdf}}
     }
     \quad 
    \subfigure[Id. = 6, Pred. = 7]{\scalebox{0.085}{\includegraphics[]{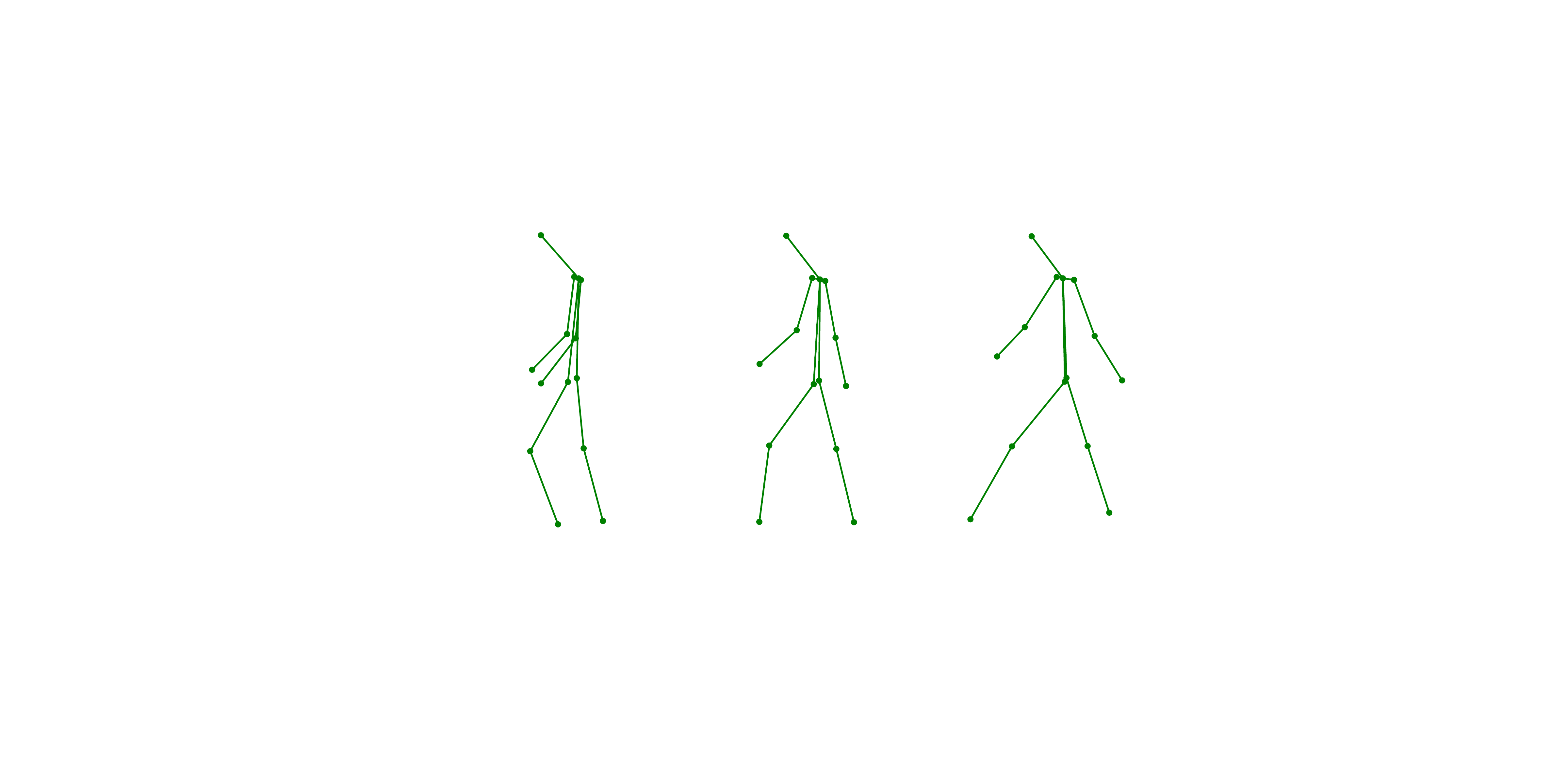} \quad \quad
         \includegraphics[]{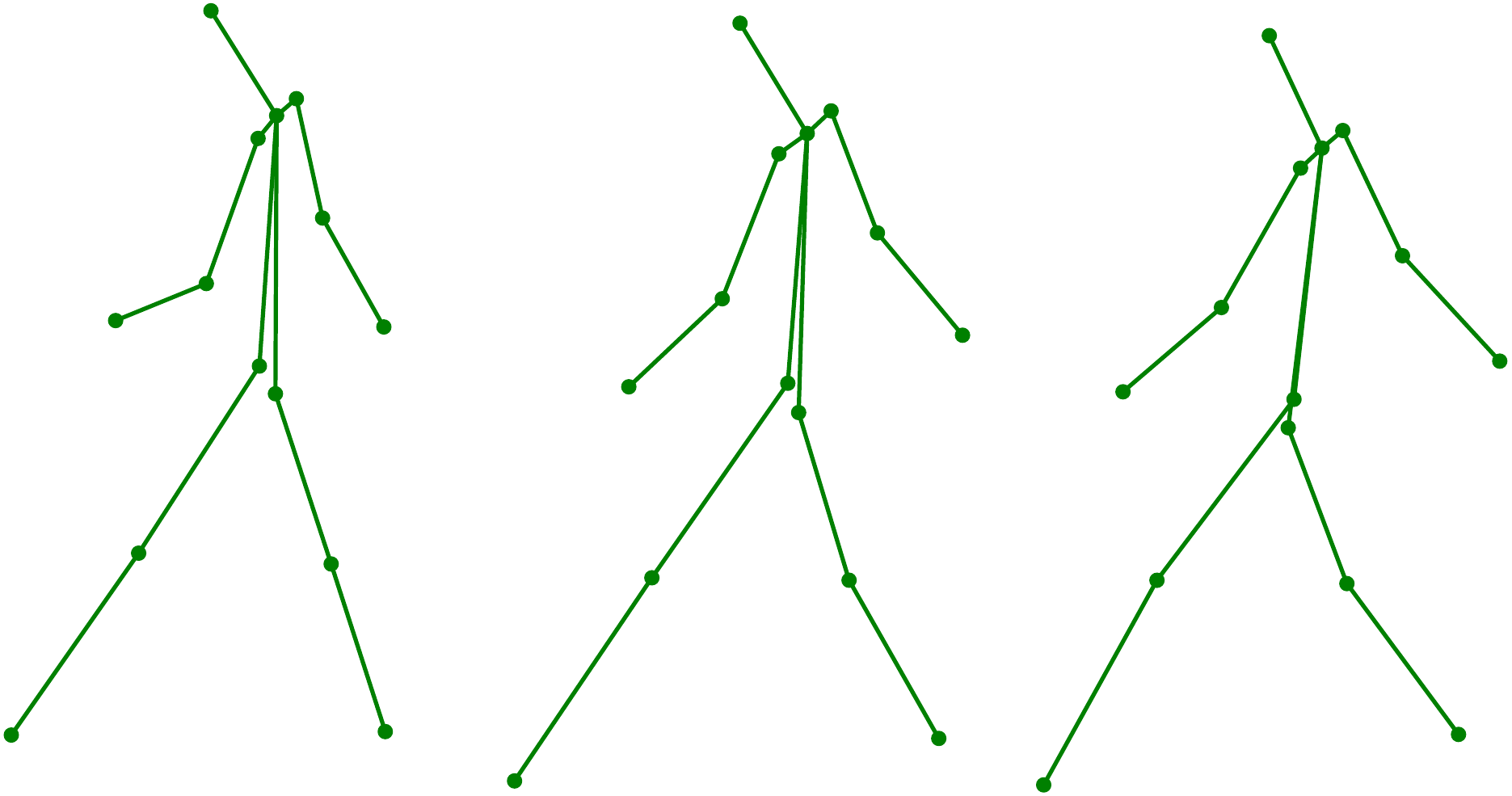}}
    }
    \caption{\hc{Visualization of typical samples in testing sets of IAS-B (a-b), KS20 (c-d), BIWI (e-f), CASIA B (g-h). Note: ``Id.'' and ``Pred.'' denotes the ground-truth label and predicted label respectively. (g) and (h) are partial samples of testing sequences on the view $90^\circ$ of CASIA B.}}
    \label{Skeleton_visual}
\end{figure}
In Fig. \ref{Skeleton_visual}, we visualize both simple samples and hard samples for Re-ID from testing sequences of different datasets. To this end, we select the classes that are most likely to be confused ($e.g.,$ ID 6 and 7 in BIWI, shown in Fig. \ref{confusion_matrices} (b)) for visualization. We obtain the following observations:  \textbf{(1)} Some skeletons from those datasets indeed contain noise. For example, in Fig. \ref{Skeleton_visual} (b), the first skeleton of the sequence suffers from obvious twisting, which can be ascribed to wrongly detected body joints, while similar noise can be observed from the former three skeletons in Fig. \ref{Skeleton_visual} (f).
In the mean time, we notice that skeleton sequences contaminated by noise are often wrongly classified, which suggest that noise degrades the person Re-ID performance.
 \textbf{(2)} The skeleton sequences that contain salient motion are more likely to be correctly classified. For example, a skeleton sequence with intense action (see Fig. \ref{Skeleton_visual} (c)) can be easily classified by our model, while those sequences with comparatively static skeletons are often hard to recognize (see Fig. \ref{Skeleton_visual} (d)).
 This actually validates that salient gait patterns in skeleton sequences will facilitate person Re-ID.
}
\section{Conclusion and Future Work}
\label{conclusion}
In this paper, we propose a generic self-supervised approach with locality-awareness to learn effective gait representations for person Re-ID. We introduce self-supervision by learning reverse skeleton sequence reconstruction as a primary pretext task, which enables our model to learn high-level semantics and discriminative gait features with unlabeled skeleton data. Other potential pretext tasks like sorting and prediction are also explored and synthesized into the self-supervised learning. To facilitate skeleton reconstruction and gait representation learning, a novel locality-aware attention mechanism and locality-aware contrastive learning scheme are proposed to incorporate the intra-sequence and inter-sequence locality into gait encoding process. Last, we propose to construct the final gait representations (CAGEs) for person Re-ID with learned context vectors. Our approach significantly outperforms existing skeleton-based Re-ID methods, and its performance is comparable or superior to depth-based and multi-modal methods. \hc{Besides, we show that our approach can be applied to 3D skeleton data estimated from large RGB-based datasets, and achieve better performance than many classic appearance-based methods.}

\hc{\textbf{Limitations.} There are three limitations in our work: First, the scale of skeleton datasets in our experiments is relatively limited when compared with popular RGB-based Re-ID datasets like DukeMTMC-reID, since large-scale skeleton-based Re-ID datasets that contain more individuals and scenarios are still unavailable. Hopefully, we will collect our own skeleton-based Re-ID datasets in the future. Second, this work basically considers the case where the skeletons are collected with relatively high quality ($e.g.,$ by device like Kinect), while the case where skeleton data are collected under a more general setting ($e.g.,$ estimated from RGB data in outdoor scenes) has not been thoroughly studied.
Finally, our approach models 3D skeletons as body-joint sequences on different dimensions, while it may be insufficient to capture underlying relations between body joints. }

\hc{Overall,} our approach showcases the effectiveness of self-supervised gait encoding on the person Re-ID task, \hc{and there are several potential directions for improvement:} \textbf{(1)} More efficient pretext tasks ($e.g.$, frame
interpolation, skeleton video generation) could be explored to improve the capture of motion semantics for better gait encoding. \hc{\textbf{(2)} Modeling 3D skeletons as graphs is able to mine richer relation information among body joints, while employing graph-based encoders ($e.g.$, graph convolutional network (GCN)) could enhance structural feature learning from skeleton graphs.} \textbf{(3)} Fine-grained spatial-temporal attention mechanisms could also be designed to extract those crucial motion patterns for person Re-ID, and more effective skeleton augmentation strategies could be considered to enhance the contrastive learning. \hc{\textbf{(4)} One important future direction is to explore 3D skeleton-based Re-ID under the general setting, so as to improve the model robustness to noise, $e.g.,$ noise incurred by skeleton/pose estimation. }
\textbf{(5)} Our model can be extended to more skeleton-related tasks, and we can expect it to be readily transferred to multi-modal learning for other pivotal vision tasks.

\rhc{\section{Ethical Statements}
Person Re-ID is an important topic with huge potential value in computer vision. However, it should be noted that improper application or abuse of Re-ID technology will pose a grave threat to the society and public privacy. Thus, we want to emphasize that benchmark datasets used in this work are either publicly available (BIWI, IAS-Lab, KGBD) or officially authorized (KS20, CASIA B). The official agents of those datasets have guaranteed that all data are collected, released, and used with the consent of subjects. All people in datasets are anonymous with simple identity numbers for privacy protection. Besides, our approach and models must only be used for research purpose.}



%



\ifCLASSOPTIONcompsoc
  \section*{Acknowledgments}
\else
  \section*{Acknowledgment}
\fi
This work was supported in part by the National Key Research and Development Program of China (Grant No. 2019YFA0706200), in part by the National Natural Science Foundation of China (Grant No. 61632014, No. 61627808, No. 62006236, No. 62072190), in part by the Hunan Provincial Natural Science Foundation (Grant No. 2020JJ5673), and in part by the NUDT Research Project (Grant No. ZK20-10).


\ifCLASSOPTIONcaptionsoff
  \newpage
\fi


\bibliographystyle{IEEEtran}
\bibliography{IEEEabrv,IEEEexample}
%



%

\begin{IEEEbiography}[{\includegraphics[width=1in,height=1.25in,clip,keepaspectratio]{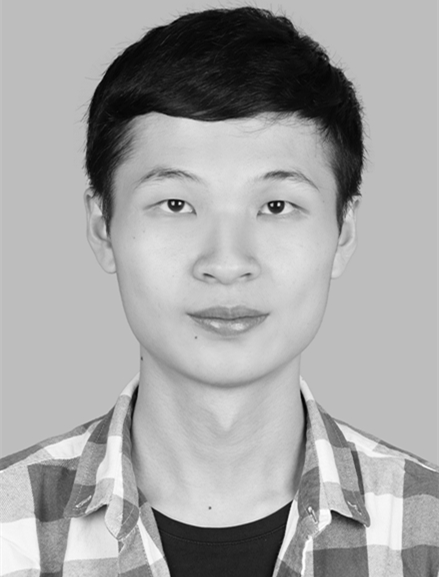}}]{Haocong Rao} received the B.Eng degree from South China University of Technology, China, in 2019. He was a visiting student with Shenzhen Institute of Advanced Technology, Chinese Academy of Sciences, China, in 2020. He is currently working toward the PhD degree at Nanyang Technological University, Singapore. He has authored/coauthored more than 5 peer-reviewed papers in highly regarded journals and conferences such as IJCAI, ACMMM, IEEE IoT journal, Information Sciences, etc. He received the Best Paper Award from IEEE Heathcom 2020. His research interests include skeleton-based person re-identification, self-supervised learning, domain adaptation and interpretable artificial intelligence.
\end{IEEEbiography}

\begin{IEEEbiography}[{\includegraphics[width=1in,height=1.25in,clip,keepaspectratio]{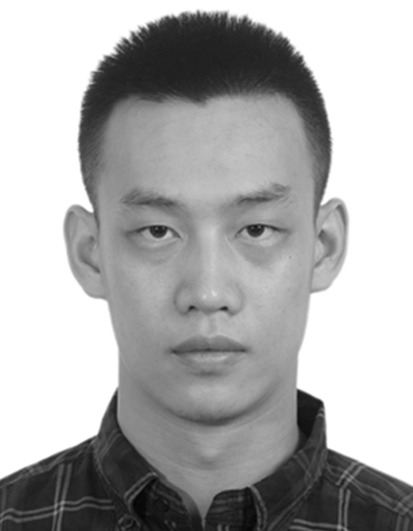}}]{Siqi Wang} received the BS degree and the PhD degree in computer science and technology from the National University of Defense Technology, China. He is currently an assistant professor with the State Key Laboratory of High Performance Computing (HPCL), National University of Defense Technology, China. His main research focuses on anomaly / outlier detection, pattern recognition and unsupervised learning. His works have been published on leading conferences and journals, such as NeurIPS, AAAI, ACM MM, Pattern Recognition, IEEE Transactions on Cybernetics and Neurocomputing. He also serves as a reviewer for several international journals, including the IEEE Transactions on Cybernetics, the IEEE Transactions on Automation Science and Engineering, Artificial Intelligence Review, and International Journal of Machine Learning and Cybernetics.
\end{IEEEbiography}


\begin{IEEEbiography}[{\includegraphics[width=1in,height=1.25in,clip,keepaspectratio]{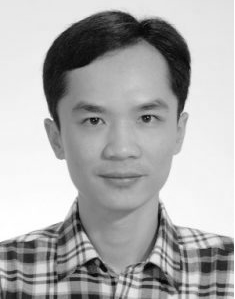}}]{Xiping Hu} received the Ph.D. degree from the University of British Columbia, Vancouver, BC, Canada. He is currently a professor with Shenzhen Institute of Advanced Technology, Chinese Academy of Sciences and Lanzhou University, China. He was the Co-Founder and CTO of Bravolol Ltd., Hong Kong, a leading language learning mobile application company with over 100 million users, and listed as the top 2 language education platform globally. He has more than 100 papers published and presented in prestigious conferences and journals, such as IEEE TMC/TPDS/TIP/JSAC/IoT journal, IEEE COMST, IEEE COMMUNICATIONS MAGAZINE, MobiCom, AAAI, IJCAI, and WWW. He has been serving as the lead guest editors of IEEE Internet of Things Journal, IEEE Transactions on Automation Science and Engineering, and WCMC. His research areas consist of mobile cyber-physical systems, crowdsensing and affective computing.
\end{IEEEbiography}


\begin{IEEEbiography}[{\includegraphics[width=1in,height=1.25in,clip,keepaspectratio]{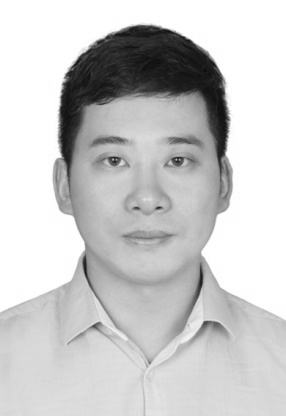}}]{MingKui Tan} is currently a professor with the School of Software Engineering at South China University of Technology. He received his Bachelor Degree in Environmental Science and Engineering in 2006 and Master degree in Control Science and Engineering in 2009, both from Hunan University in Changsha, China. He received the Ph.D. degree in Computer Science from Nanyang Technological University, Singapore, in 2014. From 2014-2016, he worked as a Senior Research Associate on computer vision in the School of Computer Science, University of Adelaide, Australia. His research interests include machine learning, sparse analysis, deep learning and large-scale optimization.
\end{IEEEbiography}


\begin{IEEEbiography}[{\includegraphics[width=1in,height=1.25in,clip,keepaspectratio]{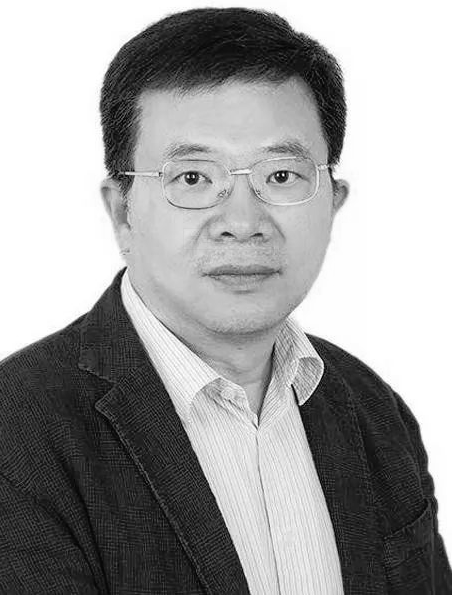}}]{Yi Guo} received his Ph.D. degree from
the University of Greifswald, Germany, in 1997. He is currently
the chief of Neurology in the Second Clinical Medical College
of Jinan University, a member of the cerebrovascular disease
group of the Chinese Medical Association neurology branch,
and the chairman of the Shenzhen Medical Association of Neurology and the Shenzhen Medical Association of Psychosomatic
Medicine. His major research areas are cerebrovascular diseases, dementia, sleep disorder, depression, and anxiety.
\end{IEEEbiography}


\begin{IEEEbiography}[{\includegraphics[width=1in,height=1.25in,clip,keepaspectratio]{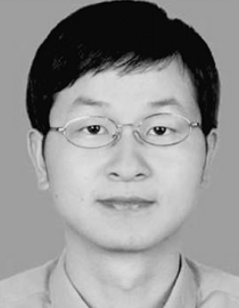}}]{Jun Cheng} received the B.Eng. and M.Eng. degrees from the University of Science and Technology of China, Hefei, China, in 1999 and 2002, respectively, and the Ph.D. degree from the Chinese University of Hong Kong, Hong Kong, in 2006. He is currently a Professor and the Founding Director of the Laboratory for Human Machine Control, Shenzhen Institute of Advanced Technology, Chinese Academy of Sciences, Shenzhen, China. He has authored or coauthored about 110 articles. His current research interests include computer visions, robotics, and machine intelligence and control.
\end{IEEEbiography}


\begin{IEEEbiography}[{\includegraphics[width=1in,height=1.25in,clip,keepaspectratio]{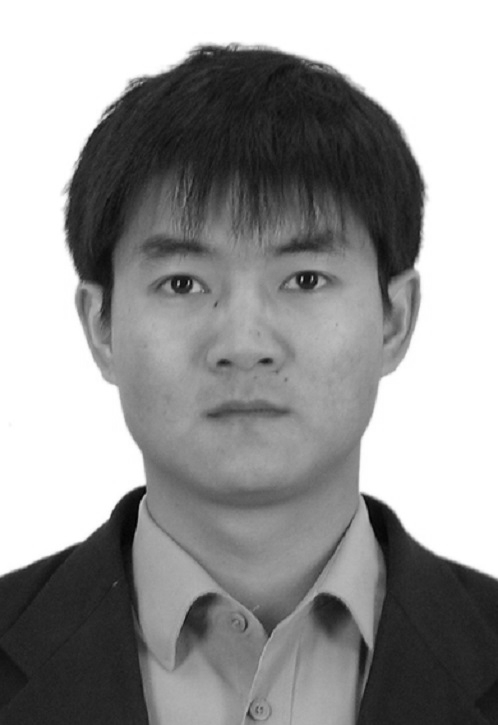}}]
{Xinwang Liu} received his PhD degree from National University of Defense Technology (NUDT), China. He is now Professor of School of Computer, NUDT. His current research interests include kernel learning and unsupervised feature learning. Dr. Liu has published 60+ peer-reviewed papers, including those in highly regarded journals and conferences such as IEEE  IEEE Transactions on Pattern Analysis and Machine Intelligence, IEEE Transactions on Knowledge and Data Engineering, IEEE Transactions On Image Processing, IEEE Transactions on Neural Networks and Learning Systems, IEEE Transactions on Multimedia, IEEE Transactions on Information Forensics and Security, NeurIPS, ICCV, CVPR, AAAI, IJCAI, etc.
\end{IEEEbiography}

\begin{IEEEbiography}[{\includegraphics[width=1in,height=1.25in,clip,keepaspectratio]{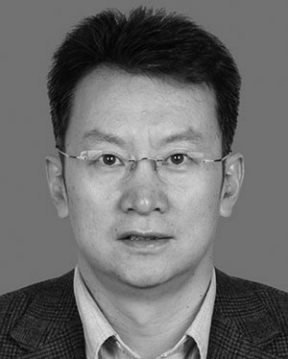}}]{Bin Hu} (M'05-SM'10) is currently a Professor of the School of Information Science and Engineering, Lanzhou University, Adjunct Professor with Tsinghua University, Beijing, China, and Guest Professor with ETH Zurich, Zurich, Switzerland. He is also IET Fellow, Co-Chairs of IEEE SMC TC on Cognitive Computing, and Member at Large of ACM China, Vice President of International Society for Social Neuroscience (China committee), etc. His work has been funded as a PI by the Ministry of Science and Technology, National Science Foundation China, European Framework Programme 7, EPSRC, and HEFCE UK, etc., also published more than 100 papers in peer-reviewed journals, conferences, and book chapters including Science, Journal of Alzheimer’s Disease, PLoS Computational Biology, IEEE TRANSACTION ON INTELLIGENT SYSTEMS, AAAI, BIBM, EMBS, CIKM, ACM SIGIR, etc. He has served as Chairs/Co-Chairs in many IEEE international conferences/workshops, and Associate Editors in peer-reviewed journals on Cognitive Science and Pervasive Computing, such as IEEE TRANSACTION ON AFFECTIVE COMPUTING, Brain Informatics, IET Communications, Cluster Computing, Wireless Communications, and Mobile Computing, The Journal of Internet Technology, Wiley’s Security and Communication Networks, etc.
\end{IEEEbiography}






\end{document}